\documentclass[runningheads]{llncs}

\usepackage{eccv}

\usepackage{eccvabbrv}

\usepackage{graphicx}
\usepackage{booktabs}
\usepackage[table]{xcolor}

\usepackage{graphicx}
\usepackage{booktabs}
\usepackage{multicol}
\usepackage{colortbl}
\usepackage{tikz,pgfplots}
\pgfplotsset{compat=newest}
\usepackage{tcolorbox}
\usepackage{algorithm}
\usepackage{algpseudocode}
\usepackage{silence}
\usepackage{fontawesome5}
\usepackage{multirow}

\usepackage[accsupp]{axessibility}  

\usepackage{hyperref}
\usepackage{orcidlink}

\definecolor{mygreen}{RGB}{0, 170, 95}
\definecolor{myred}{RGB}{190, 60, 60}
\definecolor{mygray}{gray}{0.65}
\definecolor{lightblue}{HTML}{cfedfc}
\definecolor{lightred}{HTML}{fae0d2}
\definecolor{lightpink}{HTML}{ffdee5}
\definecolor{lightgreen}{HTML}{0FBD83}
\definecolor{lightyellow}{HTML}{FAA410}
\definecolor{palegreen}{HTML}{9da894}
\definecolor{deepblue}{HTML}{6A95B1}
\definecolor{color1}{HTML}{006EB8}
\definecolor{topThreeBlue}{HTML}{08306B} 
\definecolor{topOneBlue}{HTML}{90C5E0}   
\definecolor{clusterGold}{HTML}{FFB300}  
\definecolor{thrTeal}{HTML}{668A80}      
\definecolor{pinkishred}{HTML}{EA5C5C}      
\definecolor{red2}{RGB}{252, 54, 65}

\newcommand{\ours}{\textsc{B2G}\xspace}
\newcommand{\oursfix}{\textsc{B2G}\textsubscript{fix}\xspace}
\newcommand{\oursdynclus}{\textsc{B2G}\textsubscript{dyn-clus}\xspace}
\newcommand{\oursdynperm}{\textsc{B2G}\textsubscript{dyn-perm}\xspace}

\newcommand{\gc}[1]{\textcolor{gray}{#1}}
\newcommand{\pc}[1]{\textcolor{topThreeBlue}{#1}}
\newcommand{\hc}[1]{\textcolor{black}{#1}}

\hypersetup{
  colorlinks   = true,
  urlcolor     = color1,
  linkcolor    = gray,
  citecolor   = topOneBlue
}

\usepackage{tikz}
\newcommand{\tikzxmark}{%
\tikz[scale=0.23] {
    \draw[line width=0.7,line cap=round] (0,0) to [bend left=6] (1,1);
    \draw[line width=0.7,line cap=round] (0.2,0.95) to [bend right=3] (0.8,0.05);
}}
\usepackage{microtype}
\usepackage{makecell}
\usepackage[symbol]{footmisc}

\begin{document}

\title{
Blind to Position, Biased in Language: \protect\\
Probing Mid-Layer Representational Bias in
Vision-Language Encoders for Zero-Shot
Language-Grounded Spatial Understanding
}

\titlerunning{Blind to Position, Biased in Language}

\author{
Na Min An\inst{1}\thanks{Equal contribution. Contact: \email{naminan@kaist.ac.kr}, \email{rkswlsj13@kaist.ac.kr}} \and
Inha Kang\inst{1}$^{\textcolor{gray}{*}}$ \and
Minhyun Lee\inst{2} \and  
Hyunjung Shim\inst{1}\thanks{Corresponding author.}
}

\authorrunning{An et al.}

\institute{
$^{1}$Korea Advanced Institute of Science and Technology (KAIST) \\
$^{2}$Samsung Electronics \\
}


\maketitle

\begin{abstract}
Vision–Language Encoders (VLEs) are widely adopted as the backbone of zero-shot referring image segmentation (RIS), enabling text-guided localization without task-specific training. However, prior works underexplored the underlying biases within mid-layer representations that preserve positional and language-specific information.
Through layer-wise investigation, we reveal that the conventionally used final-layer multimodal embeddings prioritize global semantic alignment, leading to two coupled consequences. First, vision embeddings exhibit weak sensitivity to positional cues. Second, multilingual text embeddings form language-dependent geometric shifts within the shared space.
Motivated by these findings, we identify an underexplored pathway within VLE mid-layers to construct a spatial map, applicable for improving zero-shot RIS by 1–7 mIoU on nine RefCOCO benchmarks.
Furthermore, leveraging mixed-language mid-layer embeddings yields enhanced spatial grounding accuracy (+7–8 mIoU and IoU@50), albeit with increased inference cost, and also improves performance on the zero-shot text-to-image retrieval task.
Our work opens up the discussion about the effects of effective representational bias probing of VLEs for enhanced spatial grounding.
\end{abstract}
\section{Introduction}
\label{intro}

Language-grounded spatial understanding is a fundamental capability for embodied AI and interactive vision.
Tasks such as language-conditioned navigation and manipulation require agents to map free-form instructions to precise regions and actions within cluttered scenes~\cite{anderson2018visionandlanguagenavigationinterpretingvisuallygrounded, chen2020touchdownnaturallanguagenavigation, qi2020stanza, shridhar2020alfredbenchmarkinterpretinggrounded, shridhar2021cliportpathwaysroboticmanipulation}.
In real deployments, the same intent may be expressed through diverse phrasings and across multiple languages. This motivates multilingual grounding benchmarks that systematically extend English datasets via translation, reaching a million-expression scale~\cite{ku2020roomacrossroommultilingualvisionandlanguagenavigation,nogueira2025comprehensionmultilingual}.
Despite this progress, recent analyses still reveal persistent weaknesses in fine-grained spatial reasoning~\cite{kamath2023whats, tong2024eyes} as well as language-dependent representational biases~\cite{zhang2024comfort}.

These two issues stem from a shared architectural bottleneck: Most systems rely on the final-layer global embedding of pretrained Vision-Language Encoders (VLEs) for cross-modal scoring. Common multimodal learning approaches~\cite{radford2021learning, tschannen2025siglip} concentrate alignment on this single representation, causing diverse visual and linguistic signals to collapse into a compact semantic space. As deeper layers increasingly optimize for high-level abstraction, they progressively discard fine-grained positional structure while introducing language-dependent geometric shifts, likely influenced by the training imbalance and tokenization asymmetry. Crucially, precise spatial cues and language-stable representations remain encoded within the \emph{intermediate} layers, prior to the final abstraction bottleneck.

The limitation becomes particularly evident in referring image segmentation (RIS), which evaluates whether a VLE can ground language to spatially precise image regions.
RIS therefore provides a natural testbed for diagnosing these two biases, as the task isolates the core cross-modal challenge of mapping language descriptions to pixel-level masks. In particular, \emph{zero-shot} RIS strongly penalizes spatially insensitive or cross-lingually unstable scoring interfaces, especially under translation-based multilingual benchmarks where the image-mask pair remains fixed while only the query language varies.

\begin{figure}[t!]
    \centering
    \begin{subfigure}[t]{\textwidth}
        \centering
        \vspace{0pt}
        \includegraphics[width=\textwidth]{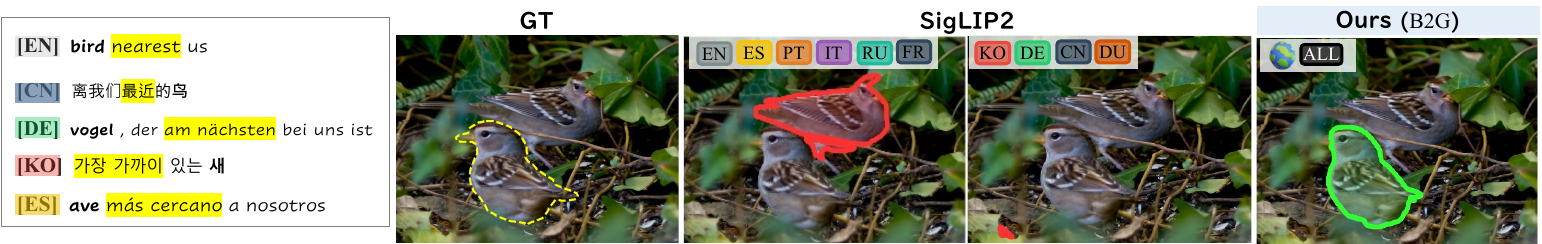}
        \captionsetup{skip=4pt}
        \caption{Qualitative comparison of zero-shot RIS on multilingual and spatial queries}
        \label{fig:quail_ex_intro}
    \end{subfigure}
    
    \begin{subfigure}[t]{0.455\textwidth}
        \centering
        \vspace{0pt}
        \includegraphics[width=\textwidth]{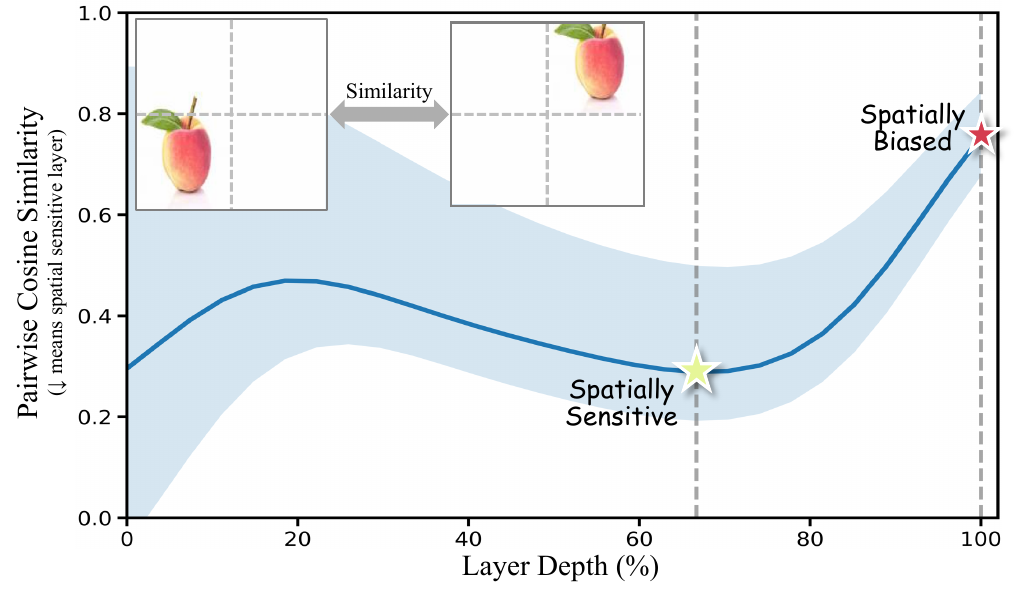}
        \captionsetup{skip=0pt}
        \caption{Spatial probe}
        \label{fig:intro_spatial}
    \end{subfigure}
    \begin{subfigure}[t]{0.535\textwidth}
        \centering
        \vspace{0pt}
        \includegraphics[width=\textwidth]{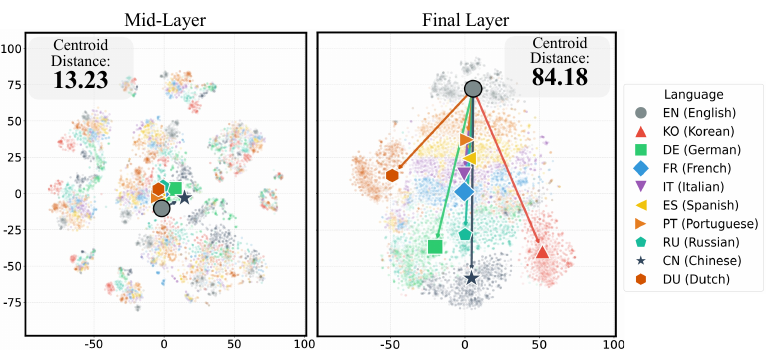}
        \captionsetup{skip=7pt}
        \caption{Multilingual t-SNE probe}
        \label{fig:intro_multilingual}
    \end{subfigure}
    \caption{\textbf{Task and preliminary diagnosis.} \textbf{(a)} Our proposed \ours accurately localizes referred regions in multilingual zero-shot RIS. \textbf{(b)} A tile-permutation probe reveals intermediate layers of vision encoders are spatially sensitive, while the final embedding is nearly position-invariant. \textbf{(c)} Semantically equivalent multilingual queries exhibit systematic geometric shifts in the latent space for the final layer than the middle layer.}
    \label{fig:teaser}
    \vspace{-1em}
\end{figure}

Pretrained VLEs~\cite{radford2021learning, zhai2023sigmoid, fang2023data, li2022blip, tschannen2025siglip} have now become standard backbones for zero-shot RIS, enabling mask grounding without any dataset-specific fine-tuning~\cite{subramanian2022reclip, yu2023zero, suo-etal-2023-text, wang2025iterprime, liu2025hybrid, ni2023ref, han2024zero, luddecke2022image} (\cref{fig:quail_ex_intro}).
Most training-free pipelines follow a proposal-and-rank design:
an off-the-shelf segmentor generates candidate masks~\cite{wang2022freesolo, cheng2022masked, liang2023open, kirillov2023segment}, and a frozen VLE ranks them using image--text similarity.
Since modern segmentors often produce multiple overlapping or plausible masks, successful zero-shot RIS critically depends on whether the VLE's ability to disambiguate the query and localize the correct region.
When the ranking signal is dominated by global semantics, models frequently exhibit ``opposite visualization''~\cite{li2023clip} (later in \cref{fig:analysis_layer}), selecting background context instead of the intended foreground target. This behavior highlights that the final-layer embedding is a suboptimal interface for pixel-level grounding.
Yet existing training-free improvements mainly address spatial disambiguation under English queries, while multilingual robustness typically relies on training new encoders—leaving no unified training-free solution that simultaneously mitigates both spatial and language biases on frozen VLEs.

Our analysis pinpoints where the missing cues remain accessible within the model.
To identify position-sensitive visual representations, we introduce a spatial equivariance probe based on controlled image tile permutation. Intermediate layers become increasingly sensitive to spatial arrangement, whereas the final embedding is nearly position-invariant (\cref{fig:intro_spatial}).
We observe an analogous phenomenon in multilingual text embeddings.
Even for semantically identical sentences, the final layers exhibit systematic language-dependent shifts (\cref{fig:intro_multilingual}). While these shifts have little impact on text-to-text similarity, they significantly distort cross-modal similarity with visual features.
Together, these findings reveal a unified opportunity: Spatial and multilingual cues remain encoded within pretrained VLEs but are attenuated at the final alignment interface, and can be recovered by selectively leveraging intermediate representations at test time.

Building on this insight, we propose \textbf{B}iased \textbf{to} \textbf{G}rounded (\ours), a training-free framework following a unified \emph{probe $\rightarrow$ select $\rightarrow$ ground} paradigm.
Rather than replacing existing pipelines, \ours acts as a plug-in that strengthens the mask-ranking signal of proposal-based zero-shot RIS methods~\cite{yu2023zero, liu2025hybrid}.
For spatial grounding, \ours extracts features from a position-sensitive intermediate vision layer to construct a dense patch-level similarity map (\texttt{P-Map}), whose high-response regions form spatial clusters that guide mask reranking.
To improve multilingual robustness, \ours probes the text encoder across $N$ translations to identify a language-stable layer and compute a mid-layer multilingual centroid, which is injected into the anchor query to stabilize representation geometry before final contextualization.
The framework operates entirely on frozen VLEs, supporting both fixed (\oursfix) and automatic (\oursdynclus, \oursdynperm) layer selection, without fine-tuning, gradient backpropagation~\cite{selvaraju2016grad}, or recomputation~\cite{bousselham2024grounding,li2023clip}.

\ours consistently improves zero-shot RIS across standard benchmarks\cite{nagaraja2016modeling, mao2016generation, wu2020phrasecut}, surpassing the strongest prior training-free baseline by 2.15 mIoU on RefCOCO, 5.47 mIoU on RefCOCO+, and 5.75 mIoU on RefCOCOg (CLIP ViT-B/16)~\cite{liu2025hybrid}. Ours also substantially improves multilingual zero-shot RIS, increasing average IoU@50 from 45.13 to 53.41 and mIoU from 42.42 to 49.62 across nine translated benchmarks~\cite{nogueira2025comprehensionmultilingual}.
Beyond RIS, our mid-layer multilingual stabilization also improves robustness in zero-shot multilingual text-to-image retrieval (+4.22, +2.77, +6.42 Recall@1 on RefCOCO, RefCOCO+, RefCOCOg). These results suggest that spatial and multilingual biases largely stem from the final-layer abstraction interface rather than the absence of grounding information itself. By probing intermediate representations, \ours unlocks spatially precise and language-robust grounding from frozen VLEs.
\section{Methodology}\label{method}

\begin{figure}[t]
    \centering
    \resizebox{\linewidth}{!}{
    \includegraphics{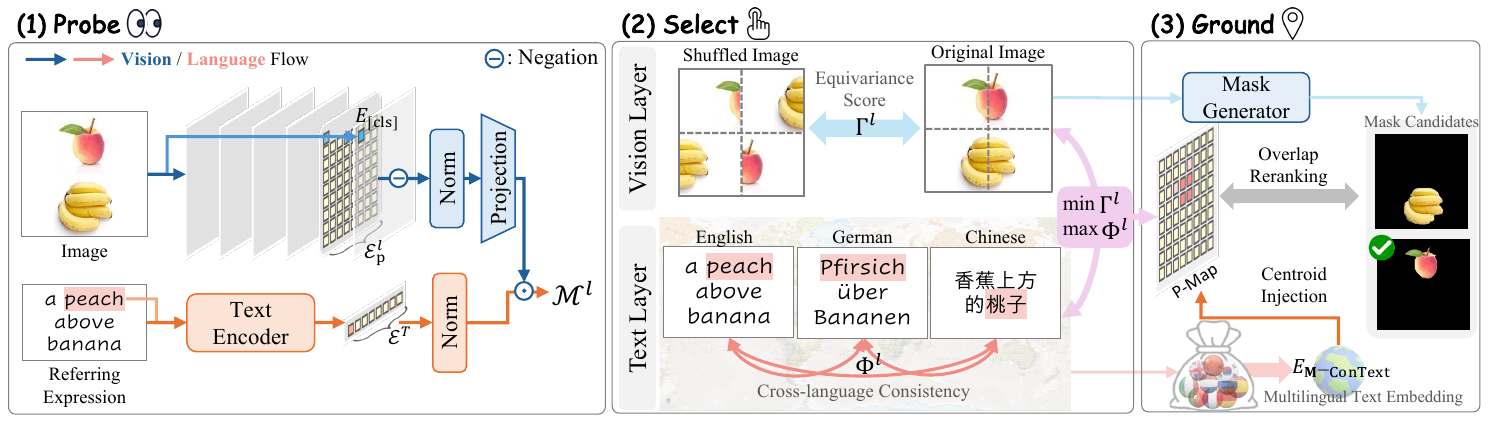}
    }
    \vspace{-1.5em}
    \caption{\textbf{Overview of Proposed Framework.} 
    \ours follows a unified \emph{\textbf{probe} $\rightarrow$ \textbf{select} $\rightarrow$ \textbf{ground}} paradigm.
    It \emph{probes} mid-layer VLE representations to identify 
    spatially sensitive and language-stable embeddings, 
    \emph{selects} the most informative layers, 
    and performs mask \emph{grounding} through spatial map, \texttt{P-Map}, with multilingual centroid fusion.}
    \label{fig:method}
\end{figure}

\paragraph{\textbf{\emph{Problem setup.}}}
In zero-shot RIS, we are given an image $I$, a referring expression $T$, and $K$ mask proposals $\{M_k\}_{k=1}^{K}$ from an off-the-shelf segmentor.
Let $\mathcal{R}(I,M_k)$ denote the region extraction operator used by prior training-free pipelines.
Following standard practice, we obtain a region embedding (of masked or cropped images specified by $M_k$) by applying the frozen vision encoder to $\mathcal{R}(I, M_k)$ and pooling its final-layer representation.
We denote by $E_{\mathrm{txt}}(T)$ the text embedding used for scoring (\cref{sec:method2} replaces it with a multilingual-stabilized variant).
A frozen VLE then scores each proposal by global image--text similarity:
\begin{equation}
S_k = \cos\!\Big(E_{\mathrm{vis}}(\mathcal{R}(I,M_k)),\; E_{\mathrm{txt}}(T)\Big),
\qquad
\hat{M}=\arg\max_{k} S_k .
\label{eq:baseline_score}
\end{equation}
\ours keeps the proposal generator and $\mathcal{R}(\cdot)$ unchanged, and focuses on strengthening the grounding signal by selecting more informative internal representations and adding a lightweight reranking cue on top of the baseline score.

\paragraph{\textbf{\emph{Proposed framework.}}}
Our key premise is that the final-layer alignment interface of pretrained VLEs is optimized for global semantic matching, which can attenuate the structural cues required for precise grounding and further amplify the English-dominant alignment bias introduced during pretraining.
This manifests as (1) weakened positional sensitivity in vision features and (2) language-dependent geometric shifts in multilingual text features (\cref{fig:teaser}\subref{fig:intro_spatial}--\subref{fig:intro_multilingual}).
\ours therefore follows a unified \emph{probe $\rightarrow$ select $\rightarrow$ ground} paradigm (\cref{fig:method}): (1) \emph{probe} mid-layer representations to identify structure-preserving layers. (2) \emph{select} a position-sensitive vision layer and a language-stable text layer. (3) perform a mask \emph{ground}ing using patch-level similarity and multilingual centroid injection.

\vspace{0.5em}

\begin{figure}[t]
    \centering
    
    \begin{subfigure}[t]{0.35\linewidth}
        \centering
        \includegraphics[width=\linewidth]{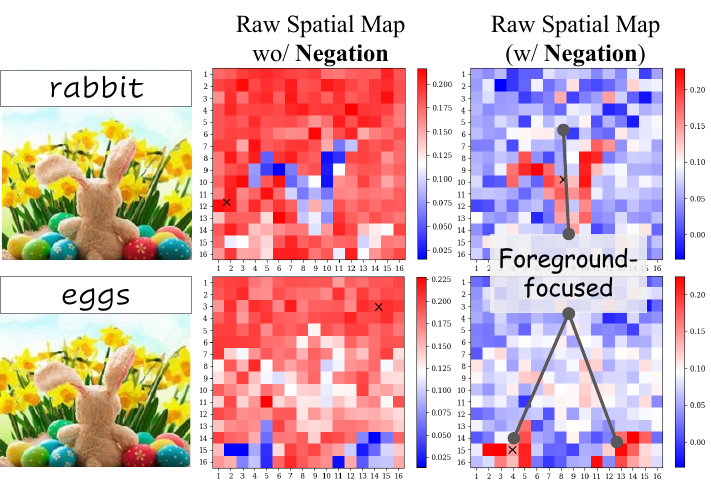}
        \caption{Raw patch-level spatial map}
        \label{fig:pmap_raw}
    \end{subfigure}
    \hfill
    \begin{subfigure}[t]{0.64\linewidth}
        \centering
        \includegraphics[width=\linewidth]{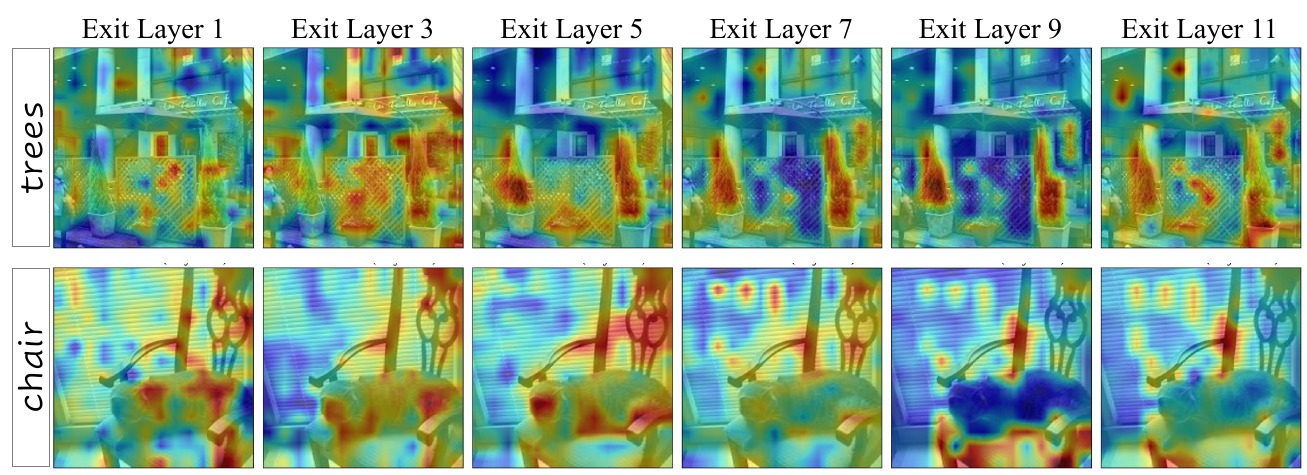}
        \caption{Layer-wise final spatial map (\texttt{P-Map})}
        \label{fig:pmap_final}
    \end{subfigure}
    \vspace{-0.5em}
    \caption{\textbf{Spatial Map Visualization.} \textbf{(a)} Raw spatial map with and without negation operation. \textbf{(b)} Final spatial maps generated across varying exit layers.}
    \label{fig:pmap}
    \vspace{-0.5em}
\end{figure}

\subsection{\textbf{B}iased \textbf{to} \textbf{G}rounded (\ours) for (Non-Multilingual) VLE}
\label{sec:method1}

\renewcommand{\thefootnote}{\arabic{footnote}}
\paragraph{\textbf{\emph{Stage 1: Spatial bias \emph{probing}}.}}
The central challenge for spatial grounding at intermediate layers is that patch-level embeddings tend to suppress foreground activations in favor of global context, pushing target-region similarities into the negative domain~\cite{li2023clip}.
Our approach addresses this via two steps. We first construct a raw patch--text similarity map to localize spatial responses. We then apply a deterministic sign inversion to correct the foreground suppression.
At several mid-layer exits, the referred foreground patches become systematically anti-correlated with the query direction in the joint embedding space, leading to inverted cosine responses  (\cref{fig:pmap_raw}).
Since $\cos(-u,v)=-\cos(u,v)$, a deterministic sign flip converts this inverted activation into a standard high means relevant map. We apply the flip to patch features before the shared projection\footnote{For models that do not have an image-to-text projection layer, such as SigLIPs, we use a post-layer normalization (LN) layer instead.}, which is empirically
stable across VLE backbones.

Specifically, given an input image $I$, we extract patch-level embeddings from the $l$–\emph{th} exit layer of the visual encoder, $\mathcal{E}_{\mathrm{p}}^{l}\in\mathbb{R}^{p\times p\times d^{*}}$.
To compare them with text in the joint embedding space, we use the model's post LN (per block) and project them using the pretrained projection matrix $\mathbb{W}_{d^{*}}^{d}$:
\[
E_{\mathrm{p}}^{l}
=
\mathrm{LN}(\mathcal{E}_{\mathrm{p}}^{l})
\cdot
\mathbb{W}_{d^{*}}^{d}
\in
\mathbb{R}^{p \times p \times d}.
\]
For the text embedding, we use $E_{\mathrm{ConText}}$, our hybrid text representation that fuses global sentence-level features with context-augmented local noun features (\texttt{CT}, Appendix for details) for scoring; \cref{sec:method2} replaces it with the
multilingual-stabilized $E_{\mathbf{M}\text{-}\mathrm{ConText}}$. We compute a raw patch-level similarity map as follows:
\[
\widetilde{\mathcal{M}}^{l}
=
(\widetilde{\mathcal{M}}^{l}_{ij})_{1 \le i,j \le p}
\in
\mathbb{R}^{p \times p},
\quad
\widetilde{\mathcal{M}}^{l}_{ij}
=
\cos\!\left(E_{\mathrm{p}}^{l}[i,j],\, E_{\mathrm{ConText}}\right).
\]
However, direct cosine similarity at mid layers may exhibit the known ``opposite visualization'' phenomenon~\cite{li2023clip}, where foreground patches receive relatively low responses. 
To deterministically correct this effect, we negate the patch embeddings before normalization, yielding our raw spatial map (later interpolated to yield the final spatial map, \texttt{P-Map}), $\mathcal{M}^{l}\in\mathbb{R}^{p\times p}$:
\[
\widetilde{E}_{\mathrm{p}}^{l}[i,j]
=
\mathrm{LN}\!\left(-\mathcal{E}_{\mathrm{p}}^{l}[i,j]\right)
\cdot
\mathbb{W}_{d^{*}}^{d},
\qquad
\mathcal{M}^{l}_{ij}
=
\cos\!\left(\widetilde{E}_{\mathrm{p}}^{l}[i,j],\, E_{\mathrm{ConText}}\right).
\]
This sign inversion restores foreground-localized activations and produces more interpretable spatial maps without additional training time and memory costs.

\paragraph{\textbf{\emph{Stage 2: Vision layer \emph{selection}}.}}
Although intermediate layers exhibit varying degrees of spatial sensitivity, not all layers yield a valid spatial map (\cref{fig:pmap}).
We therefore select a position-sensitive layer using a tile-permutation probe.
Given an input image $I$, we construct a spatially permuted variant $I^{\pi}$ by shuffling non-overlapping tiles over the $p \times p$ grid.
This preserves local appearance statistics while disrupting global spatial configuration.

For each candidate layer $l$, we extract patch embeddings from both $I$ and $I^{\pi}$, project them as above, and restore spatial correspondence by applying the inverse permutation $\pi^{-1}$ to the shuffled patch features.
We then compute an equivariance score as follows:
\vspace{-1em}
\[
\hat{E}_{\mathrm{p}}^{l}(I^{\pi})
=
\pi^{-1}\!\left(
E_{\mathrm{p}}^{l}(I^{\pi})
\right),
\qquad
\Gamma^{l}
=
\frac{1}{p^2}
\sum_{i=1}^{p^2}
\cos\!\left(
E_{\mathrm{p}}^{l}(I)[i],\,
\hat{E}_{\mathrm{p}}^{l}(I^{\pi})[i]
\right).
\vspace{-0.5em}
\]
Larger $\Gamma^{l}$ indicates that patch features remain similar after permutation (\ie, weak dependence on spatial arrangement), while smaller values indicate stronger position sensitivity.

To determine the optimal layer ($l^{*}_{\text{vis}}$), we investigate three selection strategies using a calibration set: dynamic selection via a permutation probe (\oursdynperm), cluster coherence (\oursdynclus), and a fixed mid-layer heuristic (\oursfix).
Empirically, all three strategies yield highly comparable grounding performance (details in the Appendix). This consistency suggests that the preservation of spatial structure is a robust, inherent architectural property of VLE mid-layers, rather than a fragile, sample-specific artifact. Consequently, \ours can reliably operate with the minimal cost \oursfix strategy in practice without sacrificing accuracy (detailed formulations are provided in the Appendix).

\paragraph{\textbf{\emph{Stage 3: Mask \emph{grounding} by overlap reranking}.}}
Given \texttt{P-Map} $\mathcal{M}^{l^{*}_{\mathrm{vis}}}\in\mathbb{R}^{p\times p}$, we convert it into a spatial guidance mask.
We normalize $\mathcal{M}^{l^{*}_{\mathrm{vis}}}$ to $[0,1]$, threshold it with $\delta$, and extract connected components on the $p\times p$ grid to obtain clusters $\{C_r\}$.
Each cluster is upsampled to the image resolution, and we merge them into a binary guidance mask $G$.

For each proposal $M_k$, we compute the baseline similarity score $S_k$ defined in \cref{eq:baseline_score}, alongside an overlap score $O_k$. 
We use $O_k = \mathrm{IoU}(M_k, G)$ by default; using the mean \texttt{P-Map} response within $M_k$ yields similar trends (Appendix).
We subsequently rerank the proposals using a combined score:
\begin{equation}
S'_k = S_k + \lambda\, O_k,
\qquad
\hat{M}=\arg\max_{k} S'_k .
\vspace{-0.5em}
\end{equation}
To reduce noise, overlap reranking is applied only to the top-$\kappa$ proposals under $S_k$.
Overall, \ours extracts dense spatial cues from frozen mid-layer features and uses them as a lightweight reranking signal (denoted as \texttt{TC}; Top Candidate selection), requiring neither gradient-based saliency nor recomputation of attention.

\vspace{0.5em}
\ours challenges the prevailing assumption that VLE is spatially blind by revealing latent spatial knowledge from its intermediate representations. Unlike prior methods that rely on computationally intensive gradient-based saliency or attention maps, \ours derives spatial guidance from a simple yet effective spatial map, \texttt{P-Map}. This training-free extraction avoids recomputation and remains faithful to the model's intrinsic spatial structure.

Our spatial probe demonstrates that VLE mid-layers preserve position-sensitive structures that are largely attenuated at the final layer. Interestingly, a similar phenomenon emerges in the text encoder, where intermediate text layers maintain a more language-consistent embedding geometry compared to the final representation. This structural parallel motivates our unified framework, which selects the most informative mid-layers from both modalities as detailed in the subsequent section.


\subsection{\textbf{B}iased \textbf{to} \textbf{G}rounded (\ours) for Multilingual VLE}
\label{sec:method2}

\paragraph{\textbf{\emph{Stages 1--2: Multilingual bias \emph{probing} and layer \emph{selection}}.}}
When the same referring expression is translated into multiple languages, the resulting text embeddings should ideally lie in the same region of the shared multimodal space. In practice, however, language-specific surface variations, such as morphology, word order, and tokenization, introduce systematic geometric shifts that accumulate across transformer layers.

Given an input expression $T$, we first generate semantically aligned translations $\{T^{(1)},\dots,T^{(N)}\}$ across $N$ languages, designating the initial expression $T^{(1)}=T$ as the anchor language.
We then process these expressions through the text encoder up to layer $l$. 
From this layer, we extract a pooled sentence representation $h^{l}_{(n)}\in\mathbb{R}^{d}$ utilizing the hidden state of the end-of-text token.
Subsequently, we quantify the cross-language consistency by
\[
\Phi^{l}=
\frac{1}{\binom{N}{2}}
\sum_{a<b}
\cos\!\left(h^{l}_{(a)},h^{l}_{(b)}\right).
\]
Larger $\Phi^{l}$ indicates a more language-stable geometry at layer $l$.
We select the most consistent layer by
$
l^{*}_{\mathrm{txt}}=\arg\max_{l}\Phi^{l}.
$

\paragraph{\textbf{\emph{Stage 3: Multilingual centroid for \emph{grounding}}.}}
To stabilize the language-dependent geometry, we form a multilingual centroid at the selected intermediate layer $l^{*}_{\mathrm{txt}}$.
Given translations $\{T^{(n)}\}_{n=1}^{N}$, we first obtain their token-level embeddings and propagate them through the text encoder to extract pooled mid-layer embeddings $h^{l^{*}_{\mathrm{txt}}}_{(n)}$ and compute their average $\bar{h}^{\,l^{*}_{\mathrm{txt}}}$ across languages.
We then inject this centroid into the anchor-language hidden state at layer $l^{*}_{\mathrm{txt}}$ by replacing the end-of-text (EOT) token representation.
Let $\mathcal{H}^{l^{*}_{\mathrm{txt}}}_{(n)}$ denote the token-wise hidden states of $T^{(n)}$ at layer $l^{*}_{\mathrm{txt}}$ and $eot$ its EOT index.
We set
$
\mathcal{H}^{l^{*}_{\mathrm{txt}}}_{(n)}[eot] \leftarrow \bar{h}^{\,l^{*}_{\mathrm{txt}}}
$
and also integrate the centroid of the global sentence context and local noun-phrase context.
$
\mathcal{E}_{\mathbf{M}\text{-}\mathrm{ConText}} = \gamma \mathcal{E}_{\mathbf{M}\text{-}\mathrm{Sen}} + (1-\gamma) \mathcal{E}_{\mathbf{M}\text{-}\mathrm{NounCtx}}.
$
Then, this merged embedding is propagated through the subsequent layers and the final projection head to obtain the final embedding $E_{\mathbf{M}\text{-}\mathrm{ConText}}$.
By performing both multilingual and contextual fusion at an intermediate layer, the model preserves language-invariant semantic structure before later layers amplify language-specific variations, producing a stable language-agnostic feature while keeping the text encoder frozen.

\paragraph{Test-time usage and mask scoring.}
During inference, \ours pipeline operates without gradient updates. 
First, the text query $E_{\mathbf{M}\text{-}\mathrm{ConText}}$ is computed strictly once, restricting the multilingual overhead to $N$ early-exit partial passes and a single continuation pass. 
Second, for an image, a single forward pass through the frozen vision encoder extracts mid-layer patch embeddings. These are combined with $E_{\mathbf{M}\text{-}\mathrm{ConText}}$ to construct the \texttt{P-Map} and generate the spatial guidance mask $G$. 
Finally, we evaluate the similarity $S_k$ (\cref{eq:baseline_score}) between $E_{\mathbf{M}\text{-}\mathrm{ConText}}$ and all $\kappa$ candidates. To resolve spatial ambiguities, we rerank the top-$\kappa$ candidates by combining $S_k$ with their overlap $O_k$ against $G$ to produce the final mask $\hat{M}$.

\section{Experiments}\label{exp}

\subsection{Experimental Setup}\label{exp:setup}

\paragraph{\textbf{\emph{Tasks and evaluation protocols.}}}
We evaluate B2G under three complementary protocols: standard English zero-shot RIS, multilingual zero-shot RIS, and zero-shot multilingual text-to-image retrieval. Unless otherwise noted, \ours is used as a training-free plug-in on frozen VLEs under the same pipeline.

\paragraph{\textbf{\emph{Evaluation datasets and metrics.}}}
Following prior zero-shot RIS works~\cite{yu2023zero, suo-etal-2023-text, ni2023ref, sun2024clip, wang2025iterprime, liu2025hybrid}, we report standard RIS results on RefCOCO~\cite{nagaraja2016modeling}, RefCOCO+~\cite{nagaraja2016modeling}, and RefCOCOg~\cite{mao2016generation}, which differ in linguistic complexity and positional cues--RefCOCO contains more explicit spatial terms, while RefCOCOg features longer expressions. We also report results on PhraseCut~\cite{wu2020phrasecut} under \emph{all} and \emph{unseen} (COCO class~\cite{lin2014microsoft}) settings. For RIS, the main metrics are mean Intersection over Union (mIoU) (if unspecified), overall IoU (oIoU), and IoU@50, which measures the percentage of predictions whose IoU with the ground-truth mask exceeds 0.5.

\paragraph{\textbf{\emph{Implementation details.}}}
To ensure a fair comparison with previous zero-shot RIS methods, we use the same frozen VLE backbones as the compared baselines, including CLIP (ViT-B/32 and ViT-B/16)~\cite{radford2021learning}, as well as BLIP (Appendix)~\cite{li2022blip}, SigLIP~\cite{zhai2023sigmoid}, DFN~\cite{fang2023data}, and SigLIP2~\cite{tschannen2025siglip}. Since \oursfix uses a fixed exit layer, we determine the optimal exit layer ($l^{*}$) of the visual encoder and the initial threshold ($\delta$) on a small unlabeled calibration split comprising 10\% of the RefCOCOg val (U) set, and keep them fixed at test time.

\subsection{Main Results on Standard Zero-shot RIS}

\begin{table*}[t!]
\centering
\vspace{-1em}
\caption{\textbf{Comparison of Zero-shot RIS Performance.} Ours consistently achieves superior performance over existing feature extraction approaches, Global-Local~\cite{yu2023zero} and HybridGL~\cite{liu2025hybrid}, across different backbone encoders. For fair comparison, all the listed methods have been applied with \texttt{CT} and \texttt{SG} (ablation in later). The baseline in \colorbox{lightred}{red} and statistically significant improvements over the baseline in \colorbox{lightblue}{blue} for each backbone.}
\vspace{-0.7em}
\label{tab:main_ris}
\renewcommand{\arraystretch}{0.9} 
\resizebox{0.88\linewidth}{!}{%
\footnotesize 
\begin{tabular}{@{} l c *{9}{>{\scriptsize}c} >{\scriptsize}l }
\toprule
\multicolumn{1}{c}{\multirow{2}{*}{\textbf{Method}}} & {\scriptsize \textbf{Pre-trained}} &
\multicolumn{3}{c}{\textbf{RefCOCOg}} &
\multicolumn{3}{c}{\textbf{RefCOCO}} &
\multicolumn{3}{c}{\textbf{RefCOCO+}} &
\multicolumn{1}{l}{\multirow{2}{*}{\textbf{Avg.}}} \\ 

& {\scriptsize \textbf{Segmentor}} & 
\multicolumn{1}{c}{val} & \multicolumn{1}{c}{test} & \multicolumn{1}{c}{valG} & 
\multicolumn{1}{c}{val} & \multicolumn{1}{c}{testA} & \multicolumn{1}{c}{testB} & 
\multicolumn{1}{c}{val} & \multicolumn{1}{c}{testA} & \multicolumn{1}{c}{testB} &  \\

\cmidrule(lr){1-1}\cmidrule(lr){2-2}\cmidrule(lr){3-5}
\cmidrule(lr){6-8}\cmidrule(lr){9-11}
\cmidrule(lr){12-12}

\rowcolor{gray!15}
\multicolumn{12}{c}{\textbf{CLIP ViT-B/32}} \\

Global-Local & \multirow{3}{*}{SAM}
& 44.71 & 45.97 & 45.75 & 35.99 & 35.36 & 36.15 & 32.50 & 33.62 & 30.56 & 37.85 \\

HybridGL & & 46.54 & 47.00 & 47.34 & 40.67 & 41.84 & 39.05 & 37.52 & 40.73 & 33.40 & \cellcolor{lightred}{41.57} \\
\arrayrulecolor{gray!65}
\cmidrule(lr){1-1}\cmidrule(lr){3-5}\cmidrule(lr){6-8}
\cmidrule(lr){9-11}\cmidrule(lr){12-12}
\arrayrulecolor{black}
\hspace{0.5em}$+\ours$ & & 46.89 & 47.20 & 47.67 & 41.07 & 42.15 & 39.41 & 38.54 & 42.12 & 33.89 & \cellcolor{lightblue}{42.55} \texttt{\tiny{(+0.98)}} \\

\arrayrulecolor{gray!65}
\midrule
\arrayrulecolor{black}

Global-Local & \multirow{5}{*}{{\shortstack{Mask2\\Former}}}
& 47.44 & 46.78 & 47.68 & 43.19 & 46.81 & 39.32 & 39.47 & 44.74 & 34.19 & 43.29 \\

HybridGL &
& 48.83 & 48.35 & 49.26 & 41.72 & 44.46 & 37.64 & 40.91 & 47.11 & 33.28 & \cellcolor{lightred}{43.51} \\

\arrayrulecolor{gray!65}
\cmidrule(lr){1-1}\cmidrule(lr){3-5}\cmidrule(lr){6-8}
\cmidrule(lr){9-11}\cmidrule(lr){12-12}
\arrayrulecolor{black}

\hspace{0.5em}$+\oursfix$ & 
& 53.47 & 53.84 & 53.88 & 48.58 & 54.39 & 42.01 & 46.34 & 53.60 & 36.80 & \cellcolor{lightblue}{49.21} \texttt{\tiny{(+5.70)}} \\

\hspace{0.5em}$+\oursdynclus$ & 
& 53.56 & 53.62 & 53.83 & 48.62 & 54.37 & 41.57 & 45.73 & 52.57 & 36.36 & \cellcolor{lightblue}{48.91} \texttt{\tiny{(+5.40)}} \\ 

\hspace{0.5em}$+\oursdynperm$ &
& 53.42 & 53.76 & 54.07 & 48.55 & 54.22 & 41.68 & 46.07 & 53.22 & 36.28 & \cellcolor{lightblue}{49.03} \texttt{\tiny{(+5.52)}} \\ 

\midrule
\rowcolor{gray!15}
\multicolumn{12}{c}{\textbf{CLIP ViT-B/16}} \\

Global-Local & \multirow{3}{*}{SAM}
& 48.54 & 49.49 & 48.82 & 40.06 & 40.00 & 40.36 & 35.15 & 37.33 & 33.35 & 41.90 \\

\arrayrulecolor{gray!65}
\cmidrule(lr){1-1}\cmidrule(lr){3-5}\cmidrule(lr){6-8}
\cmidrule(lr){9-11}\cmidrule(lr){12-12}
\arrayrulecolor{black}

HybridGL & & 50.37 & 50.52 & 50.41 & 44.74 & 46.48 & 43.26 & 40.17 & 44.44 & 36.19 & \cellcolor{lightred}{45.18} \\

\hspace{0.5em}$+\ours$ & & 50.75 & 50.74 & 50.77 & 45.17 & 46.82 & 43.65 & 41.28 & 45.94 & 36.72 & \cellcolor{lightblue}{46.24} \texttt{\tiny{(+1.06)}} \\

\arrayrulecolor{gray!65}
\midrule
\arrayrulecolor{black}

HybridGL & \multirow{5}{*}{{\shortstack{Mask2\\Former}}}
& 49.53 & 50.05 & 49.92 & 45.36 & 49.44 & 40.78 & 41.81 & 47.60 & 35.34 & 45.54 \\

Global-Local &
& 49.29 & 49.26 & 48.96 & 47.38 & 53.11 & 41.49 & 40.86 & 45.96 & 35.34 & \cellcolor{lightred}{45.74} \\
\arrayrulecolor{gray!65}
\cmidrule(lr){1-1}\cmidrule(lr){3-5}\cmidrule(lr){6-8}
\cmidrule(lr){9-11}\cmidrule(lr){12-12}
\arrayrulecolor{black}

\hspace{0.5em}$+\oursfix$ & 
& 54.42 & 55.62 & 54.71 & 50.05 & 55.38 & 43.01 & 46.83 & 54.08 & 37.67 & \cellcolor{lightblue}{50.20} \texttt{\tiny{(+4.46)}} \\

\hspace{0.5em}$+\oursdynclus$ & 
& 54.37 & 55.23 & 54.58 & 49.97 & 55.04 & 43.48 & 46.21 & 52.50 & 37.61 & \cellcolor{lightblue}{49.89} \texttt{\tiny{(+4.15)}} \\ 

\hspace{0.5em}$+\oursdynperm$ & 
& 54.57 & 55.45 & 54.63 & 50.18 & 55.47 & 43.71 & 46.64 & 52.89 & 37.63 & \cellcolor{lightblue}{50.13} \texttt{\tiny{(+4.39)}} \\ 

\midrule
\rowcolor{gray!15}
\multicolumn{12}{c}{\textbf{SigLIP ViT-B/16}} \\

Global-Local & \multirow{4}{*}{{\shortstack{Mask2\\Former}}}
& 49.67 & 49.33 & 49.87 & 48.77 & 55.00 & 41.39 & 42.64 & 48.98 & 33.97 & \cellcolor{lightred}{46.62} \\
\arrayrulecolor{gray!65}
\cmidrule(lr){1-1}\cmidrule(lr){3-5}\cmidrule(lr){6-8}
\cmidrule(lr){9-11}\cmidrule(lr){12-12}
\arrayrulecolor{black}

\hspace{0.5em}$+\oursfix$ & 
& 52.35 & 51.80 & 52.01 & 46.05 & 51.34 & 39.49 & 46.10 & 52.66 & 37.47 & \cellcolor{lightblue}{47.70} \texttt{\tiny{(+1.08)}} \\

\hspace{0.5em}$+\oursdynclus$ & 
& 52.43 & 51.90 & 51.97 & 46.67 & 52.42 & 40.78 & 45.73 & 52.22 & 37.74 & \cellcolor{lightblue}{47.98} \texttt{\tiny{(+1.36)}} \\

\hspace{0.5em}$+\oursdynperm$ & 
& 52.12 & 51.43 & 51.78 & 45.56 & 51.29 & 39.72 & 45.95 & 52.36 & 37.66 & \cellcolor{lightblue}{47.54} \texttt{\tiny{(+0.92)}} \\ 

\midrule
\rowcolor{gray!15}
\multicolumn{12}{c}{\textbf{DFN VIT-H/14}} \\

Global-Local & \multirow{4}{*}{{\shortstack{Mask2\\Former}}}
& 48.84 & 49.19 & 48.84 & 45.74 & 52.85 & 38.53 & 39.11 & 46.17 & 32.51 & \cellcolor{lightred}{44.64} \\
\arrayrulecolor{gray!65}
\cmidrule(lr){1-1}\cmidrule(lr){3-5}\cmidrule(lr){6-8}
\cmidrule(lr){9-11}\cmidrule(lr){12-12}
\arrayrulecolor{black}

\hspace{0.5em}$+\oursfix$ & 
& 56.23 & 55.88 & 55.68 & 50.91 & 56.59 & 44.79 & 47.05 & 54.12 & 38.91 & \cellcolor{lightblue}{51.13} \texttt{\tiny{(+6.49)}} \\

\hspace{0.5em}$+\oursdynclus$ & 
& 55.74 & 55.98 & 55.57 & 51.26 & 57.30 & 44.42 & 47.38 & 54.08 & 38.66 & \cellcolor{lightblue}{51.60} \texttt{\tiny{(+6.96)}} \\

\hspace{0.5em}$+\oursdynperm$ & & 55.35 & 55.94 & 55.26 & 51.10 & 57.44  & 43.60 & 47.25 & 54.39 & 38.19 & \cellcolor{lightblue}{50.95} \texttt{\tiny{(+6.31)}} \\ 

\bottomrule

\end{tabular}
}
\vspace{-1.5em}
\end{table*}
\paragraph{\textbf{\emph{Zero-shot RIS.}}}
As shown in \cref{tab:main_ris} and \cref{fig:qualitative} (oIoU and different VLE results in Appendix), \ours consistently yields additional improvements, even when integrated with superior feature extraction methods~\cite{yu2023zero, liu2025hybrid} enhanced by both \texttt{CT} and \texttt{SG} (spatial guider implemented from \cite{liu2025hybrid}) across varying VLE backbones. Particularly, \ours attains noticeable gains for DFN~\cite{fang2023data}, where it achieves +6.49, +6.96, and +6.32 mIoU with \oursfix, \oursdynclus, and \oursdynperm, respectively. Across CLIP ViT-B/32 and CLIP ViT-B/16 (SAM), the performance differences among selection variants are negligible; for readability, we denote them uniformly as \ours. Mask2Former shows better synergy with our method, similar to Wang \etal~\cite{wang2025iterprime}, due to its ability to generate less noisy, semantic masks~\cite{cheng2022masked, liang2023open}. \cref{fig:qualitative_b} indicates that the top candidate masks using our \texttt{P-Map} are highly informative in spatial grounding, as they yield more diverse top-$\kappa$ candidate masks than the previous method.

\begin{figure}[t]
    \centering
    
    \begin{subfigure}[t]{0.47\textwidth}
        \vspace{0pt}
        \centering
        \includegraphics[width=\linewidth]{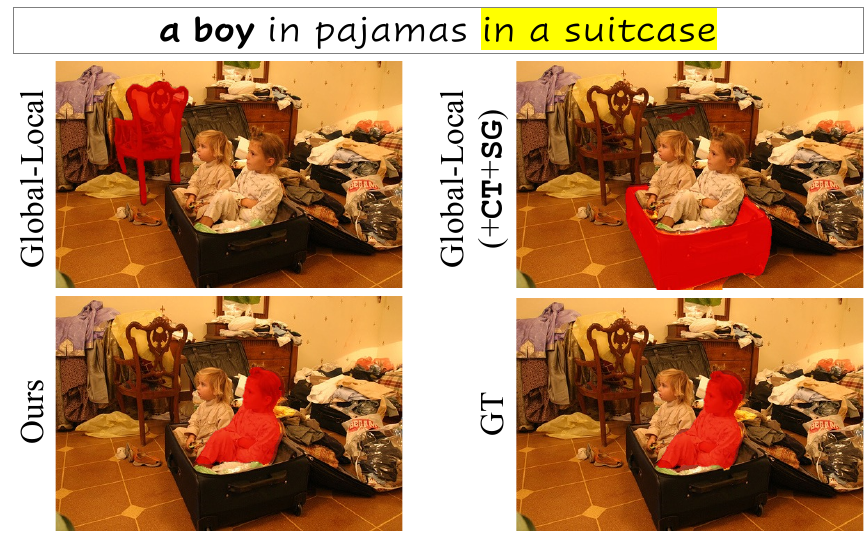}
        \caption{Comparison of Top-1 Selected Masks}
        \label{fig:qualitative_a}
    \end{subfigure}
    \hfill
    \begin{subfigure}[t]{0.52\textwidth}
        \vspace{0pt}
        \centering
        \includegraphics[width=\linewidth]{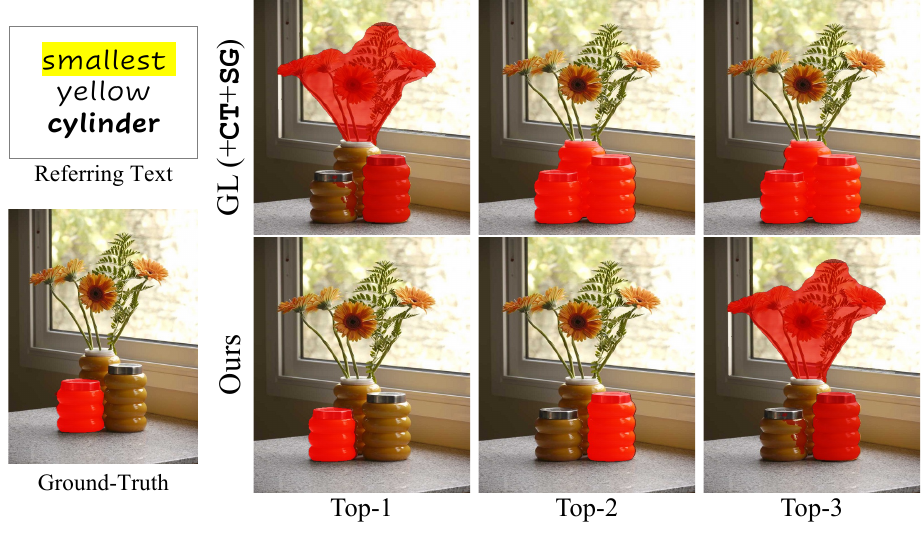}
        \caption{Comparison of Top-$\kappa$ Candidate Masks}
        \label{fig:qualitative_b}
    \end{subfigure}
    \vspace{-0.5em}
    \caption{\textbf{Qualitative Comparison on Zero-shot RIS.} \textbf{(a)} \ours shows strong spatial grounding ability. \textbf{(b)} The masks generated using \texttt{P-Map} are diverse and accurate.}
    \label{fig:qualitative}
    \vspace{-0.5em}
\end{figure}

\subsection{Spatial Grounding Analysis and Challenging Benchmarks}

\begin{figure*}[!t]
    \centering
    
    \begin{subfigure}[!t]{0.24\textwidth}
        \centering
        \vspace{0pt}
        \includegraphics[width=\linewidth]{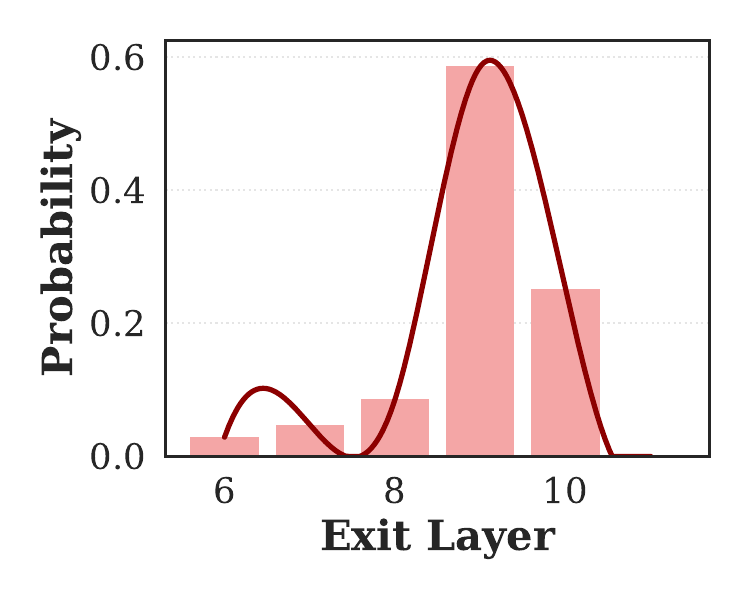}
        \caption{CLIP ViT-B/32}
    \end{subfigure}
    \begin{subfigure}[!t]{0.24\textwidth}
        \centering
        \vspace{0pt}
        \includegraphics[width=\linewidth]{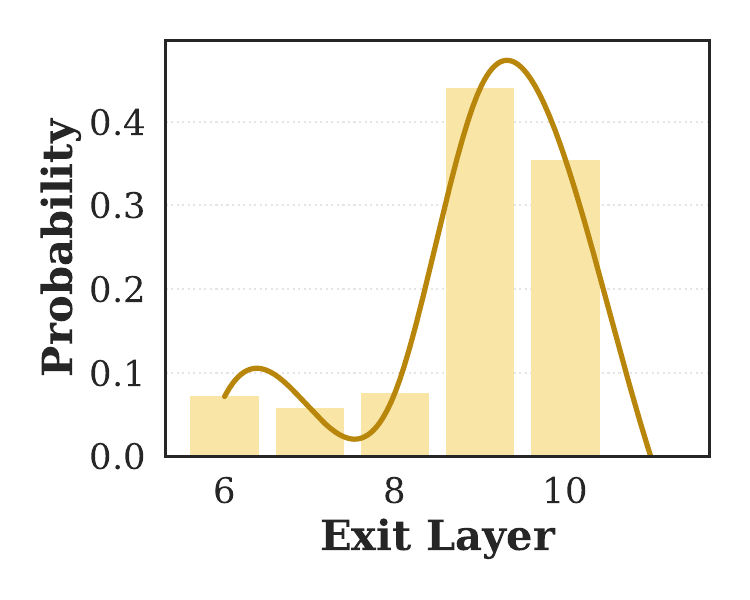}
        \caption{CLIP ViT-B/16}
    \end{subfigure}
    \begin{subfigure}[!t]{0.24\textwidth}
        \centering
        \vspace{0pt}
        \includegraphics[width=\linewidth]{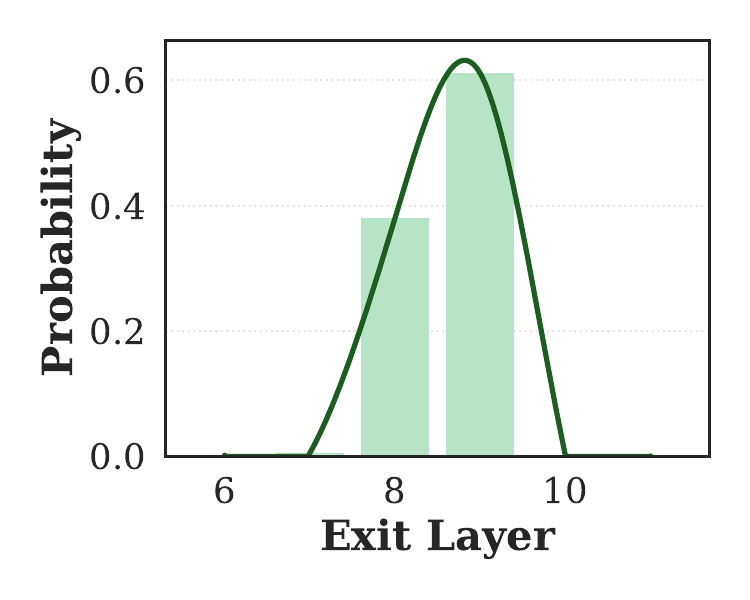}
        \caption{SigLIP ViT-B/16}
    \end{subfigure}
    \begin{subfigure}[!t]{0.24\textwidth}
        \centering
        \vspace{0pt}
        \includegraphics[width=\linewidth]{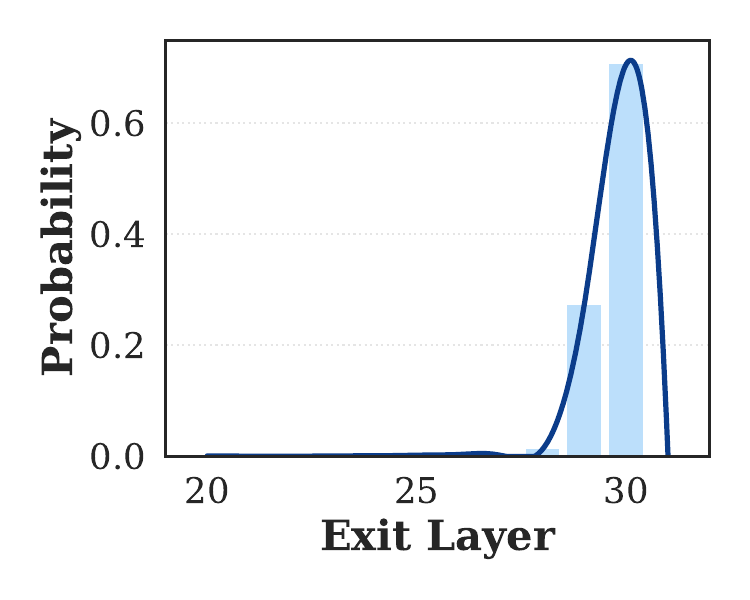}
        \caption{DFN ViT-H/14}
    \end{subfigure}
    \vspace{-0.7em}
    \caption{\textbf{Automatically Selected Layers.}
    We observe a high probability concentration in middle layers automatically selected for \oursdynperm ($n=66$k).}
    \vspace{-1em}
    \label{fig:analysis_layer}
\end{figure*}

\paragraph{\textbf{\emph{Selected intermediate layers.}}}
Different VLEs exhibit distinct layer-wise bias profiles, possibly stemming from variations in their training objectives, projection heads, and tokenization. Consequently, the layer that best preserves spatial equivariance or multilingual consistency is not universal across backbones. This motivates our automatic layer selection strategy: rather than prescribing a fixed layer for each model family, \ours explicitly identifies where structural properties are preserved and adaptively selects the most informative intermediate layer. As shown in \cref{fig:analysis_layer}, the selected layers mostly concentrate in mid-depth regions across diverse backbones. While the optimal layer may vary, this consistent mid-depth concentration supports our claim that intermediate representations preserve grounding-relevant structure, offering a reliable localization signal.

\begin{figure*}[t!]
    \centering
    \begin{subfigure}[t]{0.40\linewidth}
        \vspace{-0.5em}
        \centering
        \includegraphics[width=\linewidth]{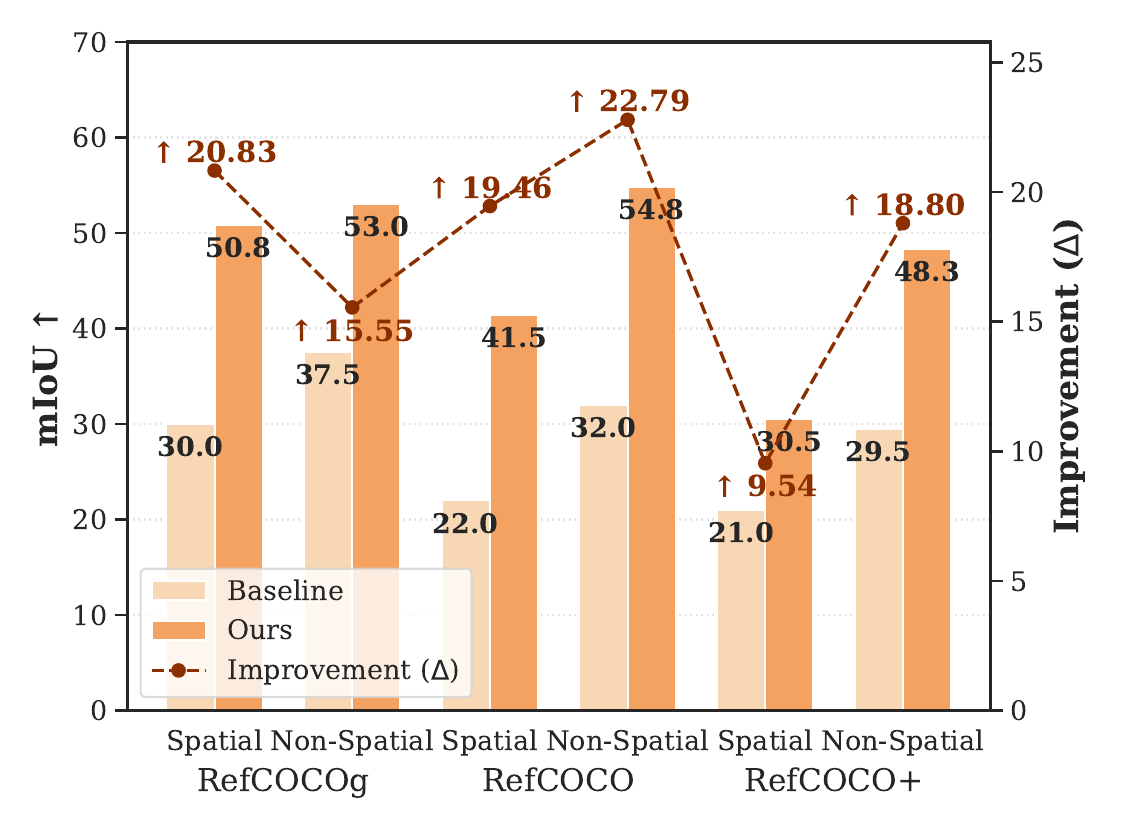}
        \caption{Subset performance}
        \label{fig:subset_perf}
    \end{subfigure}
    \begin{subfigure}[t]{0.32\linewidth}
        \vspace{-0.1em}
        \centering
        \includegraphics[width=\linewidth]{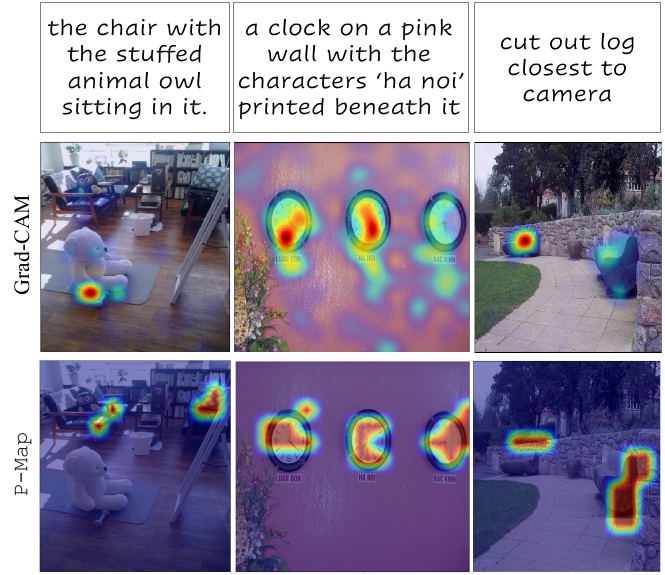}
        \caption{Spatial map comparison}
        \label{fig:spatial_map}
    \end{subfigure}
    \hfill
    \begin{subfigure}[t]{0.25\linewidth}
        \vspace{0pt} 
        \centering
        \setlength{\tabcolsep}{2pt}
        \renewcommand{\arraystretch}{1.53}
        \resizebox{\linewidth}{!}{%
        \begin{tabular}{@{}cccc}
        \toprule
        \textbf{Method} & \textbf{Train?} & \textbf{All} & \textbf{Unseen} \\ 

          \cmidrule(lr){1-1}\cmidrule(lr){2-2}\cmidrule(lr){3-4}
         CRIS~\cite{wang2022cris} & $\checkmark$ & 16.30 & 14.62 \\
         GL$^\dag$~\cite{yu2023zero} & \tikzxmark & 22.43 & 21.54 \\
         TAS~\cite{suo-etal-2023-text} & \tikzxmark & 25.64 & 24.66 \\
         LAVT~\cite{yang2022lavt} & $\checkmark$ & 28.68 & 29.14 \\
         PRIS~\cite{yu2024pseudo} & $\checkmark$ & 32.75 & 33.52 \\
         IRP~\cite{wang2025iterprime} & \tikzxmark & 38.10  & 37.90 \\
         HGL$^\dag$~\cite{liu2025hybrid} & \tikzxmark & 37.60 &  37.91 \\
         
         \rowcolor{gray!15} \ours & \tikzxmark & \textbf{38.53} & \textbf{38.31} \\
        \bottomrule
        \end{tabular}
        }
        \caption{PhraseCut results}
        \label{tab:phrasecut_perf}
    \end{subfigure}
    \vspace{-0.7em}
    \caption{\textbf{Challenging Dataset Performance.} Ours achieves higher and more accurate performance in both spatial and non-spatial subsets and the PhraseCut dataset.}
    \label{fig:analysis_ris}
\end{figure*}

\paragraph{\textbf{\emph{Spatial grounding analysis.}}}
To evaluate the fine-grained spatial grounding ability of VLE (CLIP ViT-B/16), we conduct a targeted evaluation on samples that contain explicit spatial cues (\eg, ``left'', ``closest to''). To identify such samples, we utilize Qwen2.5-14B-Instruct~\cite{yang2024qwen2} to classify each referring expression as spatial or non-spatial and extract the cue if present (prompt details in the Appendix). These automatically generated annotations take up 50.72\%, 62.40\%, and 23.48\% of the validation samples in RefCOCO benchmarks. \cref{fig:subset_perf} shows significant gains using \ours over the baseline, with +20.83\% and +15.55\% mIoU improvements on the spatial and non-spatial subsets of RefCOCOg.

These improvements demonstrate that our spatial map \texttt{P-Map} contributes to the gains on both spatial and non-spatial subsets with accurate localization (\cref{fig:pmap}). For instance, we observe that \texttt{P-Map} better captures target objects in many cases (\eg, ``chair'' in the first column) than Grad-CAM~\cite{selvaraju2016grad} (\cref{fig:spatial_map}). Comprehensively, \texttt{P-Map} achieves better average performance (\eg, +6.9 mIoU) with lower inference latency (\eg, $-$2.95 \emph{sec}. per sample, details in the Appendix). This leads to our approach achieving the best results even on the very challenging PhraseCut dataset. We obtain oIoU scores of 38.53 (\emph{all}) and 38.31 (\emph{unseen}), outperforming SoTA zero-shot and supervised methods (\cref{tab:phrasecut_perf}). These results highlight the robustness and generalizability of \ours across various datasets.

\subsection{Multilingual Grounding Results}\label{exp:multi_grounding}

\begin{figure}[t]
    \centering
    \vspace{-0.5em}
    \begin{subfigure}[t]{0.45\linewidth}
        \centering
        \includegraphics[width=\linewidth]{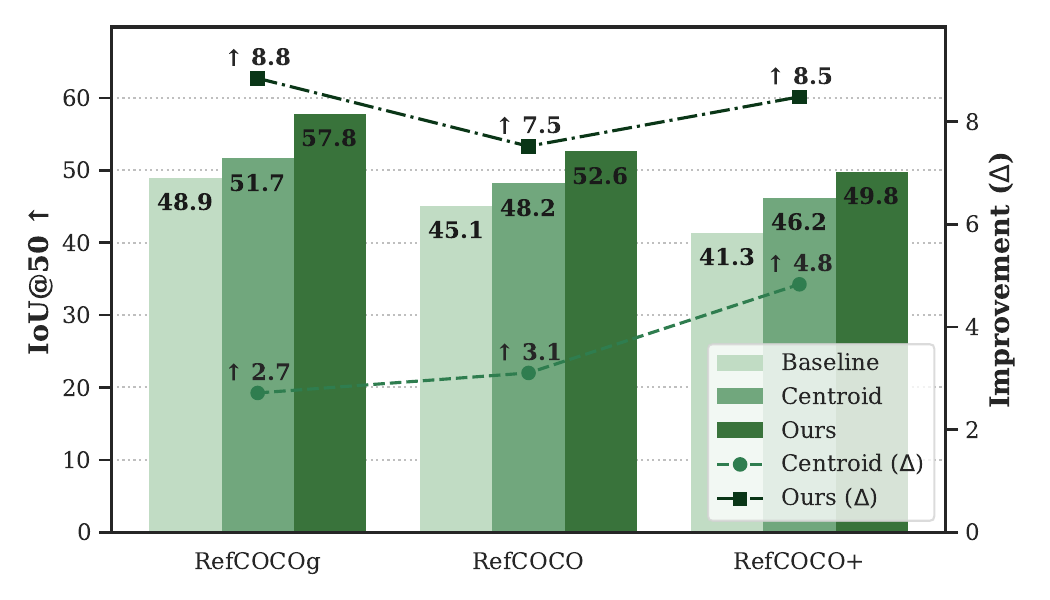}
        \caption{Spatial grounding performance}
        \label{fig:multilingual_perf}
    \end{subfigure}
    \begin{subfigure}[t]{0.45\linewidth}
        \centering
        \includegraphics[width=\linewidth]{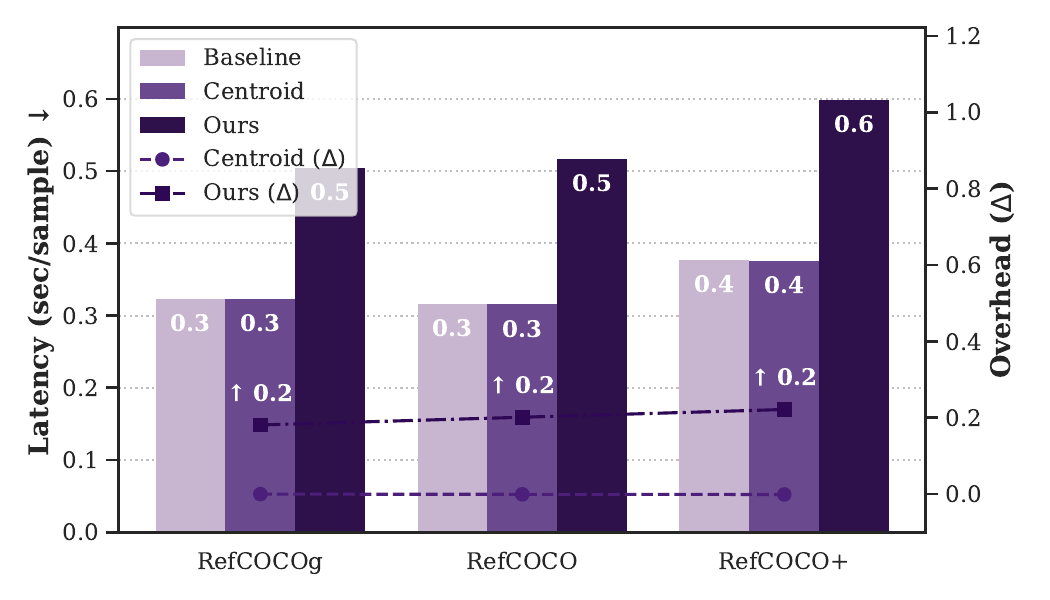}
        \caption{Inference cost}
        \label{fig:multilingual_cost}
    \end{subfigure}
    \vspace{-0.7em}
    \caption{\textbf{Comparison of Zero-shot Multilingual RIS and Inference-Cost Performance.}
    The proposed mid-layer representation extraction method achieves 
    \textbf{(a)} higher spatial grounding performance (+8--9\%)
    \textbf{(b)} at increased inference cost ($\times$1.5--1.7).}
    \label{fig:multilingual}
    \vspace{-1em}
\end{figure}

\begin{figure}[t]
    \centering
    \resizebox{0.95\linewidth}{!}{
    \includegraphics{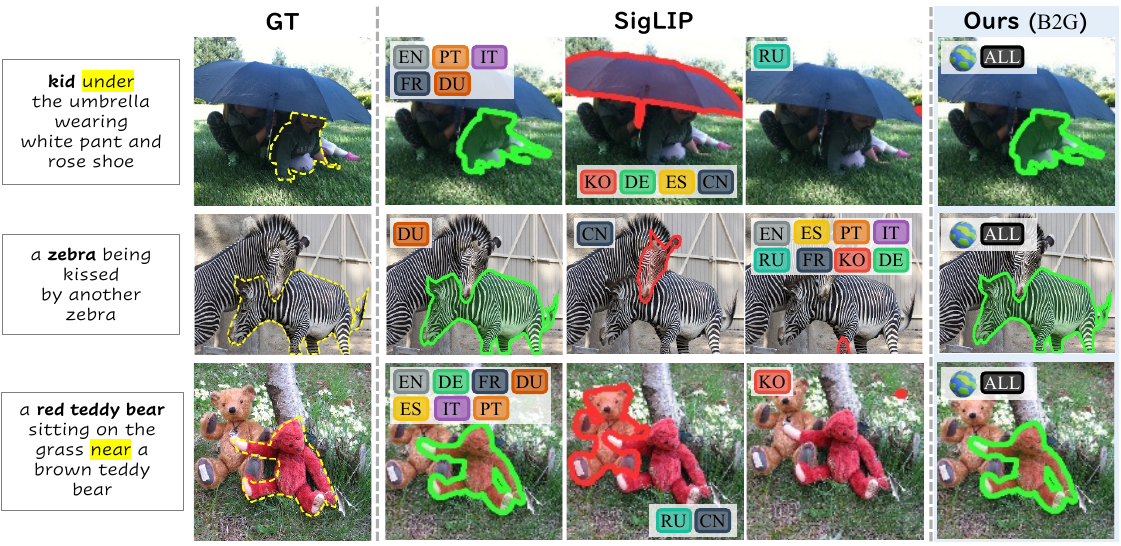}
    }
    \caption{\textbf{Qualitative on Zero-shot Multilingual RIS.} \ours leverages the centroid of mid-layer multilingual embeddings for robust and language-consistent spatial grounding.}
    \label{fig:exp_quali}
    \vspace{-1em}
\end{figure}

\paragraph{\textbf{\emph{Zero-shot multilingual RIS.}}}
\cref{fig:multilingual,fig:exp_quali} demonstrate that the proposed mid-layer multilingual centroid fusion strategy consistently improves spatial grounding results over both the SigLIP2 baseline and the final-layer centroid fusion strategy. The baseline achieves an average of 45.13 IoU@50 across all nine benchmarks, which increases to 48.67 with centroid aggregation and further to 53.41 with the proposed \ours. A similar trend is observed for mIoU: 42.42 $\rightarrow$ 46.07 (+3.65) $\rightarrow$ 49.62 (+7.20). These improvements incur modest additional computation from mid-layer feature extraction and similarity computation, reflecting a natural accuracy--cost trade-off.

\paragraph{\textbf{\emph{Qualitative multilingual grounding.}}}
\Cref{fig:exp_quali} illustrates the efficacy of \ours in mitigating both spatial blindness and language-induced biases. The baseline frequently exhibits severe localization failures due to geometric drift when processing non-English queries and struggles with explicit spatial relations (\eg, confusing the actor and receiver in \emph{a zebra being kissed}). By probing structure-preserving intermediate layers, \ours constructs a \texttt{P-Map} to recover precise positional cues. Concurrently, injecting a language-stable multilingual centroid at the selected text layer rectifies erratic geometric shifts across translated expressions. This stabilization yields a robust, language-agnostic representation that ensures highly accurate grounding across diverse languages without parameter updates, and even corrects baseline failures on standard English queries.

\begin{table*}[b]
\centering
\vspace{-1em}
\renewcommand{\arraystretch}{0.8}
\caption{\textbf{Zero-Shot Multilingual Retrieval Results (Recall@1).} We compare the SigLIP2 and \ours. Across all datasets and 10 diverse languages, particularly rescuing languages that suffer from severe geometric drift without degrading the other languages.}
\vspace{-0.7em}
\label{tab:multilingual_retrieval}
\resizebox{0.9\textwidth}{!}{%
\begin{tabular}{ll cccccccccc c}
\toprule
\multirow{2}{*}{\textbf{Dataset}} & \multirow{2}{*}{\textbf{Method}} & \multicolumn{10}{c}{\textbf{Target Language (Recall@1 $\uparrow$)}} & \multirow{2}{*}{\textbf{Avg.}} \\
\cmidrule(lr){3-12}
 & & EN & DE & FR & IT & ES & PT & RU & KO & CN & DU & \\
\cmidrule(lr){1-1}\cmidrule(lr){2-2}\cmidrule(lr){3-12}\cmidrule(lr){13-13}
\multirow{3}{*}{\shortstack[l]{RefCOCO}} 
 & SigLIP2 & 61.37 & 13.67 & 52.42 & 45.53 & 49.29 & 53.94 & 52.45 & 46.83 & 45.19 & 45.87 & 46.66 \\
\arrayrulecolor{gray!50}\cmidrule(lr){2-13}\arrayrulecolor{black}
 & +\ours & \textbf{62.67} & \textbf{25.82} & \textbf{55.56} & \textbf{47.83} & \textbf{50.44} & \textbf{56.40} & \textbf{56.37} & \textbf{52.00} & \textbf{51.41} & \textbf{50.31} & \textbf{50.88} \\[-0.8ex]
 &  & \scriptsize\texttt{\textcolor{deepblue}{+1.30}} & \scriptsize\texttt{\textcolor{deepblue}{+12.15}} & \scriptsize\texttt{\textcolor{deepblue}{+3.14}} & \scriptsize\texttt{\textcolor{deepblue}{+2.30}} & \scriptsize\texttt{\textcolor{deepblue}{+1.15}} & \scriptsize\texttt{\textcolor{deepblue}{+2.46}} & \scriptsize\texttt{\textcolor{deepblue}{+3.92}} & \scriptsize\texttt{\textcolor{deepblue}{+5.17}} & \scriptsize\texttt{\textcolor{deepblue}{+6.22}} & \scriptsize\texttt{\textcolor{deepblue}{+4.44}} & \scriptsize\texttt{\textcolor{deepblue}{+4.22}} \\

\midrule
\multirow{3}{*}{\shortstack[l]{RefCOCO+}} 
 & SigLIP2 & 30.96 & 3.96 & 19.34 & 18.52 & 21.05 & 21.05 & 20.76 & 20.80 & 21.84 & 19.82 & 19.81 \\
\arrayrulecolor{gray!50}\cmidrule(lr){2-13}\arrayrulecolor{black}
 & +\ours & \textbf{31.64} & \textbf{9.43} & \textbf{21.49} & \textbf{22.03} & \textbf{23.95} & \textbf{23.50} & \textbf{23.37} & \textbf{23.77} & \textbf{24.23} & \textbf{22.42} & \textbf{22.58} \\[-0.8ex]
 &  & \scriptsize\texttt{\textcolor{deepblue}{+0.68}} & \scriptsize\texttt{\textcolor{deepblue}{+5.47}} & \scriptsize\texttt{\textcolor{deepblue}{+2.15}} & \scriptsize\texttt{\textcolor{deepblue}{+3.51}} & \scriptsize\texttt{\textcolor{deepblue}{+2.90}} & \scriptsize\texttt{\textcolor{deepblue}{+2.45}} & \scriptsize\texttt{\textcolor{deepblue}{+2.61}} & \scriptsize\texttt{\textcolor{deepblue}{+2.97}} & \scriptsize\texttt{\textcolor{deepblue}{+2.39}} & \scriptsize\texttt{\textcolor{deepblue}{+2.60}} & \scriptsize\texttt{\textcolor{deepblue}{+2.77}} \\

\midrule
\multirow{3}{*}{\shortstack[l]{RefCOCOg}} 
 & SigLIP2 & 56.00 & 12.85 & 48.44 & 46.94 & 49.02 & 45.83 & 49.52 & 43.83 & 39.56 & 45.41 & 43.74 \\
\arrayrulecolor{gray!50}\cmidrule(lr){2-13}\arrayrulecolor{black}
 & +\ours & \textbf{57.71} & \textbf{26.40} & \textbf{51.45} & \textbf{48.85} & \textbf{51.48} & \textbf{49.50} & \textbf{53.19} & \textbf{47.51} & \textbf{46.79} & \textbf{48.72} & \textbf{50.16} \\[-0.8ex]
 &  & \scriptsize\texttt{\textcolor{deepblue}{+1.71}} & \scriptsize\texttt{\textcolor{deepblue}{+13.55}} & \scriptsize\texttt{\textcolor{deepblue}{+3.01}} & \scriptsize\texttt{\textcolor{deepblue}{+1.91}} & \scriptsize\texttt{\textcolor{deepblue}{+2.46}} & \scriptsize\texttt{\textcolor{deepblue}{+3.67}} & \scriptsize\texttt{\textcolor{deepblue}{+3.67}} & \scriptsize\texttt{\textcolor{deepblue}{+3.68}} & \scriptsize\texttt{\textcolor{deepblue}{+7.23}} & \scriptsize\texttt{\textcolor{deepblue}{+3.31}} & \scriptsize\texttt{\textcolor{deepblue}{+6.42}} \\

\bottomrule
\end{tabular}
}
\vspace{-1.5em}
\end{table*}

\subsection{Zero-shot Multilingual Text-to-Image Retrieval}

To validate our method's language-agnostic semantics, we evaluate zero-shot multilingual text-to-image retrieval. As shown in \cref{tab:multilingual_retrieval}, the baseline final-layer embeddings of SigLIP2 suffer from severe geometric drift in non-English languages. While English achieves a robust $49.44\%$ average Recall@1, languages such as German and Chinese collapse to $10.16\%$ and $35.53\%$, respectively. Without any parameter updates, our mid-layer centroid injection yields substantial improvements across all nine non-English languages, achieving average gains of $10.39\%$ for DE and $5.28\%$ for CN. \ours avoids typical generalization trade-offs by not only rescuing underperforming languages but also further enhancing the strong English baseline ($+1.23\%$ average gain). This confirms that our method acts as a targeted structural regularizer, effectively unlocking equitable and high-performance multilingual retrieval from biased VLEs.

\subsection{Ablation and Sensitivity Studies}

\begin{table}[b!]
        \vspace{-0.7em}
        \caption{\textbf{Ablation Study Results.} \textbf{(a)} Using the baseline feature extraction methods--Global-Local (GL)~\cite{yu2023zero} and HybridGL (HGL)~\cite{liu2025hybrid}, all the subcomponents are integral for building \ours. \textbf{(b)} \ours consistently achieves higher performance for a number of top candidate masks ($k$: \# of automatically generated clusters that differ per data sample). \textbf{(c)} Omitting negation or clustering reduces both mIoU and oIoU performance.}
        \vspace{-1em}
        \begin{subtable}[t]{0.38\linewidth}
        \centering
        \vspace{0pt}
        \resizebox{\linewidth}{!}{%
        \begin{tabular}{@{}lccccccc}
        \toprule
        & \multirow{2}{*}{\texttt{\textbf{CT}}} & \multirow{2}{*}{\texttt{\textbf{TC}}} & \multirow{2}{*}{\texttt{\textbf{SG}}} & \multicolumn{2}{c}{\textbf{ViT-B/32}} & \multicolumn{2}{c}{\textbf{ViT-B/16}} \\
        & & & & mIoU & oIoU & mIoU & oIoU\\ 
        \cmidrule(lr){1-1}\cmidrule(lr){2-4}\cmidrule(lr){5-6}\cmidrule(lr){7-8}
        HGL & & & & 41.54 & 29.89 & 46.59 & 37.04 \\
        \rowcolor{gray!15}
          \ours & $\checkmark$ & & & \textbf{42.58} & \textbf{30.77} & \textbf{47.01} & \textbf{37.36} \\
         \midrule
        GL & & & & 47.07 & 35.73 & 50.30 & \textbf{38.35} \\
        \rowcolor{gray!15}
          \ours & $\checkmark$ & & & \textbf{49.65} & \textbf{37.18} & \textbf{50.65} & 38.24 \\
        GL & $\checkmark$ & $\checkmark$ & & 47.70 & 36.32 & 50.68 & \textbf{38.84} \\
        \rowcolor{gray!15}
          \ours & $\checkmark$ & $\checkmark$ & & \textbf{50.16}  & \textbf{37.63} & \textbf{51.16} & \textbf{38.84} \\
        GL & $\checkmark$ & $\checkmark$ & $\checkmark$ & 50.42 & 38.17 & 50.91 & 39.58 \\
        \rowcolor{gray!15}
          \ours & $\checkmark$ & $\checkmark$ & $\checkmark$ & \textbf{52.24} & \textbf{41.43} & \textbf{54.42} & \textbf{44.14} \\
        \bottomrule
        \end{tabular}
        }
        \caption{Subcomponent ablation}
        \label{tab:subcomponent}
    \end{subtable}
    \hfill
    \begin{subtable}[t]{0.31\linewidth}
        \centering
        \vspace{0pt}
        \resizebox{\linewidth}{!}{%
        \begin{tabular}{@{}ccccc}
        \toprule
        \texttt{\textbf{TC}} & \multicolumn{2}{c}{\textbf{GL+\texttt{\textbf{CT}}+\texttt{\textbf{SG}}}} &\multicolumn{2}{c}{\cellcolor{gray!15}\textbf{\ours}} \\
        \textbf{Mask \#} & mIoU & oIoU &\cellcolor{gray!15}mIoU  & \cellcolor{gray!15}oIoU \\ 
        \cmidrule(lr){1-1}\cmidrule(lr){2-3}\cmidrule(lr){4-5}
        2 & 50.47 & 40.28 & \cellcolor{gray!15}\textbf{52.08} & \cellcolor{gray!15}\textbf{40.72} \\ 
        3 & 50.79 & 41.29 & \cellcolor{gray!15}\textbf{52.24} & \cellcolor{gray!15}\textbf{41.43} \\ 
        4 & 49.42 & 40.81 & \cellcolor{gray!15}\textbf{51.92} & \cellcolor{gray!15}\textbf{41.40} \\
        $k$ & \textendash & \textendash & \cellcolor{gray!15}\textbf{51.80} & \cellcolor{gray!15}\textbf{41.43} \\ 
        \bottomrule
        \end{tabular}
        }
        \caption{Comparison of zero-shot RIS performances using top candidates selected by the original and our candidate mask scores on RefCOCOg val (U) dataset}
        \label{tab:tc_number}
    \end{subtable}
    \hfill
    \begin{subtable}[t]{0.28\linewidth}
        \centering
        \vspace{0pt}
        \resizebox{\linewidth}{!}{%
        \begin{tabular}{@{}c cccc}
        \toprule
        \textbf{Neg.} & \textbf{Clus.} & \textbf{Refg} & \textbf{Ref}& \textbf{Ref+} \\
        \cmidrule(lr){1-2}\cmidrule(lr){3-5}
        \multicolumn{5}{c}{mIoU} \\
        \midrule
         & $\checkmark$ & 49.87 & 40.08 & 41.24 \\
        $\checkmark$ &  & 50.36 & 41.40 & 41.50 \\
        \rowcolor{gray!15}
        $\checkmark$ & $\checkmark$ & \textbf{54.92} & \textbf{49.48} & \textbf{46.19} \\
        \midrule
        \midrule
        \multicolumn{5}{c}{oIoU} \\
        \midrule
         & $\checkmark$ & 41.53 & 35.22 & 35.22 \\
        $\checkmark$ &  & 41.16 & 35.32 & 34.53 \\
        \rowcolor{gray!15}
        $\checkmark$ & $\checkmark$ & \textbf{45.21} & \textbf{41.43} & \textbf{38.90} \\
        \bottomrule
        \end{tabular}
        }
        \caption{Effect of neg. and clus.}
        \label{tab:neg_cluster_avg}
    \end{subtable}
    \vspace{-0.5em}
    \label{tab:ablation}
    \vspace{-2em}
\end{table}

\paragraph{\textbf{\emph{Component ablation.}}}
\cref{tab:subcomponent} first highlights the individual and combined effects of $E_{\text{ConText}}$ (\texttt{CT}), top-candidate (\texttt{TC}) mask selection via \texttt{P-Map}, and the spatial guider (\texttt{SG}). Incorporating \texttt{CT} and \texttt{TC}--both novel components of our framework--consistently improves VLE performance, with particularly notable gains on CLIP ViT-B/32 (+2.46 and +1.32 mIoU and oIoU, respectively). Importantly, the improvements increase with each component added, indicating that leveraging mid-layer representations is critical to overall performance gains. These work complementarily: \texttt{CT} refines language-aligned representations, while \texttt{TC} exploits spatial cues extracted from mid-layer features, together enabling more accurate spatial grounding.

\paragraph{\textbf{\emph{\texttt{P-Map} design effects.}}}
We further examine the robustness of \texttt{P-Map} through top candidate mask quality (\cref{tab:tc_number}). Across varying numbers of selected candidates ($k$), our method consistently outperforms Global-Local~\cite{yu2023zero} (+\texttt{CT}+\texttt{SG}), demonstrating that \texttt{P-Map} produces reliable region proposals, with the top three masks yielding the best performance. Moreover, both negation and clustering are essential (\cref{tab:neg_cluster_avg}), as removing either step leads to a 4--11\% drop, underscoring their importance for robust spatial grounding.

\paragraph{\textbf{\emph{Multilingual module ablation.}}}
Text-side evaluations confirm that the language-induced geometric drift is robustly resolved by our multilingual mid-layer centroid injection, avoiding the hyperparameter sensitivity often observed in final-layer fusions. Detailed text-side ablations are provided in the Appendix.
\section{Related Work}\label{relworks}

\subsubsection{Zero-shot RIS.}
While traditional RIS relies on costly pixel-level annotations~\cite{yang2022lavt, wang2022cris} or weakly-supervised training~\cite{strudel2022weakly, yu2024pseudo}, zero-shot methods circumvent task-specific training by leveraging pretrained VLEs in a proposal-and-rank pipeline: an off-the-shelf segmentor generates candidate masks~\cite{kirillov2023segment, cheng2022masked, liang2023open, wang2022freesolo}, and a frozen VLE ranks them via image--text similarity. Training-free improvements mainly focus on mask scoring through feature fusion (Cropping, Global-Local~\cite{yu2023zero}, HybridGL~\cite{liu2025hybrid}), heuristic spatial reasoning (ReCLIP~\cite{subramanian2022reclip}), token-level similarity (Region Token~\cite{li2022adapting}), or text augmentation (TAS~\cite{suo-etal-2023-text}). Spatial guidance using diffusion models (Ref-Diff~\cite{ni2023ref}) or Grad-CAM (CaR~\cite{sun2024clip}; IteRPrimE~\cite{wang2025iterprime}) requires additional models or gradient backpropagation. Crucially, these prior methods primarily rely on final-layer multimodal embeddings, which inherently suffer from spatial insensitivity and language-dependent shifts. In contrast, our framework bypasses this alignment bottleneck by extracting a novel spatial map (\texttt{P-Map}) directly from position-sensitive intermediate layers in a single forward pass.

\subsubsection{Multilingual RIS.}

Extending RIS beyond English introduces additional challenges, as language-specific phrasing can distort cross-modal alignment. Multilingual benchmarks such as Room-Across-Room~\cite{ku2020roomacrossroommultilingualvisionandlanguagenavigation}, xGQA~\cite{pfeiffer2022xgqacrosslingualvisualquestion}, and COMFORT~\cite{zhang2024comfort}, along with large-scale resources like WIT~\cite{Srinivasan_2021} and XM3600~\cite{thapliyal2022crossmodal3600massivelymultilingualmultimodal}, enable multilingual evaluation and pretraining. Notably, many of these evaluation datasets are constructed by translating English-centric benchmarks, which inherently introduces translation-induced semantic drift~\cite{nogueira2025comprehensionmultilingual}. Prior works typically mitigate such biases through resource-intensive pretraining and distillation (MURAL~\cite{jain2021muralmultimodalmultitaskretrieval}, UC2~\cite{zhou2021uc2universalcrosslingualcrossmodal}, mCLIP~\cite{chen-etal-2023-mclip}) or by developing new multilingual backbones (SigLIP~\cite{zhai2023sigmoid}, SigLIP2~\cite{tschannen2025siglip}). In contrast, our method improves multilingual grounding at test time by probing frozen VLEs to identify language-stable intermediate layers and using a multilingual centroid to stabilize cross-modal similarity.
\section{Conclusion}\label{conclusion}

We revisit zero-shot referring image segmentation through the lens of layer-wise representational bias in pretrained Vision–Language Encoders. Our analysis shows that final-layer multimodal embeddings, though optimized for global semantic alignment, suppress structurally informative signals.
This results in weakened positional sensitivity in vision embeddings and language-dependent geometric shifts in multilingual text embeddings. By probing intermediate layers and selectively leveraging structurally preserved representations, we mitigate both spatial and cross-lingual gaps without architectural changes or fine-tuning.
Beyond RIS, the identified mid-layer pathway may also benefit other multimodal tasks that require fine-grained structural reasoning. A deeper theoretical understanding of how alignment objectives shape layer-wise geometry could further enable bias-aware use of pretrained VLEs.


\bibliographystyle{splncs04}
\bibliography{main}

\clearpage
\newpage
\appendix
\renewcommand{\thefigure}{S\arabic{figure}}
\renewcommand{\thetable}{S\arabic{table}}
\renewcommand{\theequation}{S\arabic{equation}}
\setcounter{figure}{0}
\setcounter{table}{0}
\setcounter{equation}{0}

\section*{Appendix}\label{appendix}

\begin{itemize}
  \item \cref{app:method}: \textbf{Method Details.}
  We provide omitted formulations and implementation details for context-token extraction, \texttt{P-Map} construction, visual-layer selection, and reranking.

  \item \cref{app:relwork}: \textbf{Supplementary Related Work.}
  We further discuss prior studies on referring image segmentation and multilingual vision-language encoders.

  \item \cref{app:ris}: \textbf{Additional Zero-shot RIS Results.}
  We present additional evidence for the core zero-shot RIS setting, including spatial-map diagnostics, mask-generator analysis, qualitative comparisons, and sensitivity studies.

  \item \cref{app:multi}: \textbf{Multilingual Benchmark Details and Text-Side Analysis.}
  We clarify the multilingual datasets and evaluation protocols used in our experiment and examine the multilingual stabilization module through runtime analysis, text-side retrieval ablations, cross-lingual synergy matrices, and qualitative examples.

  \item \cref{app:quali}: \textbf{Further Qualitative Results.}
  We show qualitative results on zero-shot RIS and zero-shot text-to-image retrieval.

  \item \cref{app:dis}: \textbf{Discussion.}
  We summarize limitations, future directions, and broader impact.
\end{itemize}

\section{Method Details}\label{app:method}

This section provides supplementary methodological details that support the descriptions in the main text. We first explain the extraction of context tokens used to construct refined hybrid text features (\cref{app:method_ct}), followed by the formulation of the proposed spatial map (\cref{app:method_tc}). We then describe the visual layer selection strategy based on cluster coherence (\cref{app:method_layer}) and present alternative formulations for the reranking score for top candidate mask selection (\cref{app:method_overlap}).

\subsection{Context Token (\texttt{CT}) Extraction for Refined Hybrid Text Features}\label{app:method_ct}

Previous studies have demonstrated the effectiveness of hybrid text features that combine global and local-level features~\cite{yu2023zero, liu2025hybrid}. Herein, the local text feature encodes only the primary noun phrase. Specifically, it is commonly extracted as the primary noun phrase~\cite{yu2023zero, liu2025hybrid} or chunk~\cite{wang2025iterprime}. However, this could leave out important spatial cues that might appear after the selected noun phrase. Subsequently, without appending contextual cues, different expressions may collapse to the same resulting mask, limiting spatial disambiguation. Hence, we hypothesize that augmenting context information during the generation of local-level text features could enhance spatial grounding.

Overall, we construct hybrid text features that fuse both global and context-aware local representations. Given the referring expression $t$, we first extract global-level text features by feed-forwarding the entire text input into the text encoder ($\phi_{\mathcal{T}}$), resulting in $E_{\text{Sen}}(=\phi_{\mathcal{T}}(t))$. For local features, we extract two parts: the primary noun phrase ($N_{O}$) and the context token ($N_{C}$). Then, these two chunks are concatenated at an input level and encoded jointly, generating $E_{\text{Noun}}(=\phi_{\mathcal{T}}([N_{O}|N_{C}]))$. Lastly, we merge the global-level and local-level text features by addition per hidden dimension: $E_{\text{ConText}}=\gamma E_{\text{Sen}}+ (1-\gamma)E_{\text{Noun}} \in \mathbb{R}^{d}$ (\emph{attention-level} merging in \cref{app:res_attn}). Our final hybrid text feature is then used for generating our context-aware spatial map. Unlike previous hybrid feature approaches that restrict local features to a single noun phrase, our formulation guides the model with better spatial cues from the contextual tokens for localizing target objects. This yields a more discriminative textual representation, serving as the foundation for our context-aware spatial map introduced in the next subsection.

For implementation details, we use the NLP software library, Stanza~\cite{qi2020stanza}, to automatically extract $N_{O}$ and $N_{C}$, where $N_{C}$ are noun or adjective tokens that are not contained in $N_{O}$ that may appear on the right side of the primary noun phrase/chunk~\cite{wang2025iterprime}. Note that we observe a slight performance improvement using Stanza instead of spaCy~\cite{honnibal2017spacy}. For instance, using Stanza and spaCy on our method (\ours) in the RefCOCOg (val) dataset results in IoU performances of 52.24 and 51.50, respectively. Based on our preliminary analysis, we notice that this method is particularly beneficial when $N_{C}$ consists of noun chunks that are well-captured by the model (\eg, ``woman smiling'') and color adjectives (qualitative results in \cref{app:res_qual}).

\subsection{Novel Spatial Map (\texttt{P-Map}) Generation}\label{app:method_tc}
\begin{figure*}[ht!]
\centering
\includegraphics[width=\textwidth]{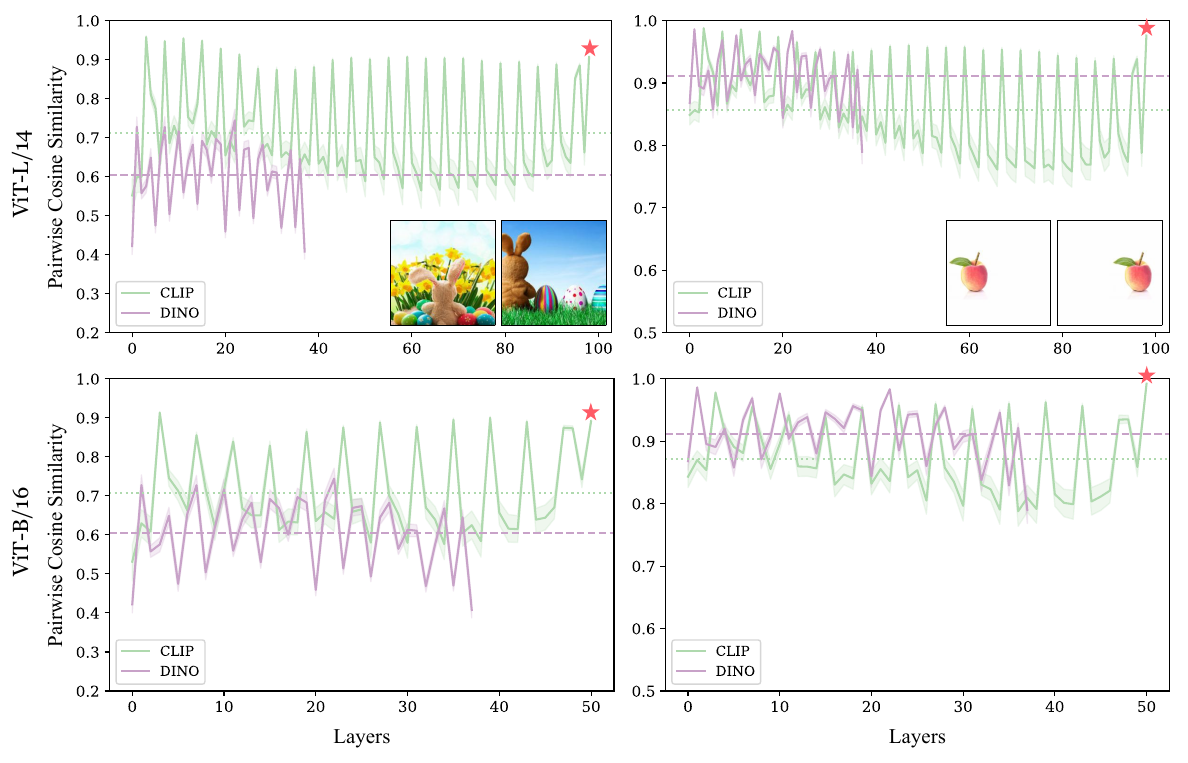}
\caption{\textbf{Cosine similarity trend across layers for image pairs containing objects at different locations.} We observe a decreasing overall trend of similarity for two VLEs, but a high peak at the last layer.}
\label{fig:layer_all}
\vspace{+3em}
\end{figure*}

We explain in detail why the negation operation is principled rather than heuristic, and why it is necessary to produce an intuitive spatial map (\texttt{P-Map}) that resolves the opposite visualization problem described in previous work~\cite{li2023clip}. To begin with, the spatial map is built based on the observation that the cosine similarity between embeddings of image pairs containing identical objects located at different positions is lower in intermediate layers than that of the final layer (\cref{fig:layer_all}). This decreasing trend of similarity across layers implies that the intermediate image embeddings may better preserve spatial information than the original image features extracted from the final layer of the visual encoder.

\begin{figure*}[ht!]
\centering
\includegraphics[width=\textwidth]{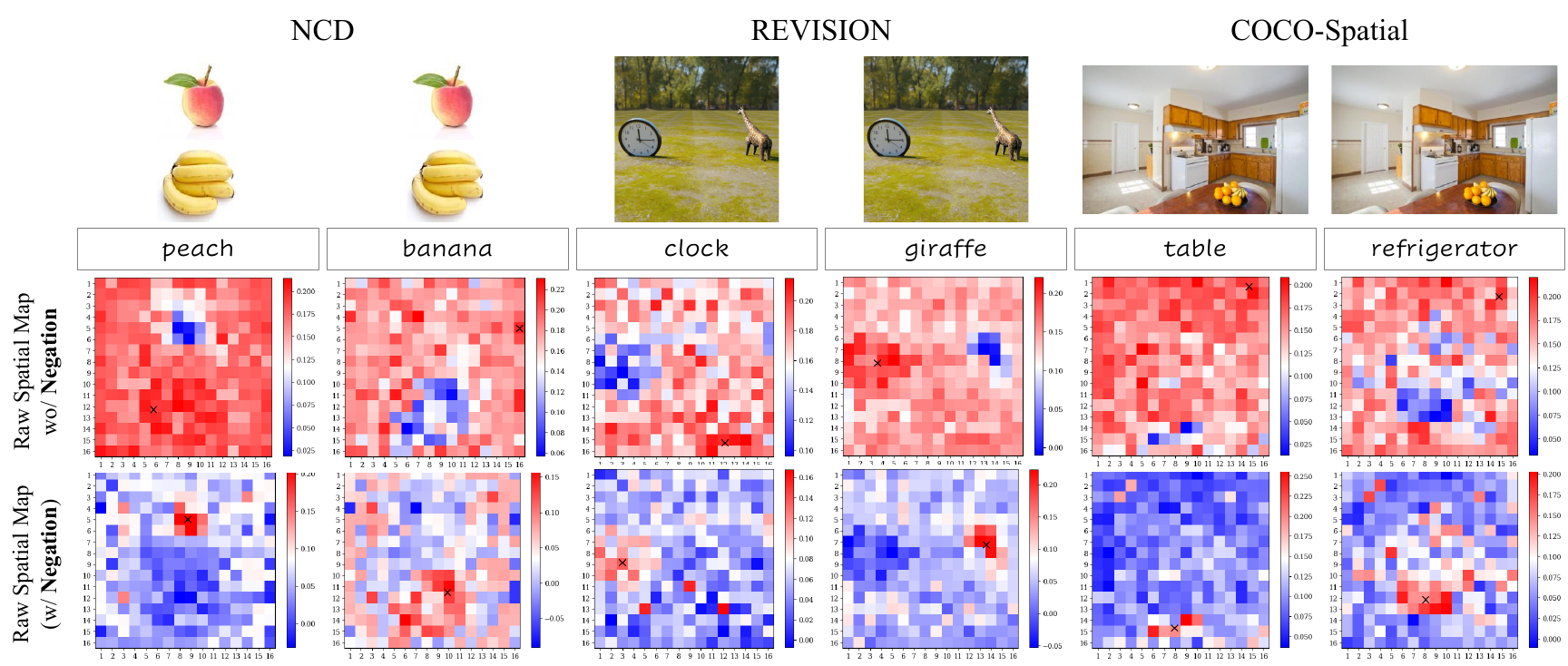}
\caption{\textbf{Opposite visualization without the negation (above) and our final raw spatial map (bottom).} The final raw spatial map with negation operation localizes the ``target objects'' and resolves the opposite visualization issue.}
\label{fig:opposite_map}
\vspace{+2em}
\end{figure*}
\begin{figure*}[ht!]
\centering
\includegraphics[width=\textwidth]{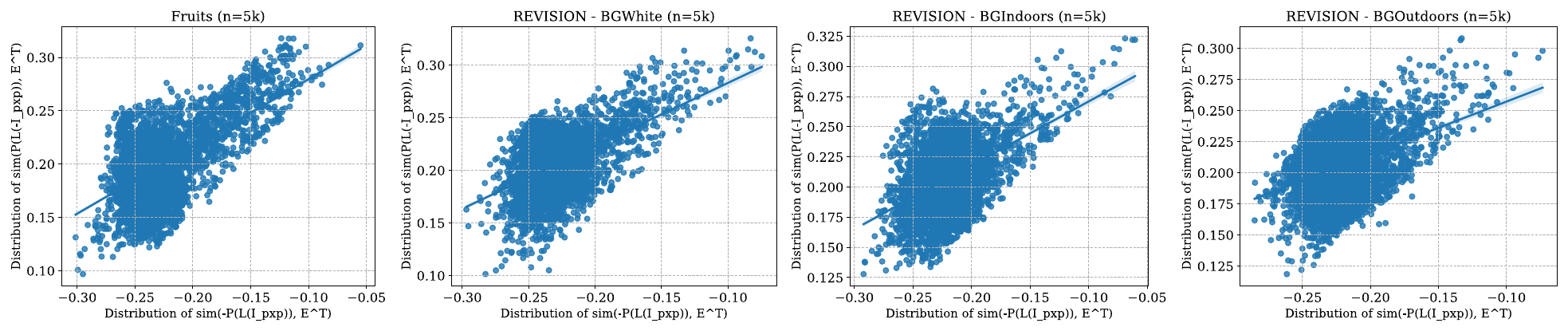}
\caption{\textbf{Correlation plots between our final spatial map (\texttt{P-Map}) image-text similarity scores (y-axis) and negative of the image-text similarity scores from the oppositely visualized spatial map (x-axis).} There exists a strong positive correlation between these two similarity score values across data samples and datasets.}
\label{fig:corr}
\end{figure*}

However, when we calculate the raw similarity between the hybrid text feature ($E_{\text{ConText}} \in \mathbb{R}^{d}$) and patch-level image features extracted from the $l$-\emph{th} exit layer ($E_{\text{p}}^{l} \in \mathbb{R}^{p \times p \times d}$), the similarity score values of the resulting similarity map ($\Tilde{\mathcal{M}}^{l} \in \mathbb{R}^{p \times p}$) is shown with opposite visualization (\cref{fig:opposite_map}). Moreover, if we simply negate the similarity map itself, the opposite visualization phenomenon is naturally resolved, but the range of similarity scores falls within $-$0.3 to 0.0 ($x$-axis in \cref{fig:corr}). We could shift these scores by an addition in order to make them fall into the original range of similarity scores (0.0 to 0.3). However, this postprocessing method requires an additional hyperparameter for the shifting and does not fundamentally solve the opposite visualization problem.

\begin{figure*}[ht!]
\centering
\includegraphics[width=\textwidth]{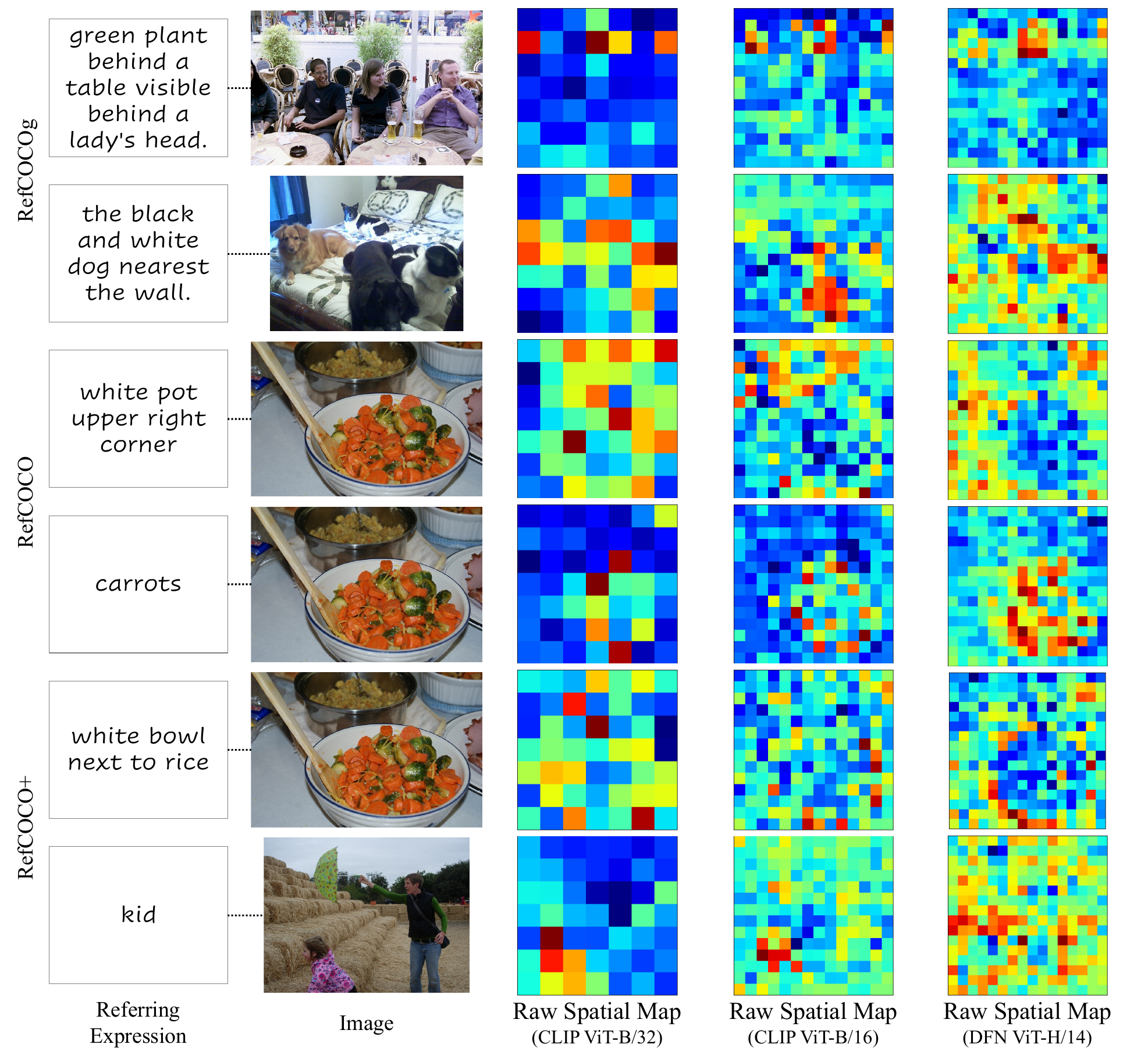}
\caption{\textbf{Comparison of raw spatial maps using different backbones for \ours.} The spatial map becomes more fine-grained for the backbone with higher dimensions for the number of patches (7 $\times$ 7 in CLIP ViT-B/32; 14 $\times$ 14 in CLIP ViT-B/16; 16 $\times$ 16 in DFN ViT-H/14).}
\label{fig:maps}
\end{figure*}
\begin{figure*}[ht!]
\centering
\includegraphics[width=\textwidth]{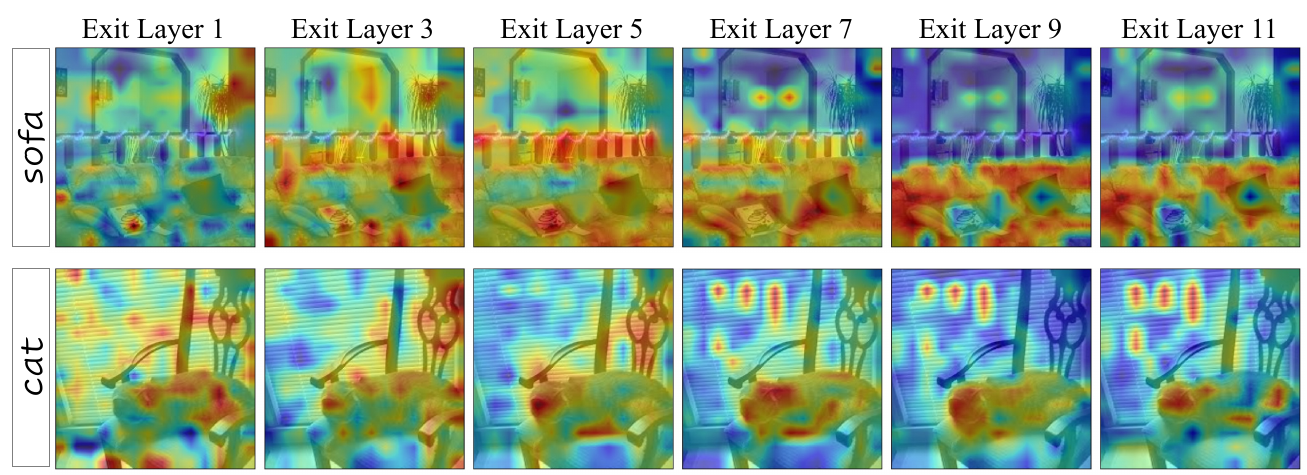}
\caption{\textbf{Comparison of our spatial map, \texttt{P-Map} generated using varying exit layers.} \texttt{P-Map} captures the most intuitive image-text similarity using the intermediate layers close to the last layer (\eg, layer 9).}
\label{fig:layer_ablation}
\end{figure*}

On the other hand, simply negating the patch-level image embeddings before feeding them into the post-layer norm and multiplying them by the projection weight naturally resolves the opposite-visualization and out-of-range issues. As shown in \cref{fig:corr}, the correlation between values of $\mathcal{M}^{l}_{ij} = \cos({\Tilde{E_{\text{p}}^{l}}[i,j], E_{\text{ConText}}})$ ($y$-axis) and $\mathcal{M}^{l}_{\text{Neg}} = -\cos({E_{\text{p}}^{l}[i, j], E_{\text{ConText}}})$ ($x$-axis) is positive ($\forall i,j = \{1, ..., p\}$). This observation is a product of the definition of cosine similarity:
\[
\cos(-x, y) = \frac{(-x) \cdot y}{\| -x \| \| y \|} = \frac{-(x \cdot y)}{\| x \| \| y \|} = -\cos(x, y)
\]
This indicates that the sign inversion directly counteracts the observed anti-correlation between foreground patch features and the query direction, providing a low-cost mechanism for recovering foreground-localized responses. Our raw spatial map ($\mathcal{M}^{l}=(\mathcal{M}^{l}_{ij})_{1 \leq i,j \leq p}$) can be found in \cref{fig:maps} across different VLE backbones. The final interpolated, spatial map, \texttt{P-Map}, constructed using different layers, can also be observed in \cref{fig:layer_ablation}.

\subsection{Visual Encoder Layer Selection and Cluster Coherence}\label{app:method_layer}

\begin{figure}
\centering
\vspace{-25pt}
\resizebox{0.7\textwidth}{!}{
  \begin{minipage}{0.7\textwidth}
    \begin{algorithm}[H]
    	\caption{Spatial Map Clustering} 
    	\begin{algorithmic}[1]
            \Require Spatial map $\mathcal{M}\in\mathbb{R}^{p \times p}$
            
            \State $\mathcal{B} = (\mathcal{B}_{ij})$, where $\forall i,j$: $\mathcal{B}_{ij} \gets [\mathcal{M}_{ij} > \delta]$
            
            \State $\mathcal{C} \gets \mathbf{0}_{p \times p}$, $l \gets 0$ 

            \For{each patch coordinate $\mathbf{k}$ in $\mathcal{B}$}
            
            \If{$\mathcal{B}[\mathbf{k}] = 1$ \textbf{and} $\mathcal{C}[\mathbf{k}] = 0$} 
                \State $l \gets l + 1$
                
                \State $Q \gets \text{new Queue()}$ 
                \State $Q.\text{push}(\mathbf{k})$, $\mathcal{C}[\mathbf{k}] \gets l$ 
                
                \While{$Q$ is not empty} 
                    \State $\mathbf{p} \gets Q.\text{pop}()$
                    \For{each neighbor $\mathbf{n}$ of $\mathbf{p}$}
                        \If{$\mathbf{n}$ is valid \textbf{and} $\mathcal{B}[\mathbf{n}]=1$ \textbf{and} $\mathcal{C}[\mathbf{n}]=0$}
                            \State $Q.\text{push}(\mathbf{n})$, $\mathcal{C}[\mathbf{n}] \gets l$
                        \EndIf
                    \EndFor
                \EndWhile
            \EndIf
        \EndFor

            \State \Return Interpolated $\mathcal{C}$
    	\end{algorithmic} 
    \end{algorithm}
  \end{minipage}
}
\caption{\textbf{Clustering algorithm.} Clusters of the spatial map are formed by threshold-exceeding adjacent patches and interpolated into the clustered spatial map.}
\label{alg:alg1}
\end{figure}

The main paper considers three strategies for instantiating the selected visual layer
$l^{*}_{\mathrm{vis}} \in \mathcal{L}_{\mathrm{vis}}$:
dynamic selection via a permutation probe (\oursdynperm),
a fixed mid-layer heuristic (\oursfix),
and dynamic selection via cluster coherence (\oursdynclus).
All three strategies share the same downstream grounding pipeline.
They differ only in how the visual layer $l^{*}_{\mathrm{vis}}$ is chosen.

\subsubsection{Dynamic permutation-based selection (\oursdynperm).}

Given an input image $I$, we construct a spatially permuted image $I^{\pi}$ by shuffling
non-overlapping tiles over the $p \times p$ grid.
For each candidate layer $l \in \mathcal{L}_{\mathrm{vis}}$,
let $E_{\mathrm{p}}^{l}(I) \in \mathbb{R}^{p \times p \times d}$ and
$E_{\mathrm{p}}^{l}(I^{\pi}) \in \mathbb{R}^{p \times p \times d}$
denote the projected patch features of the original and permuted images, respectively.
We restore spatial correspondence by applying the inverse permutation:
\[
\hat{E}_{\mathrm{p}}^{l}(I^{\pi})
=
\pi^{-1}\!\left(E_{\mathrm{p}}^{l}(I^{\pi})\right).
\]
We then compute the layer-wise equivariance score
\[
\Gamma^{l}
=
\frac{1}{p^2}
\sum_{i=1}^{p^2}
\cos\!\left(
E_{\mathrm{p}}^{l}(I)[i],\,
\hat{E}_{\mathrm{p}}^{l}(I^{\pi})[i]
\right).
\]
Larger $\Gamma^{l}$ indicates that the patch features remain similar after permutation,
implying a weaker dependence on spatial arrangement.
Therefore, \oursdynperm selects the most position-sensitive layer by
\[
l^{*}_{\mathrm{vis}}
=
\arg\min_{l \in \mathcal{L}_{\mathrm{vis}}} \Gamma^{l}.
\]

\subsubsection{Fixed mid-layer heuristic (\oursfix).}

Instead of selecting a layer separately for each image, \oursfix uses a single shared
visual layer for all test instances.
Let
\[
l_{\mathrm{vis}}^{\mathrm{fix}} \in \mathcal{L}_{\mathrm{vis}},
\qquad
\delta^{\mathrm{fix}} \in (0,1)
\]
denote the exit layer and threshold chosen once on a small calibration split.
Following the experimental setup, we determine
$l_{\mathrm{vis}}^{\mathrm{fix}}$ and $\delta^{\mathrm{fix}}$
on 10\% of the RefCOCOg val (U) split and keep them fixed at test time.
Thus, for every test image,
\[
l^{*}_{\mathrm{vis}} = l_{\mathrm{vis}}^{\mathrm{fix}},
\qquad
\delta = \delta^{\mathrm{fix}}.
\]
This removes per-sample layer search and yields the lowest inference cost among the
three strategies.

\subsubsection{Dynamic cluster-coherence selection (\oursdynclus).}

For \oursdynclus, we first compute the layer-wise spatial map (\texttt{P-Map})
$\mathcal{M}^{l}$ for each candidate visual layer $l \in \mathcal{L}_{\mathrm{vis}}$.
We then convert the continuous map into a binary mask $\mathcal{B}^{l}$
via thresholding to identify activated patches.
A Breadth-First Search (BFS) is applied to group spatially connected, adjacent patches
into clusters, producing a cluster map $\mathcal{C}^{l}$ (\cref{alg:alg1}).
Note that the number of clusters ($k_l$ or $k$) is determined automatically and may vary
across samples.

To select the optimal visual layer, we compute a cluster coherence score for each layer
based on the resulting \texttt{P-Map} clusters.
Intuitively, layers that preserve spatial structure tend to produce compact and
coherent clusters.
For a candidate visual layer $l \in \mathcal{L}_{\mathrm{vis}}$, let
\[
\mathcal{B}^{l}(u)=\mathbf{1}\!\left[\mathcal{M}^{l}(u)>\delta\right],
\qquad
u \in \Omega := \{1,\ldots,p\}\times\{1,\ldots,p\},
\]
and let $\{C_{r}^{l}\}_{r=1}^{k_l}$ denote the connected components extracted from
$\mathcal{B}^{l}$ by BFS, where
\[
C_{r}^{l}=\{u \in \Omega \mid \mathcal{C}^{l}(u)=r\}.
\]

We first min-max normalize the spatial map:
\[
\bar{\mathcal{M}}^{l}(u)=
\frac{
\mathcal{M}^{l}(u)-\min_{v\in\Omega}\mathcal{M}^{l}(v)
}{
\max_{v\in\Omega}\mathcal{M}^{l}(v)-\min_{v\in\Omega}\mathcal{M}^{l}(v)+\varepsilon
}.
\]

For each cluster $C_{r}^{l}$, we compute an inside-outside response contrast:
\[
\Delta_{r}^{l}
=
\frac{1}{|C_{r}^{l}|}\sum_{u\in C_{r}^{l}}\bar{\mathcal{M}}^{l}(u)
-
\beta
\frac{1}{|\Omega \setminus C_{r}^{l}|+\varepsilon}\sum_{u\in \Omega\setminus C_{r}^{l}}\bar{\mathcal{M}}^{l}(u),
\]
where $\beta \ge 0$ controls the penalty on diffuse activation outside the cluster. If
$\Omega \setminus C_{r}^{l}=\emptyset$, we interpret the second term as $0$.

To explicitly favor compact connected regions, we additionally define a simple
compactness factor
\[
\kappa_{r}^{l}
=
\frac{|C_{r}^{l}|}{|\partial C_{r}^{l}|+\varepsilon},
\]
where $\partial C_{r}^{l}$ is the set of boundary patches in $C_{r}^{l}$ that have
at least one 4-neighbor outside $C_{r}^{l}$.

The cluster coherence score of layer $l$ is then defined as a size-weighted average
over all connected components:
\[
\operatorname{CCS}(l)
=
\frac{1}{\sum_{r=1}^{k_l}|C_{r}^{l}|}
\sum_{r=1}^{k_l}
|C_{r}^{l}|\,\kappa_{r}^{l}\,\Delta_{r}^{l}.
\]

If a layer produces no activated component (i.e., $k_l=0$), we set
\[
\operatorname{CCS}(l)=-\infty.
\]

Finally, \oursdynclus selects
\[
l^{*}_{\mathrm{vis}}
=
\arg\max_{l\in\mathcal{L}_{\mathrm{vis}}}
\operatorname{CCS}(l).
\]

For convenience, the three strategies can be summarized as
\[
l^{*}_{\mathrm{vis}}
=
\begin{cases}
\arg\min_{l\in\mathcal{L}_{\mathrm{vis}}}\Gamma^{l},
& \text{for } \oursdynperm, \\[0.4em]
l_{\mathrm{vis}}^{\mathrm{fix}},
& \text{for } \oursfix, \\[0.4em]
\arg\max_{l\in\mathcal{L}_{\mathrm{vis}}}\operatorname{CCS}(l),
& \text{for } \oursdynclus.
\end{cases}
\]

Once $l^{*}_{\mathrm{vis}}$ is determined by any of the above rules, the corresponding \texttt{P-Map} $\mathcal{M}^{\,l^{*}_{\mathrm{vis}}}$ is converted into the spatial guidance mask
and used in the reranking stage described below.
In the simplest variant of \oursdynclus, one may set $\kappa_{r}^{l}=1$ and $\beta=1$,
which reduces the score to a pure inside-versus-outside response contrast.

\subsection{Reranking Score Variants}\label{app:method_overlap}
In the main text, the overlap score $O_k$ is defined as the IoU between the proposal
mask $M_k$ and the guidance mask $G$. As an alternative, let
$\widetilde{\mathcal{M}}^{\,l^{*}_{\mathrm{vis}}}\in[0,1]^{H\times W}$ denote the
normalized \texttt{P-Map} $\mathcal{M}^{\,l^{*}_{\mathrm{vis}}}$ upsampled to the image
resolution. We also compute $O_k$ using the mean \texttt{P-Map} response within the proposal
region, \ie,
\begin{equation}
O_k = \frac{1}{|M_k|} \sum_{i \in M_k} \widetilde{\mathcal{M}}^{\,l^{*}_{\mathrm{vis}}}[i].
\end{equation}
where $|M_k|$ denotes the number of pixels within the proposal mask. Empirically,
this formulation produces trends comparable to IoU-based overlap (non-statistically
significant drop of 0.01, averaged across RefCOCO datasets), confirming that the
effectiveness of the reranking stage is not sensitive to the specific overlap metric.

\section{Supplementary Related Work}\label{app:relwork}

This section complements the concise related-work discussion in the main paper and clarifies the multilingual benchmark and evaluation protocol used in our experiments. Since our multilingual results operate on translated referring expressions while keeping the underlying visual target fixed, a precise description of the benchmark construction and test-time usage is important for interpreting the gains of \ours.

\subsubsection{Supervised and weakly-supervised RIS.}

Traditional RIS aims to segment image regions described by free-form text expressions by relying on costly pixel-level annotations to learn dense cross-modal correspondences. Early approaches introduced language-aware vision transformers (LAVT~\cite{yang2022lavt}) and CLIP-driven decoders trained with contrastive losses (CRIS~\cite{wang2022cris}). Subsequent works focused on parameter-efficient fusion (RISCLIP~\cite{kim2023extending}, ETRIS~\cite{xu2023bridging}), prompt-driven instance refinement (Prompt-Driven RIS~\cite{shang2024prompt}), bridging SAM and CLIP representations~\cite{ito2025feature}, unified multi-task architectures (UniNeXt~\cite{lin2023uninext}), fine-grained pixel-word alignment (MagNet~\cite{chng2024mask}), and distortion-aware learning (MaskRIS~\cite{leemaskris}). To mitigate the high cost of manual masking, weakly-supervised methods emerged, relying solely on image-text pairs. These approaches enforce text-region consistency~\cite{strudel2022weakly}, maintain intra-chunk coherence~\cite{lee2023weakly}, utilize shatter-and-gather mechanisms~\cite{kim2023shatter}, or leverage foundation models to generate pseudo-masks for training (Pseudo-RIS~\cite{yu2024pseudo}).

\begin{figure*}[!t]
\centering
\includegraphics[width=0.8\textwidth]{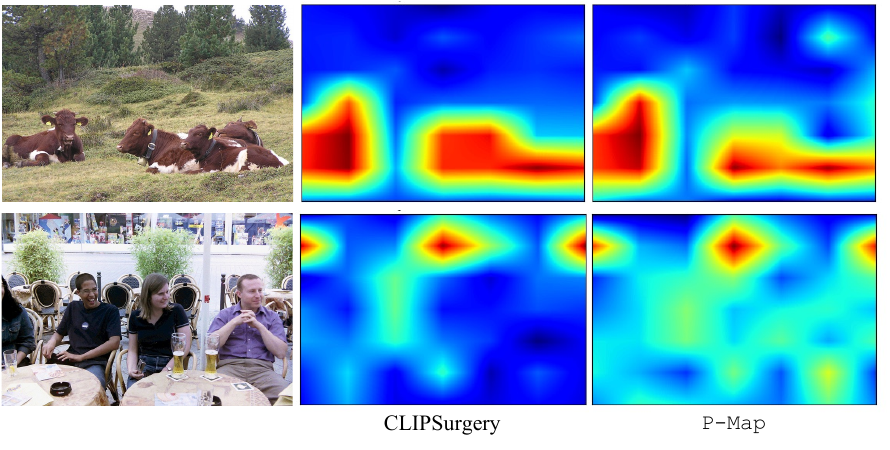}
\caption{\textbf{Activation pattern comparison between CLIPSurgery and \texttt{P-Map}.} Our spatial map shows a very similar activation as CLIPSurgery without attention recomputations.}
\label{fig:clipsurgery}
\vspace{-2em}
\end{figure*}
\begin{figure*}[!t]
\centering
\includegraphics[width=0.8\textwidth]{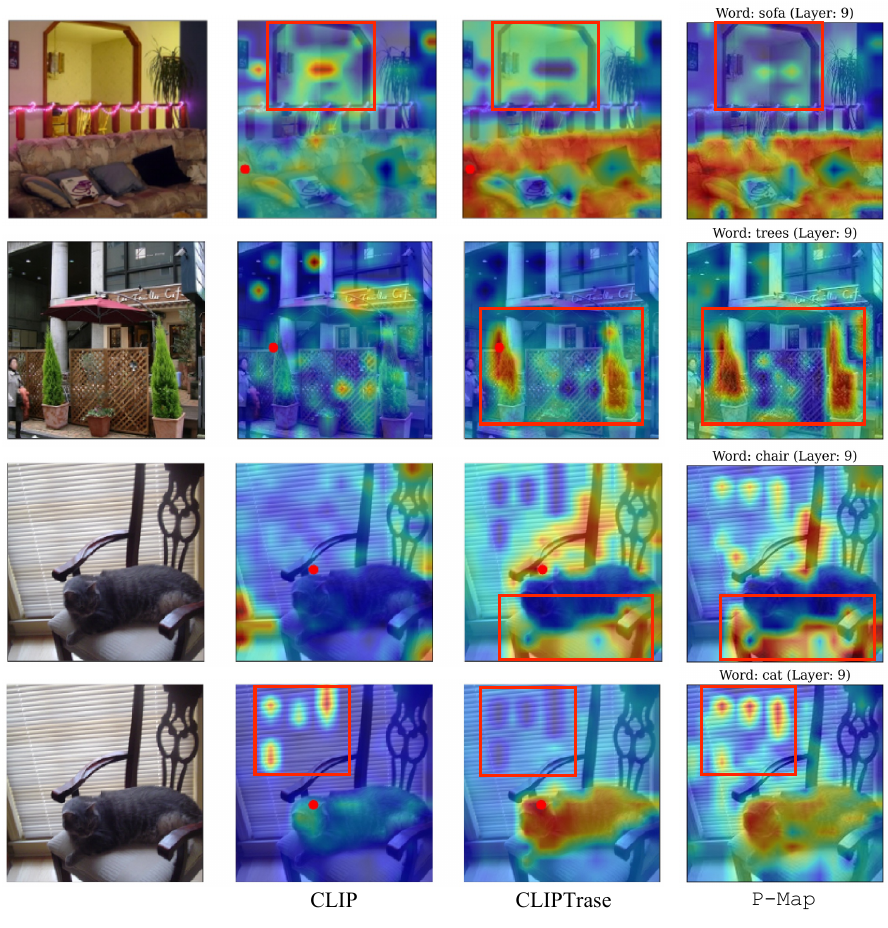}
\vspace{-1em}
\caption{\textbf{Activation pattern comparison between CLIPTrase and \ours.} Although \ours does not show completely clean activations without noises (red blobs on the last row), it can still capture highly activated areas like CLIPTrase.}
\vspace{-1em}
\label{fig:cliptrace}
\end{figure*}

\subsubsection{Multilingual VLEs and cross-lingual stabilization.}

Extending vision-language alignment beyond English introduces significant challenges, as language-specific phrasing and grammatical structures can severely distort cross-modal alignment. While large-scale resources like WIT~\cite{Srinivasan_2021} and XM3600~\cite{thapliyal2022crossmodal3600massivelymultilingualmultimodal} support multilingual pretraining, most prior works mitigate language biases through resource-intensive pretraining and distillation (\eg, MURAL~\cite{jain2021muralmultimodalmultitaskretrieval}, UC2~\cite{zhou2021uc2universalcrosslingualcrossmodal}, mCLIP~\cite{chen-etal-2023-mclip}). Recently, advanced multilingual backbones like SigLIP~\cite{zhai2023sigmoid} and SigLIP2~\cite{tschannen2025siglip} have been introduced with scaled multilingual curation and revised training objectives. These models are powerful foundations, but their standard usage still relies on the final-layer embedding as the cross-modal interface.

Our work is complementary to these training-intensive directions. Rather than training a new multilingual RIS model (\eg, the fully trained attention-anchored model in Nogueira \etal~\cite{nogueira2025comprehensionmultilingual}) or adapting the VLE weights, \ours studies whether frozen multilingual VLEs already contain a more language-stable geometry in their intermediate layers. Our method operates entirely at test time: It probes the frozen text encoder across parallel translations, identifies a language-stable intermediate layer, computes a multilingual centroid, and injects that centroid before final contextualization. Thus, our contribution is not a new multilingual backbone, but a test-time mechanism for exposing latent multilingual structure that is obscured by the final-layer abstraction bottleneck.


\section{Supplementary Zero-shot RIS Results}\label{app:ris}

This section presents additional results that are not included in the main paper. Note that all inference is conducted on a single RTX A4000 for all VLEs, except for SigLIPs and DFN ViT-H/14, where a single RTX A6000 is used. To ensure a fair comparison for calculating the inference time in \cref{tab:time}, we use the same device, RTX A6000, for all the methods.

\subsection{Spatial Map Diagnostics and Efficiency}\label{app:ris_map}

\subsubsection{Spatial map comparison with segmentation approach.}\label{app:res_map}

Our final spatial map, \texttt{P-Map}, is surprisingly comparable to several zero-shot segmentation models, such as CLIPSurgery~\cite{li2023clip} (\cref{fig:clipsurgery}) and CLIPTrase~\cite{shao2024explore} (\cref{fig:cliptrace}). CLIPSurgery is generated via recomputation of the self-attention operation ($A=\sigma(s \cdot VV^{T})V$ instead of $A=\sigma(s \cdot QK^{T})V$, where $s$: scaling factor) for every layer (before the start of the new path). Similarly, CLIPTrase also requires a recomputation of self-attention for producing a more consistent semantic correlation matrix. However, \texttt{P-Map} does not require recomputations for self-attention modules and produces intuitive activation patterns similar to those of these pre-trained segmentors.

\begin{figure*}[!t]
\centering
\includegraphics[width=\textwidth]{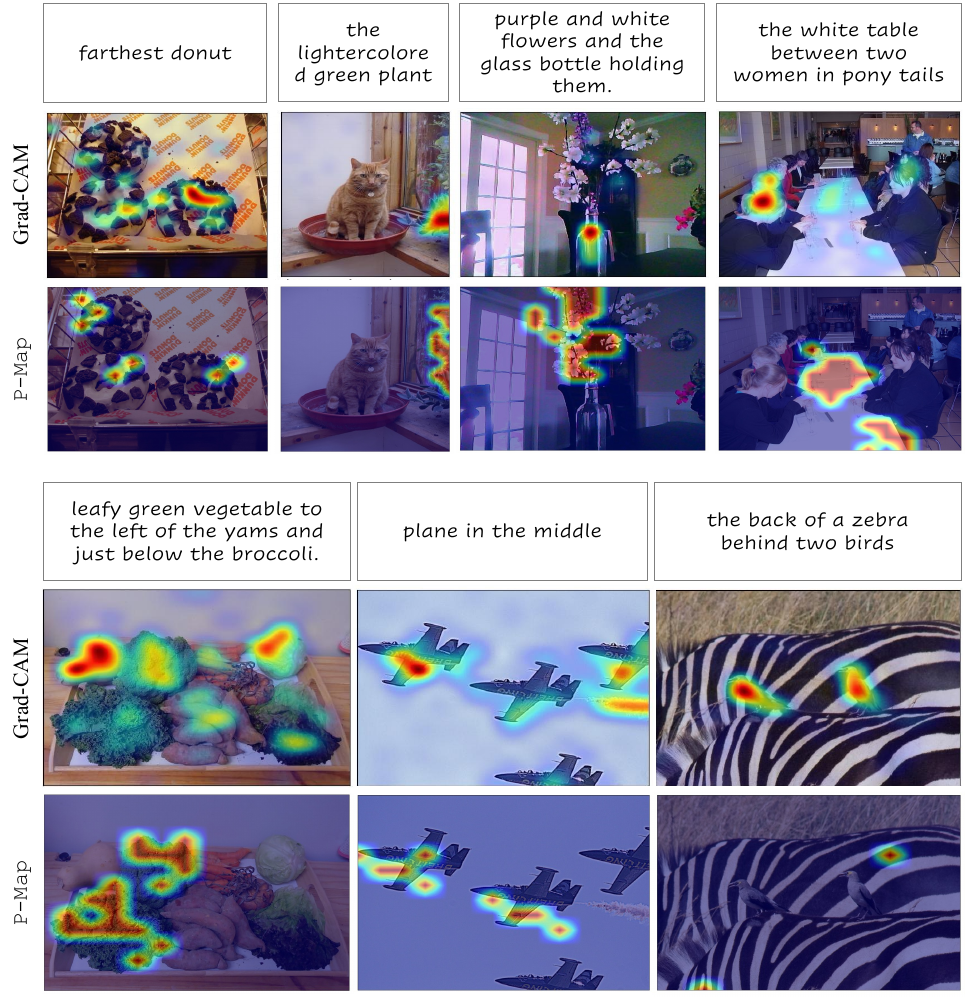}
\caption{\textbf{Qualitative comparison between Grad-CAM (IteRPrimE) and \texttt{P-Map} (\ours).} Our \texttt{P-Map} can better capture and localize the primary objects in the referring expressions.}
\label{fig:gradcam_full}
\vspace{-1em}
\end{figure*}

\subsubsection{Spatial map comparison with gradient-based approach.}\label{app:res_map_2}

When compared to Grad-CAM~\cite{selvaraju2016grad}, which is often applied in several zero-shot RIS approaches~\cite{subramanian2022reclip, wang2025iterprime, lee2023weakly}, \cref{fig:gradcam_full} demonstrates that our \texttt{P-Map} can provide better spatial cues for a primary object from a referring expression. Furthermore, our \ours (using \texttt{P-Map}) takes less inference time than IteRPrimE, which uses Grad-CAM, as shown in \cref{tab:time} while achieving better IoU performance.

\begin{table*}[!t]
\centering
\caption{\textbf{Inference time and performance comparison between IteRPrimE and \ours on validation sets.} The reported time result is in seconds per iteration, averaged across the first 100 samples for each dataset. Our methods comparatively take less time than IterPRIMe while achieving better IoU performance.}
\resizebox{0.85\textwidth}{!}{%
\begin{tabular}{@{}lcccccccc}
\toprule
&  \multicolumn{2}{c}{\textbf{RefCOCOg}} & \multicolumn{2}{c}{\textbf{RefCOCO}} & \multicolumn{2}{c}{\textbf{RefCOCO+}} & \multicolumn{2}{c}{\textbf{Avg.}} \\ 
& Time & mIoU & Time & mIoU & Time & mIoU & Time & mIoU \\ 

 \cmidrule(lr){1-1}\cmidrule(lr){2-3}\cmidrule(lr){4-5}\cmidrule(lr){6-7}\cmidrule(lr){8-9}
 IterPRIMe \cite{wang2025iterprime} & 3.20 & 46.0 & 5.42 & 40.2 & 4.82 & 44.2 & 4.48 & 43.5 \\
 \rowcolor{gray!15} \ours (CLIP ViT-B/32) & \textbf{1.27} & 52.2 & \textbf{1.45} & 46.8 & \textbf{1.44} & 44.1 & \textbf{1.38} & 47.7 \\
 \rowcolor{gray!15} \ours (CLIP ViT-B/16) & \underline{1.40} & \underline{54.4} & \underline{1.60} & \underline{50.1} & \underline{1.60} & \underline{46.8} & \underline{1.53} & \underline{50.4} \\
 \rowcolor{gray!15} \ours (DFN ViT-H/14) & 2.24 & \textbf{56.2} & 2.52 & \textbf{50.9} & 2.53 & \textbf{47.1} & 2.43 & \textbf{51.4} \\
\bottomrule
\end{tabular}
}
\label{tab:time}
\end{table*}

\begin{table*}[!t]
\centering
\caption{\textbf{Zero-shot performances of SoTA referring image extraction methods applied with three mask generators on RefCOCOg val set.} The highest IoU performances are generally achieved using Mask2Former across different methods. $\dag$ stands for the reproduced results.}
\resizebox{0.9\textwidth}{!}{%
\begin{tabular}{@{}lcccccc}
\toprule
& \multicolumn{2}{c}{\textbf{FreeSOLO}} & \multicolumn{2}{c}{\textbf{SAM}} & \multicolumn{2}{c}{\textbf{Mask2Former}} \\ 
& mIoU & oIoU & mIoU & oIoU & mIoU & oIoU \\ 
 \cmidrule(lr){1-1}\cmidrule(lr){2-3}\cmidrule(lr){4-5}\cmidrule(lr){6-7}
 Global-Local (CLIP ViT-B/32) \cite{yu2023zero} & 33.59 & 31.02 & 38.56 & 27.04 & \textbf{45.23} & \textbf{32.59} \\
 HybridGL (CLIP ViT-B/32) \cite{liu2025hybrid} & 30.66 & 27.97 & 41.54 & 29.89 & \textbf{44.27} & \textbf{32.17} \\
 HybridGL$^{\dag}$ (CLIP ViT-B/16) & 30.84 & 28.02 & \textbf{46.59} & \textbf{37.04} & 43.90 & 31.73 \\
 IteRPrimE (ALBEF) \cite{wang2025iterprime} & 29.03 & 26.36 & 25.23 & 21.27 & \textbf{46.89} & \textbf{41.09} \\
 \cmidrule(lr){1-1}\cmidrule(lr){2-3}\cmidrule(lr){4-5}\cmidrule(lr){6-7}
 Mask Upper Bound (CLIP ViT-B/32) & 52.44 & 52.76 & 74.39 & 75.76 & 76.85 & 72.73 \\
\bottomrule
\end{tabular}
}
\label{tab:masks}
\end{table*}

\subsection{Candidate Mask Quality and Additional Quantitative Results}\label{app:ris_mask}

\subsubsection{Mask generator selection.}\label{app:res_mask}
For selecting the mask generator, we compare three widely used pretrained segmentation models, FreeSOLO~\cite{wang2022freesolo}, SAM~\cite{kirillov2023segment}, and Mask2Former~\cite{cheng2022masked, liang2023open} on several SoTA zero-shot RIS approaches~\cite{yu2023zero, wang2025iterprime, liu2025hybrid}. As can be observed in \cref{tab:masks}, we observe that Mask2Former is most effective in the RefCOCOg validation set~\cite{mao2016generation}. Compared to SAM, it is also highly efficient, taking a fifth of the inference time (\eg, 150 min. \emph{vs}. 30 min. using SAM and Mask2Former on 2,573 data instances) while mostly achieving higher IoU performances across different methods and similar upper bound scores (\eg, mIoU of 74.39 and 76.85 using SAM and Mask2Former, respectively).

\subsubsection{oIoU performance.}\label{app:res_oIoU}
\cref{tab:oIoU} shows the oIoU performances of various zero-shot RIS methods, including ours, across RefCOCO datasets. Similar to the mIoU results, \ours achieves the best overall results for all the backbones. 

\begin{table*}[t!]
\centering
\vspace{-1em}
\caption{\textbf{Comparison of Zero-shot RIS (oIoU) Performance.} Consistent with mIoU results, ours consistently achieves superior performance over existing feature extraction approaches, Global-Local~\cite{yu2023zero} and HybridGL~\cite{liu2025hybrid}, across different backbone encoders. For fair comparison, all the listed methods have been applied with \texttt{CT} and \texttt{SG} (ablation in later). The baseline in \colorbox{lightred}{red} and statistically significant improvements over the baseline in \colorbox{lightblue}{blue} for each backbone.}
\vspace{-0.7em}
\label{tab:oIoU}
\renewcommand{\arraystretch}{0.9} 
\resizebox{0.88\linewidth}{!}{%
\footnotesize 
\begin{tabular}{@{} l c *{9}{>{\scriptsize}c} >{\scriptsize}l }
\toprule
\multicolumn{1}{c}{\multirow{2}{*}{\textbf{Method}}} & {\scriptsize \textbf{Pre-trained}} &
\multicolumn{3}{c}{\textbf{RefCOCOg}} &
\multicolumn{3}{c}{\textbf{RefCOCO}} &
\multicolumn{3}{c}{\textbf{RefCOCO+}} &
\multicolumn{1}{l}{\multirow{2}{*}{\textbf{Avg.}}} \\ 

& {\scriptsize \textbf{Segmentor}} & 
\multicolumn{1}{c}{val} & \multicolumn{1}{c}{test} & \multicolumn{1}{c}{valG} & 
\multicolumn{1}{c}{val} & \multicolumn{1}{c}{testA} & \multicolumn{1}{c}{testB} & 
\multicolumn{1}{c}{val} & \multicolumn{1}{c}{testA} & \multicolumn{1}{c}{testB} &  \\

\cmidrule(lr){1-1}\cmidrule(lr){2-2}\cmidrule(lr){3-5}
\cmidrule(lr){6-8}\cmidrule(lr){9-11}
\cmidrule(lr){12-12}

\rowcolor{gray!15}
\multicolumn{12}{c}{\textbf{CLIP ViT-B/32}} \\

Global-Local & \multirow{3}{*}{SAM}
& 34.53 & 36.51 & 36.40 & 31.50 & 32.03 & 30.56 & 27.17 & 29.91 & 25.31 & 31.55 \\

HybridGL & & 37.97 & 38.94 & 39.37 & 34.70 & 35.97 & 33.50 & 31.33 & 34.21 & 28.12 & \cellcolor{lightred}{34.90} \\
\arrayrulecolor{gray!65}
\cmidrule(lr){1-1}\cmidrule(lr){3-5}\cmidrule(lr){6-8}
\cmidrule(lr){9-11}\cmidrule(lr){12-12}
\arrayrulecolor{black}
\hspace{0.5em}$+\ours$ & & 37.94 & 38.90 & 39.24 & 34.86 & 36.45 & 33.67  & 32.06 & 35.39 & 28.52 & \cellcolor{lightblue}{35.23} \texttt{\tiny{(+0.33)}} \\

\arrayrulecolor{gray!65}
\midrule
\arrayrulecolor{black}

Global-Local & \multirow{5}{*}{{\shortstack{Mask2\\Former}}}
& 39.22 & 39.00 & 39.40 & 35.66 & 42.34 & 31.13 & 33.35 & 40.00 & 28.81 & 36.55 \\

HybridGL &
& 40.17 & 40.07 & 40.63 & 34.60 & 40.52 & 30.88 & 33.84 & 41.05 & 28.78 & \cellcolor{lightred}{36.73} \\

\arrayrulecolor{gray!65}
\cmidrule(lr){1-1}\cmidrule(lr){3-5}\cmidrule(lr){6-8}
\cmidrule(lr){9-11}\cmidrule(lr){12-12}
\arrayrulecolor{black}

\hspace{0.5em}$+\oursfix$ & 
& 42.20 & 42.58 & 42.84 & 38.30 & 46.76 & 33.51 & 36.63 & 45.02 & 30.15 & \cellcolor{lightblue}{39.78} \texttt{\tiny{(+3.05)}} \\

\hspace{0.5em}$+\oursdynclus$ & 
& 41.72 & 42.19 & 42.77 & 38.39 & 46.50 & 33.01 & 36.15 & 43.95 & 29.77 & \cellcolor{lightblue}{39.38} \texttt{\tiny{(+2.65)}} \\ 

\hspace{0.5em}$+\oursdynperm$ &
& 41.62 & 42.31 & 42.92 & 38.37 & 46.51 & 33.12 & 36.39 & 44.88 & 29.76 & \cellcolor{lightblue}{39.54} \texttt{\tiny{(+2.81)}} \\ 

\midrule
\rowcolor{gray!15}
\multicolumn{12}{c}{\textbf{CLIP ViT-B/16}} \\

Global-Local & \multirow{3}{*}{SAM}
& 41.84 & 42.79 & 42.15 & 33.33 & 33.27 & 33.63 & 28.45 & 30.63 & 26.65 & 35.19 \\

HybridGL & & 43.65 & 43.80 & 43.69 & 38.02 & 39.76 & 36.54 & 33.45 & 37.72 & 29.47 & \cellcolor{lightred}{38.68} \\
\arrayrulecolor{gray!65}
\cmidrule(lr){1-1}\cmidrule(lr){3-5}\cmidrule(lr){6-8}
\cmidrule(lr){9-11}\cmidrule(lr){12-12}
\arrayrulecolor{black}
\hspace{0.5em}$+\ours$ & & 44.03 & 44.02 & 44.05 & 38.47 & 40.12 & 36.95 & 34.58 & 39.24 & 30.03 & \cellcolor{lightblue}{39.50} \texttt{\tiny{(+0.82)}} \\

\arrayrulecolor{gray!65}
\midrule
\arrayrulecolor{black}

HybridGL & \multirow{5}{*}{{\shortstack{Mask2\\Former}}}
& 41.05 & 41.46 & 41.52 & 36.78 & 42.52 & 33.30 & 34.54 & 40.45 & 30.59 & 38.02 \\

Global-Local &
& 41.49 & 41.93 & 41.88 & 38.16 & 46.25 & 33.30 & 34.07 & 39.54 & 30.63 & \cellcolor{lightred}{38.58} \\
\arrayrulecolor{gray!65}
\cmidrule(lr){1-1}\cmidrule(lr){3-5}\cmidrule(lr){6-8}
\cmidrule(lr){9-11}\cmidrule(lr){12-12}
\arrayrulecolor{black}

\hspace{0.5em}$+\oursfix$ & 
& 44.13 & 46.24 & 45.26 & 40.85 & 48.45 & 34.98 & 38.39 & 46.17 & 32.15 & \cellcolor{lightblue}{41.85} \texttt{\tiny{(+3.27)}} \\

\hspace{0.5em}$+\oursdynclus$ & 
& 44.68 & 46.09 & 44.99 & 40.72 & 47.45 & 35.35 & 38.07 & 44.43 & 32.41 & \cellcolor{lightblue}{41.58} \texttt{\tiny{(+3.00)}} \\ 

\hspace{0.5em}$+\oursdynperm$ & 
& 44.30 & 46.21 & 45.18 & 40.84 & 48.06 & 35.45 & 38.33 & 44.92 & 32.47 & \cellcolor{lightblue}{41.75} \texttt{\tiny{(+3.17)}} \\ 

\midrule
\rowcolor{gray!15}
\multicolumn{12}{c}{\textbf{SigLIP ViT-B/16}} \\

Global-Local & \multirow{4}{*}{{\shortstack{Mask2\\Former}}}
& 42.85 & 42.96 & 43.29 & 41.93 & 49.51 & 35.17 & 37.03 & 43.08 & 30.36 & \cellcolor{lightred}{40.69} \\
\arrayrulecolor{gray!65}
\cmidrule(lr){1-1}\cmidrule(lr){3-5}\cmidrule(lr){6-8}
\cmidrule(lr){9-11}\cmidrule(lr){12-12}
\arrayrulecolor{black}

\hspace{0.5em}$+\oursfix$ & 
& 44.63 & 44.40 & 44.89 & 39.27 & 46.01 & 33.12 & 39.28 & 46.40 & 31.48 & \cellcolor{lightblue}{41.05} \texttt{\tiny{(+0.36)}} \\

\hspace{0.5em}$+\oursdynclus$ & 
& 44.76 & 44.60 & 45.25 & 40.07 & 47.38 & 34.52 & 39.00 & 46.18 & 32.10 & \cellcolor{lightblue}{41.54} \texttt{\tiny{(+0.85)}} \\

\hspace{0.5em}$+\oursdynperm$ & 
& 44.99 & 44.06 & 44.76 & 39.35 & 46.64 & 33.60 & 39.36 & 46.47 & 31.90 & \cellcolor{lightblue}{41.24} \texttt{\tiny{(+0.55)}} \\ 

\midrule
\rowcolor{gray!15}
\multicolumn{12}{c}{\textbf{DFN VIT-H/14}} \\

Global-Local & \multirow{4}{*}{{\shortstack{Mask2\\Former}}}
& 39.19 & 39.24 & 39.05 & 36.26 & 44.23 & 30.72 & 31.94 & 37.78 & 28.26 & \cellcolor{lightred}{36.74} \\
\arrayrulecolor{gray!65}
\cmidrule(lr){1-1}\cmidrule(lr){3-5}\cmidrule(lr){6-8}
\cmidrule(lr){9-11}\cmidrule(lr){12-12}
\arrayrulecolor{black}

\hspace{0.5em}$+\oursfix$ & 
& 46.21 & 46.03 & 46.09 & 41.24 & 48.88 & 36.15 & 38.10 & 45.11 & 33.14 & \cellcolor{lightblue}{42.33} \texttt{\tiny{(+5.59)}} \\

\hspace{0.5em}$+\oursdynclus$ & 
& 45.87 & 46.34 & 45.64 & 41.44 & 49.26 & 35.83 & 38.26 & 45.15 & 32.98 & \cellcolor{lightblue}{42.31} \texttt{\tiny{(+5.57)}} \\

\hspace{0.5em}$+\oursdynperm$ & & 45.25 & 45.90 & 45.30 & 41.20 & 49.21  & 35.22 & 38.08 & 45.26 & 32.74 & \cellcolor{lightblue}{42.02} \texttt{\tiny{(+5.28)}} \\ 

\bottomrule

\end{tabular}
}
\vspace{-1.5em}
\end{table*}

\begin{figure*}[ht!]
\centering
\includegraphics[width=0.95\textwidth]{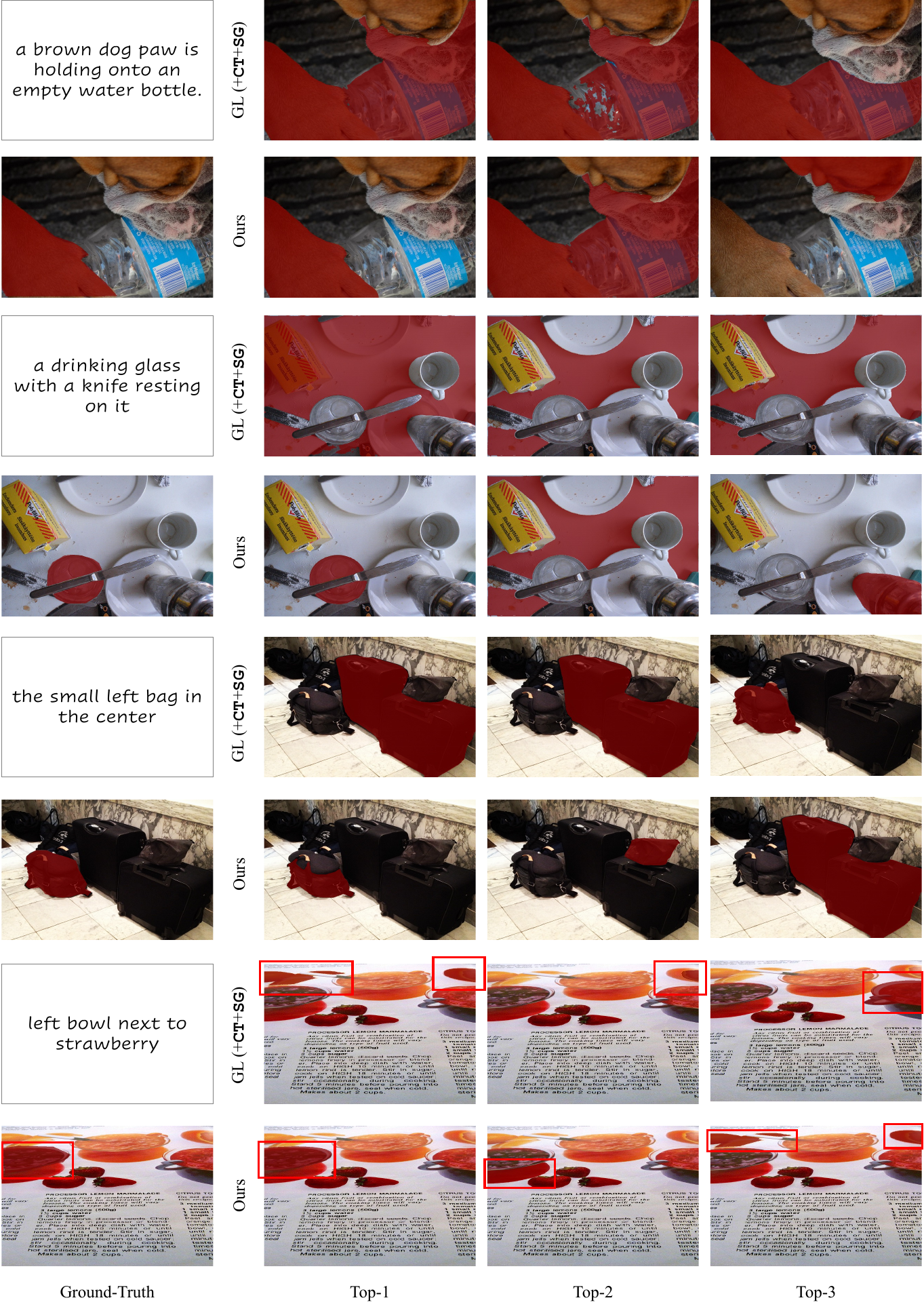}
\caption{\textbf{Qualitative comparison of top candidate masks generated using original image-text similarity from Global-Local (w/ \texttt{CT}) and averaged \texttt{P-Map} similarity score using our \ours}. The top
candidate masks generated using our method are much more diverse and accurate.}
\label{fig:qualitative_full1}
\end{figure*}
\begin{figure*}[h!]
\centering
\includegraphics[width=0.95\textwidth]{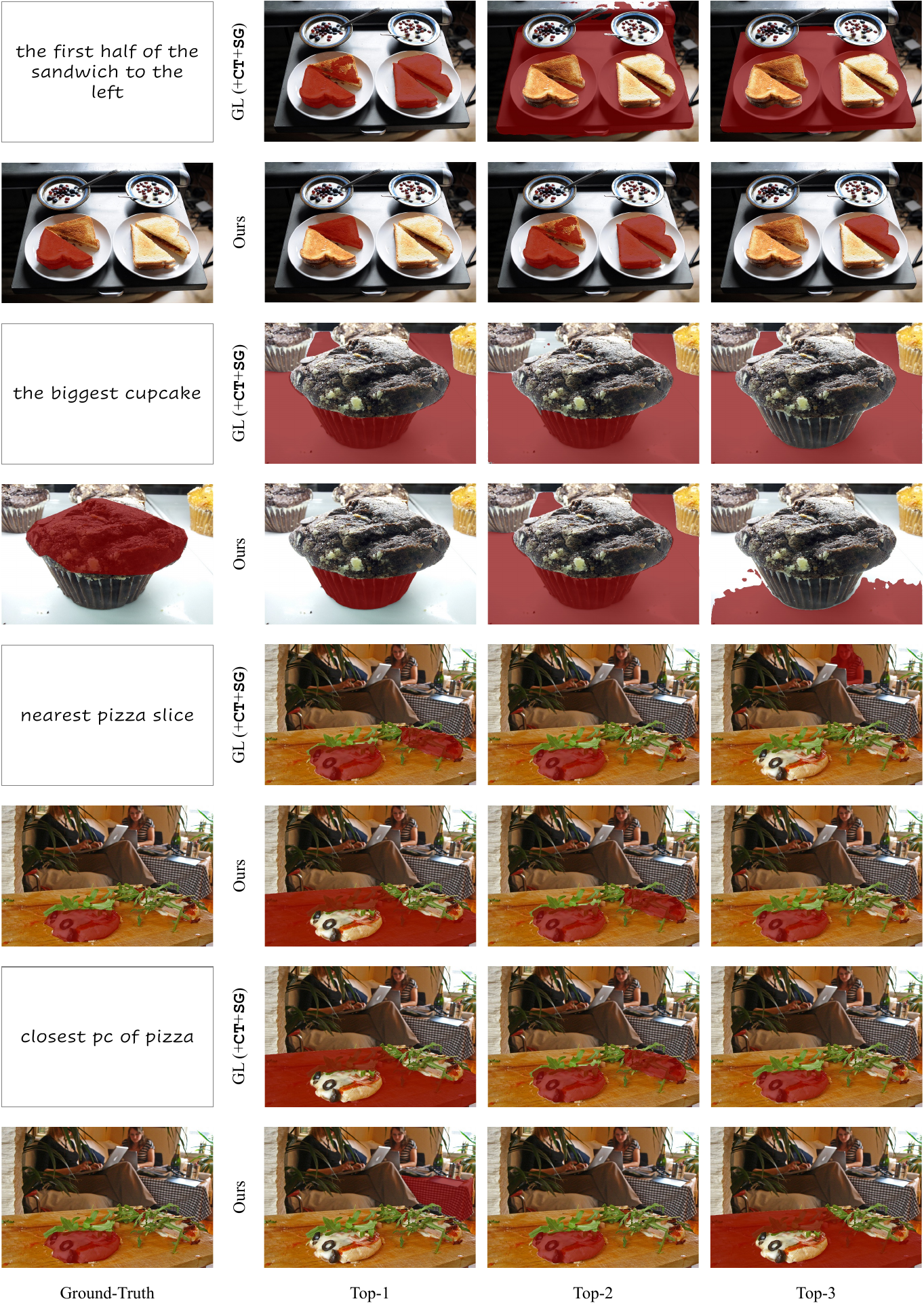}
\caption{\textbf{Qualitative comparison of top candidate masks generated using original image-text similarity from Global-Local (w/ \texttt{CT}) and averaged \texttt{P-Map} similarity score using our \ours.} The top
candidate masks generated using our method are much more diverse.}
\label{fig:qualitative_full2}
\end{figure*}

\subsection{Qualitative Analyses of Grounding Behavior}\label{app:ris_qual}

\subsubsection{Qualitative comparison.}\label{app:res_qual}
Additional qualitative results comparing the stronger Global-Local baseline used in the
main paper (\texttt{GL} + \texttt{CT} + \texttt{SG}) and \ours (top-1) are illustrated
in \cref{fig:qualitative_full1} and \cref{fig:qualitative_full2}. While
\cref{fig:qualitative_full1} shows several samples that our approach accurately segments
by predicting the top-1 candidate mask correctly, \cref{fig:qualitative_full2} shows
several challenging cases even for our approach. For example, the first-row sample in
\cref{fig:qualitative_full2} is segmented on ``the first half of the sandwich to the
\emph{right}'' instead of ``\emph{left}'' for the top-1 and top-3 predictions of
\ours. Despite this, the top candidate masks produced using our method are much more
diverse than those produced by the stronger Global-Local baseline.

\subsubsection{Context Token (\texttt{CT}) effect.}\label{app:res_ct}
\begin{figure*}[!t]
\centering
\includegraphics[width=\textwidth]{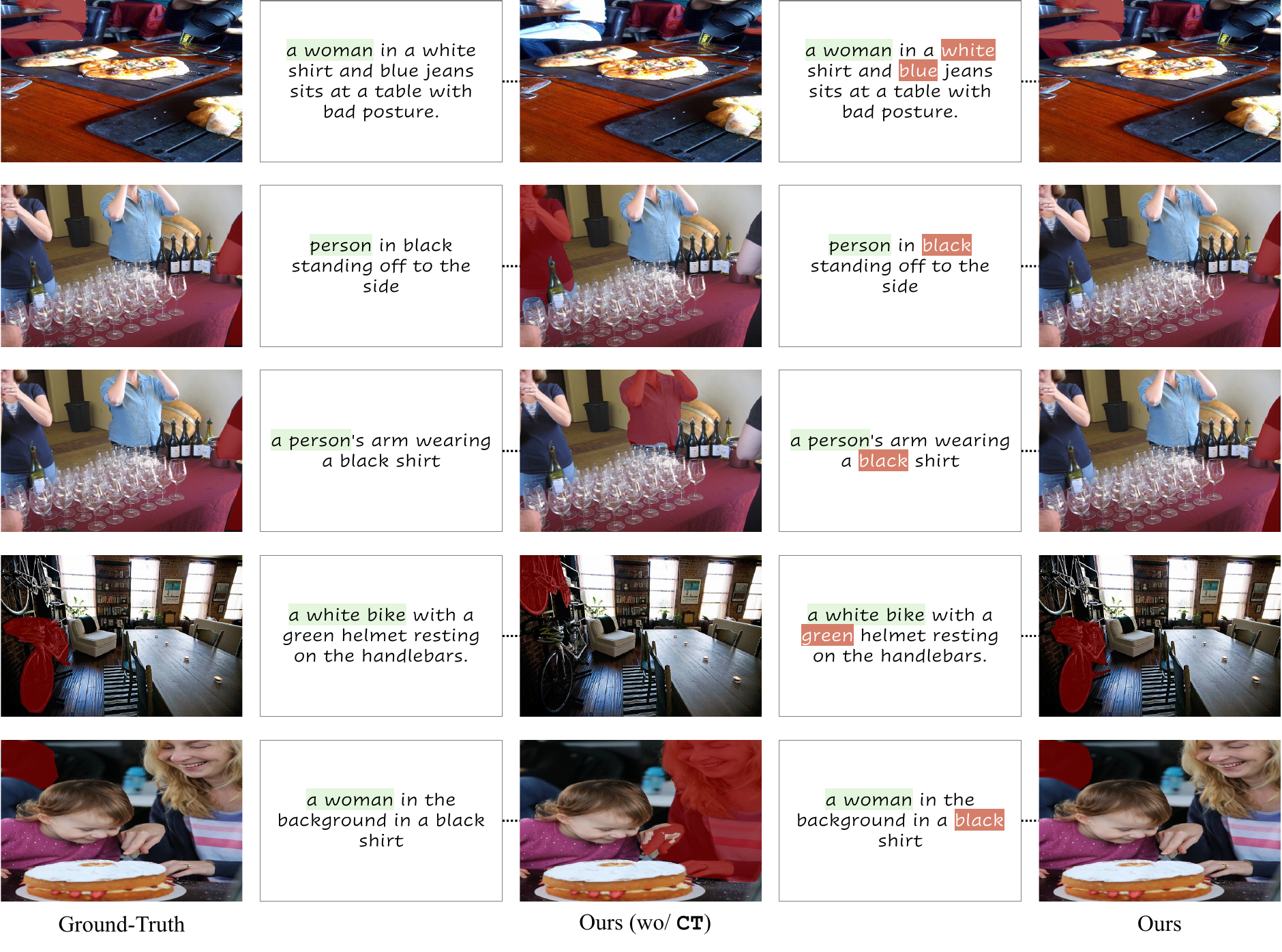}
\caption{\textbf{Qualitative comparison between \ours without and with context tokens (\texttt{CT}).} We observe an improved mask candidate using hybrid text features highlighted with proposed context tokens in the local-level text features.}
\label{fig:pc_full}
\vspace{-1em}
\end{figure*}

We also show qualitative results of the effects of our context tokens on the segmented results in \cref{fig:pc_full}. We highlight the context tokens in red that are added as the local-level features of the hybrid text features described in \cref{exp:multi_grounding}. As noted, the model could better predict the primary object from the referring expression, given the context of colors. For instance, the fourth sample in \cref{fig:pc_full} guides a model to locate the bike nearest to the `green' helmet. The last sample also corrects the model to focus on the `black' color property more than the woman wearing a dark navy jacket. We leave as future work to discover which of the context tokens within the referring expression could better guide the model for zero-shot RIS tasks.

\subsection{Protocol and Sensitivity Analyses}\label{app:ris_ablation}

\subsubsection{Spatial subset generation.}\label{app:res_qwen}
The prompt for instructing Qwen2.5-14B-Instruct~\cite{yang2024qwen2} to classify each referring expression into `Spatial' or `Non-Spatial' depending on the presence of a spatial cue is in \cref{tab:qwen_prompts}. We provide important clarifications for the model to distinguish words that have multiple definitions (\eg, right, in) and several examples for classification. The list of unique spatial cues ($n=950$) found using LLM for all the RefCOCOg, RefCOCO, RefCOCO+ validation datasets is as follows\footnote{Due to the space limit, we provide the first 173 cues only.}:

`12 o'clock',
`2nd from',
`2nd from bottom',
`3rd from bottom',
`3rd one from',
`None.',
`above',
`above left of',
`above right',
`above, left of',
`above, on, right',
`above, right',
`above, right of',
`across',
`adjacent to',
`after',
`against',
`against, on top of',
`ahead of',
`alongside',
`amongst',
`around',
`around it',
`at',
`at 4 o'clock',
`at 9 o clock',
`at the edge of',
`at the end of',
`at the end of, on the left',
`atop',
`attached to',
`back',
`back behind',
`back end of',
`back end of, on, right edge of',
`back end... front',
`back from the right',
`back left',
`back left of',
`back middle',
`back of',
`back of chair center top',
`back of the chair on the right',
`back of the pack',
`back of the room right side',
`back of, closest to',
`back of, far right',
`back of, flanking it',
`back of, in front of',
`back of, in the middle, behind',
`back of, left',
`back of, on, left',
`back on right',
`back on the right',
`back right',
`back right top',
`back row 4th from left',
`back row on left',
`back side',
`back side position',
`back top almost middle',
`back up high',
`back, in the front of',
`back, on',
`back, right',
`back, to the right',
`back... right',
`background',
`background center',
`background, far left',
`background, on',
`background, top left',
`backleft',
`backside',
`behidn',
`behind',
`behind and right of',
`behind middle',
`behind right of',
`behind right side',
`behind, front',
`behind, in, middle',
`behind, in, near',
`behind, left',
`behind, left of',
`behind, left shoulder of',
`behind, next to',
`behind, next to, on',
`behind, on the right',
`behind, on, left',
`behind, on, left side',
`behind, on, left, near',
`behind, right',
`behind, right side',
`behind, to the left of',
`below',
`below, above',
`below, in',
`below, left',
`below, on, left',
`beneath',
`beneath, on',
`beside',
`beside, closest to',
`beside, in front of',
`between',
`between, on',
`between, on the right',
`between, on, bottom left of',
`between, on, right',
`blocked by',
`blocking',
`bottom',
`bottom center',
`bottom from top',
`bottom front',
`bottom front left',
`bottom left',
`bottom left area',
`bottom left corner',
`bottom left corner of',
`bottom left of',
`bottom left portion of couch on left',
`bottom left side',
`bottom left, on',
`bottom leftcorner',
`bottom middle',
`bottom most',
`bottom most left',
`bottom of',
`bottom portion of it',
`bottom right',
`bottom right corner',
`bottom right corner of pic',
`bottom right front',
`bottom right next to',
`bottom right of',
`bottom right on',
`bottom right under',
`bottom rightcorner',
`bottom rightmost',
`bottom row',
`bottom row 2nd from left',
`bottom row second from left',
`bottom row second kid from left',
`bottom row, left',
`bottom to right',
`bottom, 2nd from top',
`bottom, first row, closest to',
`bottom, left',
`bottom, on, top, left',
`bottom-left',
`bottom-right',
`bottom-right, closest to',
`bottom/center',
`bottomright',
`by',
`by the wall',
`by the, on, left',
`by, with, to',
`center',
`center directly in front of',
`center front',
`center left',
`center leftish',
`center of',
`center of pic',
`center right',
`center row far right',
`center, behind',
`center/background',
`centermost',
`centernearest.'

\begin{table*}[h!]
\centering
\caption{\textbf{Prompts used for generating spatial subsets of RefCOCO, RefCOCO+, and RefCOCOg validation sets.} We prompt Qwen2.5-14B-Instruct to divide the samples into Spatial and Non-Spatial subsets.}
\resizebox{0.83\textwidth}{!}{
\begin{tcolorbox}[colback=white, colframe=gray, boxsep=3pt, left=3pt, right=3pt, top=3pt, bottom=3pt, width=\textwidth, sharp corners, title= System and User Prompts for Spatial Subset Generation, fontupper=\small, fontlower=\small]
You are Qwen, created by Alibaba Cloud.\\
You are a helpful assistant.\\
Please carefully analyze user sentences to see if they contain spatial words or phrases.\\
Spatial expressions are any words or phrases indicating location, direction, orientation, or proximity.\\
These include (but are not limited to): `on', `in' (meaning `inside'), `under', `front', `behind', `middle', `center', `left', `right' (direction), `closest to', `near', `beside', `above', `below', `next to', etc.\\
\\
\textbf{Important clarifications}:\\
- If `right' means `correct', it is NOT spatial.\\
- If `in' means `wearing' (\eg, `in a costume'), it is NOT spatial.\\
- Action expressions like `facing', `looking up' is NOT spatial.\\
- `closest to','near', or `middle' should be treated as spatial. Respond with a specific format so it is easy to parse.'\\
\tcblower
Below are examples of how to classify:\\
\\
Example 1:\\
Sentence: ``The chair with the stuffed animal owl sitting in it.''\\
Analysis: `in' here is adverb.\\
Answer: IsSpatial: False, SpatialExpression: None\\
\\
Example 2:\\
Sentence: ``The apple is on the table.''\\
Analysis: `on' indicates spatial.\\
Answer: IsSpatial: True, SpatialExpression: on\\
\\
Example 3:\\
Sentence: ``left side monitor''\\
Analysis: `left' indicates spatial.\\
Answer: IsSpatial: True, SpatialExpression: left\\
\\
Example 4:\\
Sentence: ``A small lamb lying closest to the adult.''\\
Analysis: `closest to' indicates a spatial relationship.\\
Answer: IsSpatial: True, SpatialExpression: closest to\\
\\
Now, analyze this new sentence:\\
\{sentence\}\\
\\
\textbf{Instructions}:\\
1) If the sentence contains any spatial relationship expression, output True and the exact expression (\eg, `on', `in', `closest to').
Important: actions or wearing attributes are NOT spatial expression. (\eg a girl facing the camera. -> `facing' is NOT means direction or spatial relationship)\\
2) If no spatial expression is found, output False and None.\\
3) Return your answer in the format exactly:
''IsSpatial: <True/False>, SpatialExpression: <expression or None>''
\end{tcolorbox}
}
\label{tab:qwen_prompts}
\end{table*}

\subsubsection{Hyperparameter tuning for \oursfix.}\label{app:res_hyp}
\begin{figure*}[!t]
    \centering
    \begin{subfigure}[t]{0.32\textwidth}
        \centering
        \includegraphics[width=\textwidth]{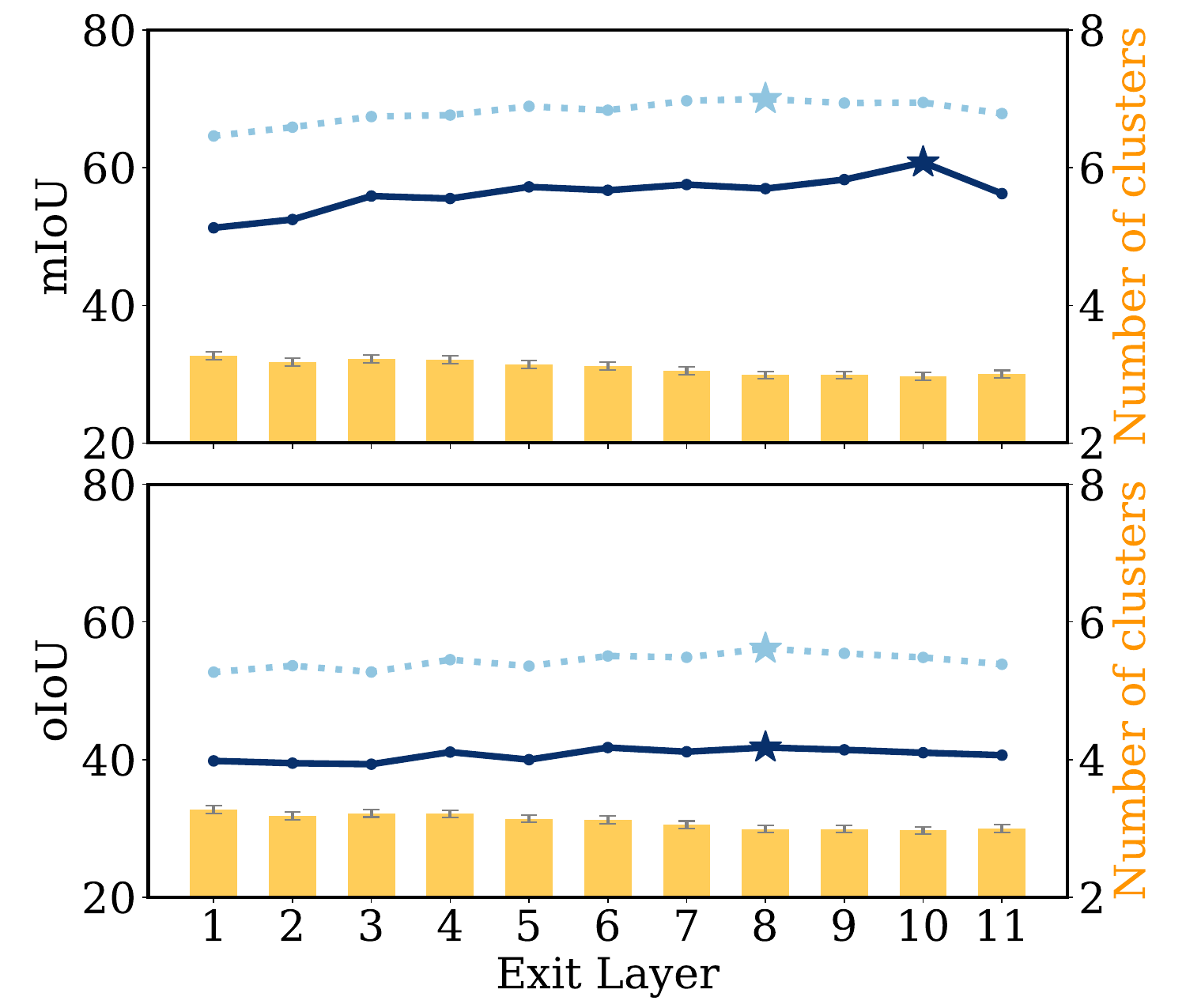}
        \caption{Exit Layer}
    \end{subfigure}
    \begin{subfigure}[t]{0.32\textwidth}
        \centering
        \includegraphics[width=\textwidth]{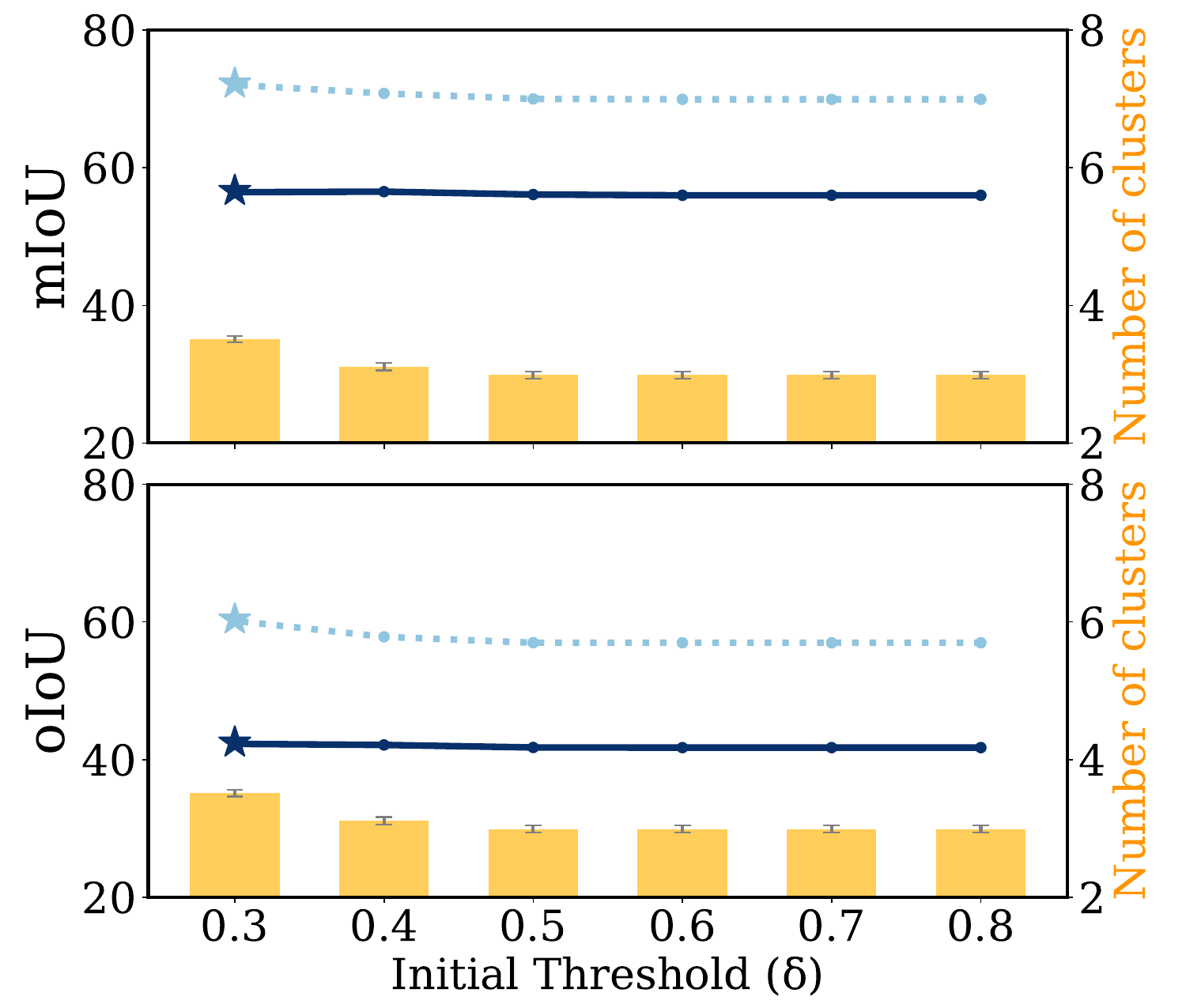}
        \caption{Threshold}
    \end{subfigure}
    \begin{subfigure}[t]{0.32\textwidth}
        \centering
        \includegraphics[width=\textwidth]{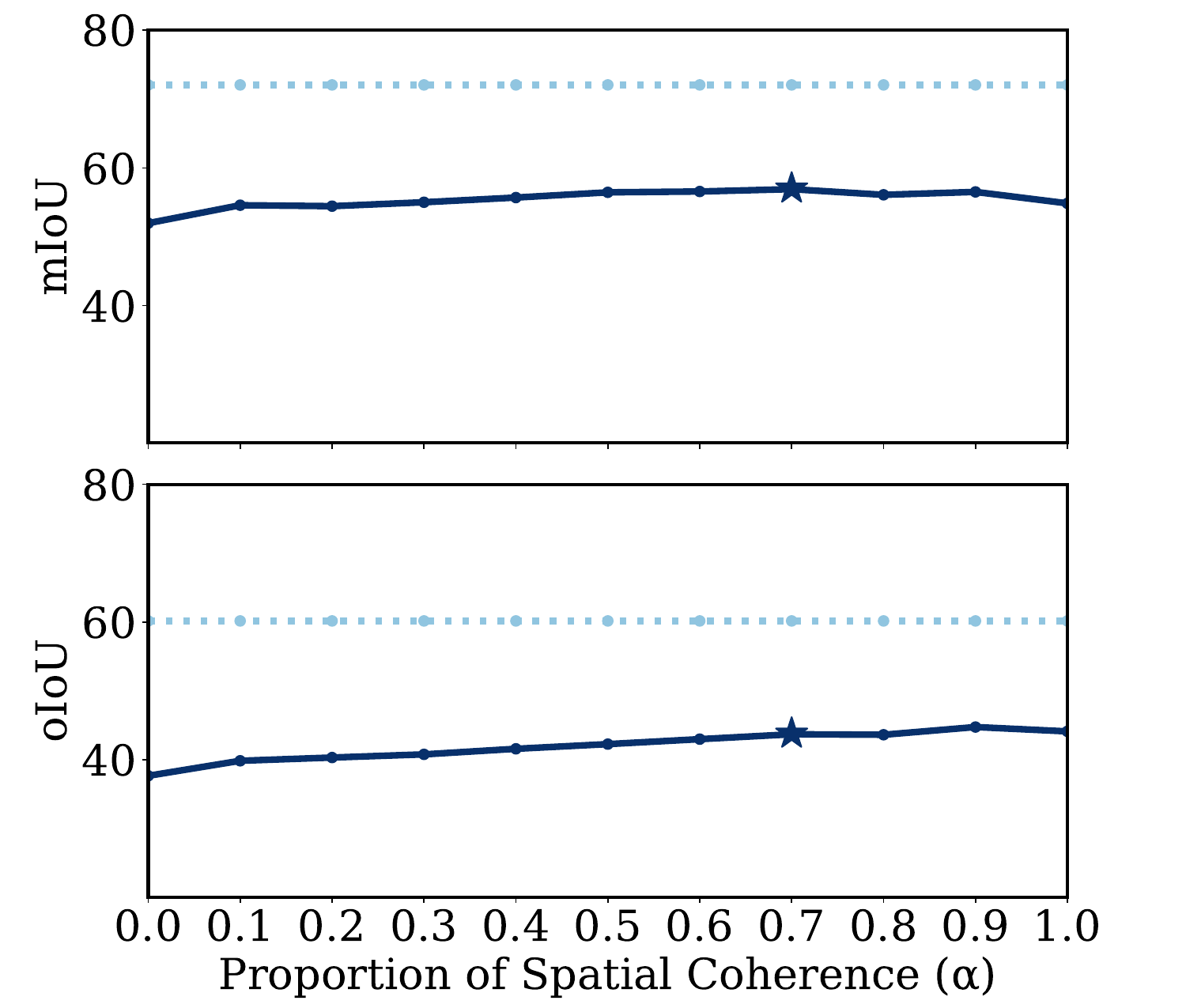}
        \caption{Spatial Coherence}
    \end{subfigure}

    \vspace{0.3em}
    \centerline{\small (a) ViT-B/16}

    \vspace{0.5em}

    \begin{subfigure}[t]{0.32\textwidth}
        \centering
        \includegraphics[width=\textwidth]{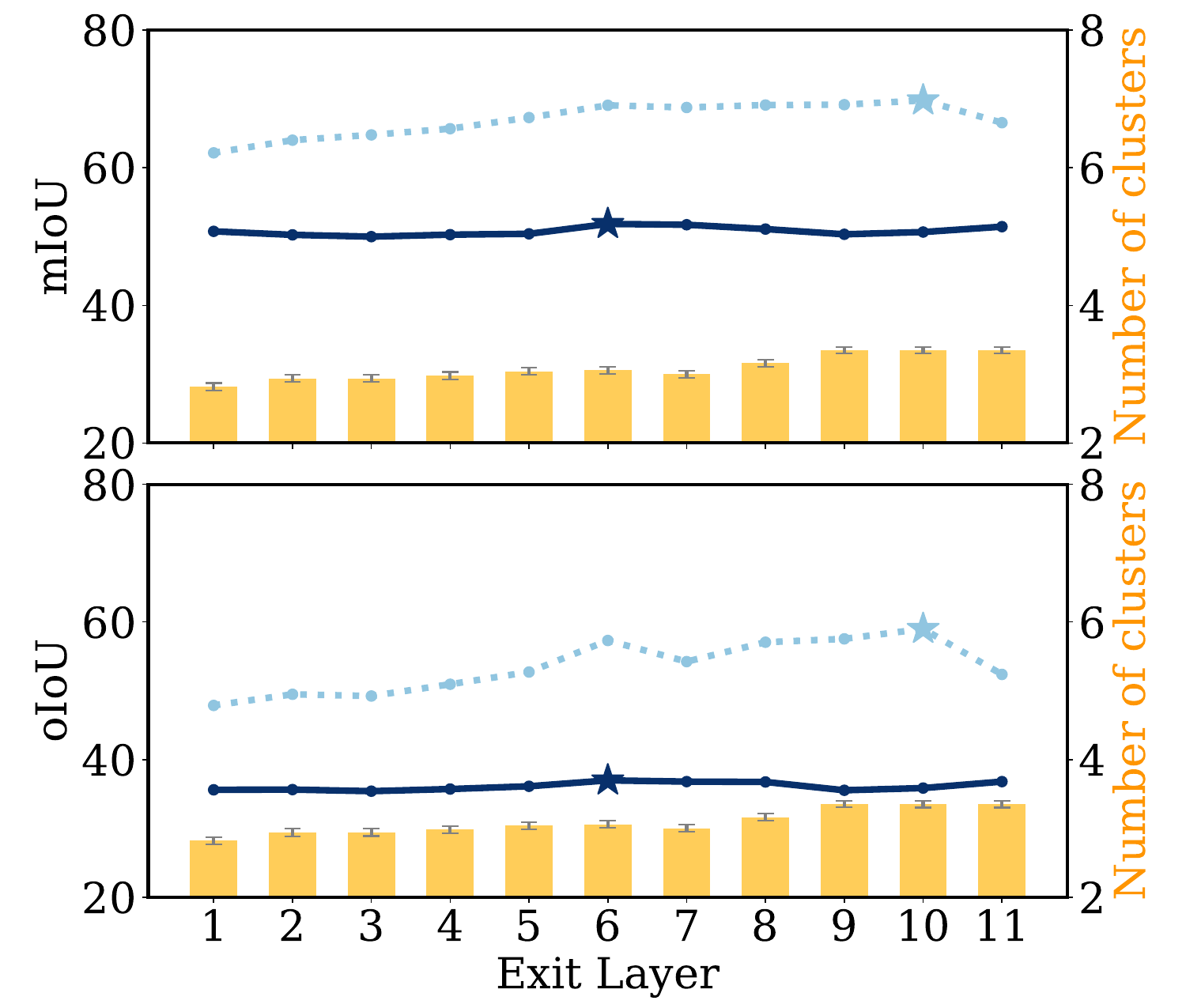}
        \caption{Exit Layer}
    \end{subfigure}
    \begin{subfigure}[t]{0.32\textwidth}
        \centering
        \includegraphics[width=\textwidth]{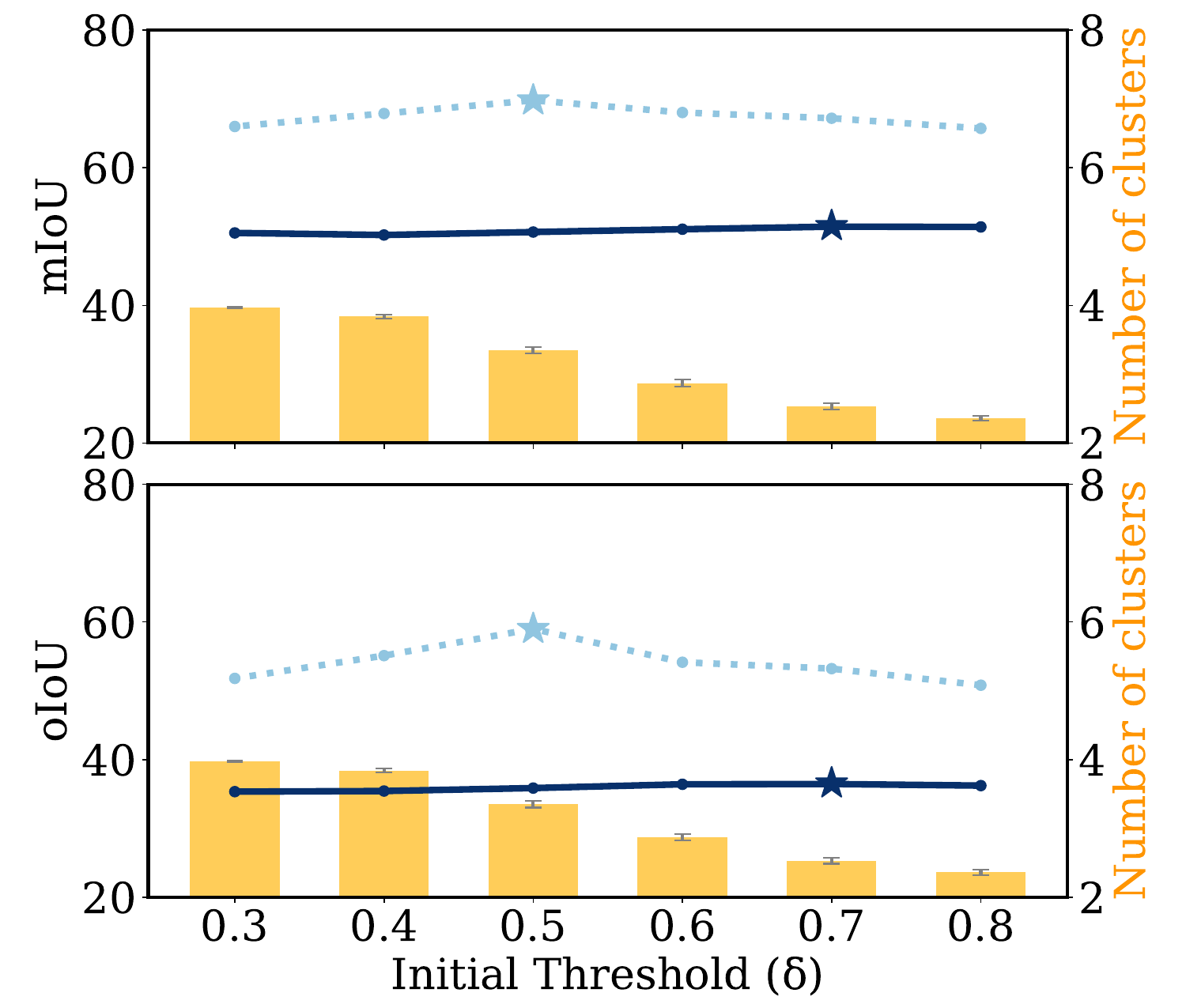}
        \caption{Threshold}
    \end{subfigure}
    \begin{subfigure}[t]{0.32\textwidth}
        \centering
        \includegraphics[width=\textwidth]{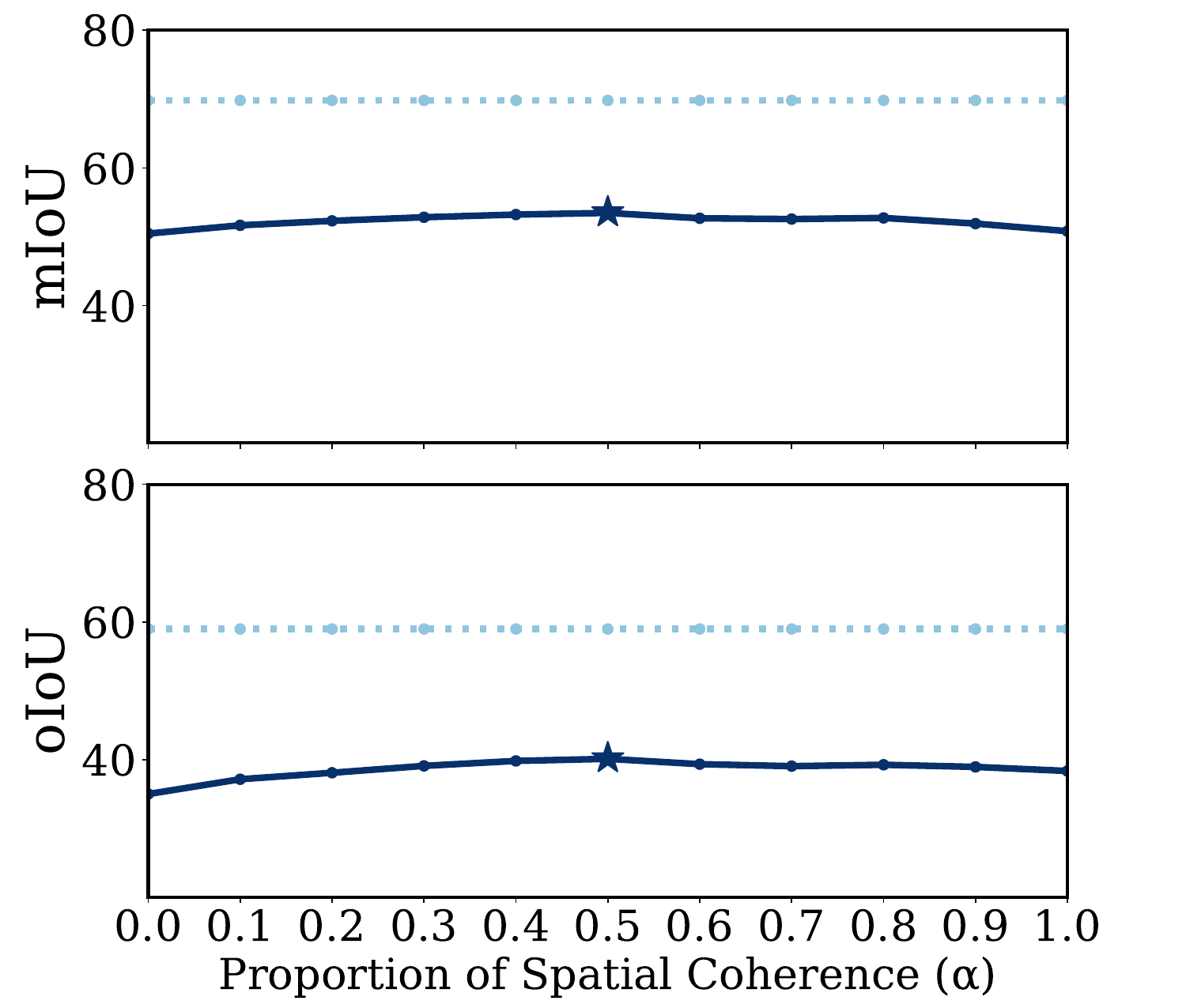}
        \caption{Spatial Coherence}
    \end{subfigure}

    \vspace{0.3em}
    \centerline{\small (b) ViT-B/32}

    \vspace{0.5em}

    \begin{subfigure}[t]{0.32\textwidth}
        \centering
        \includegraphics[width=\textwidth]{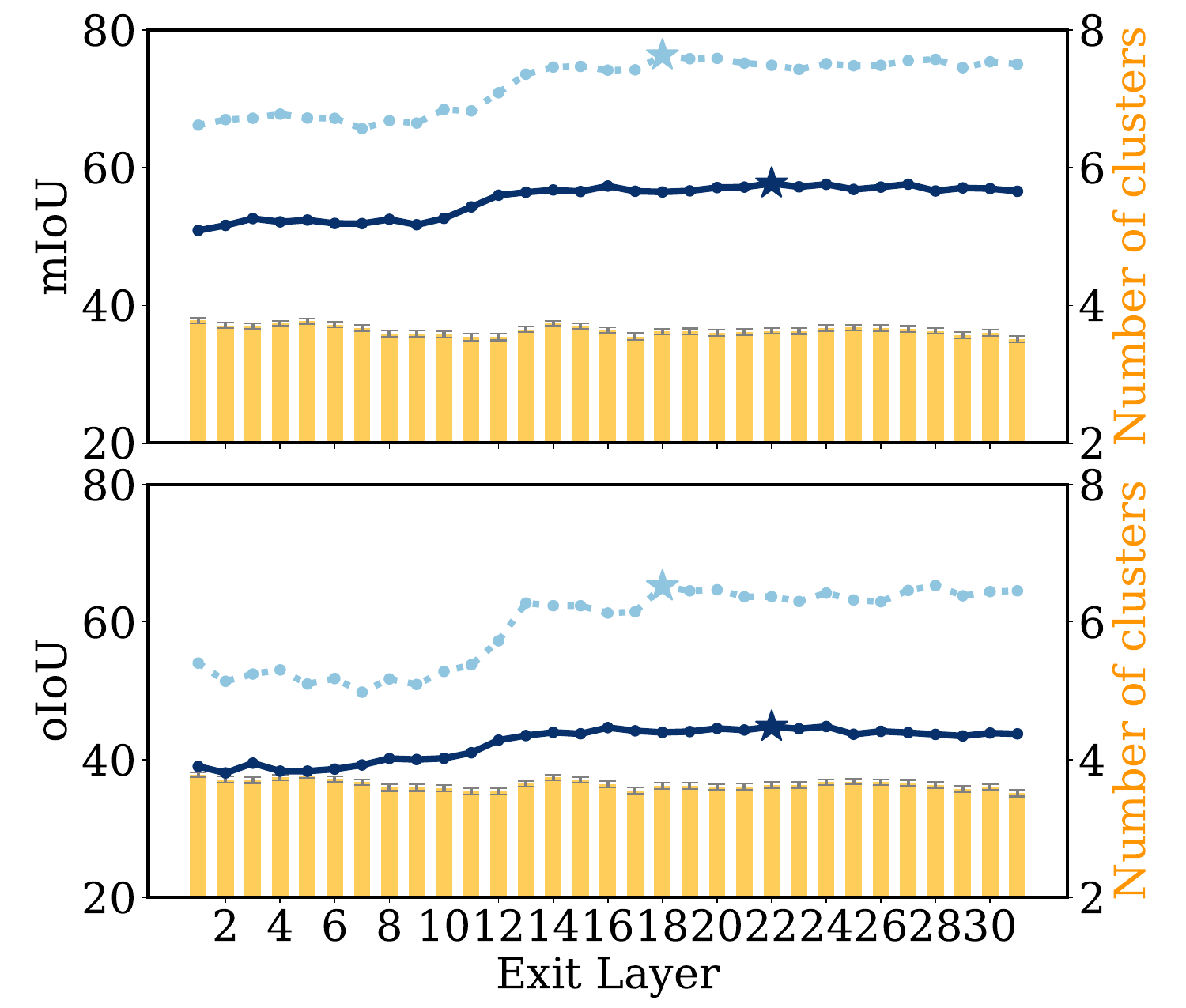}
        \caption{Exit Layer}
    \end{subfigure}
    \begin{subfigure}[t]{0.32\textwidth}
        \centering
        \includegraphics[width=\textwidth]{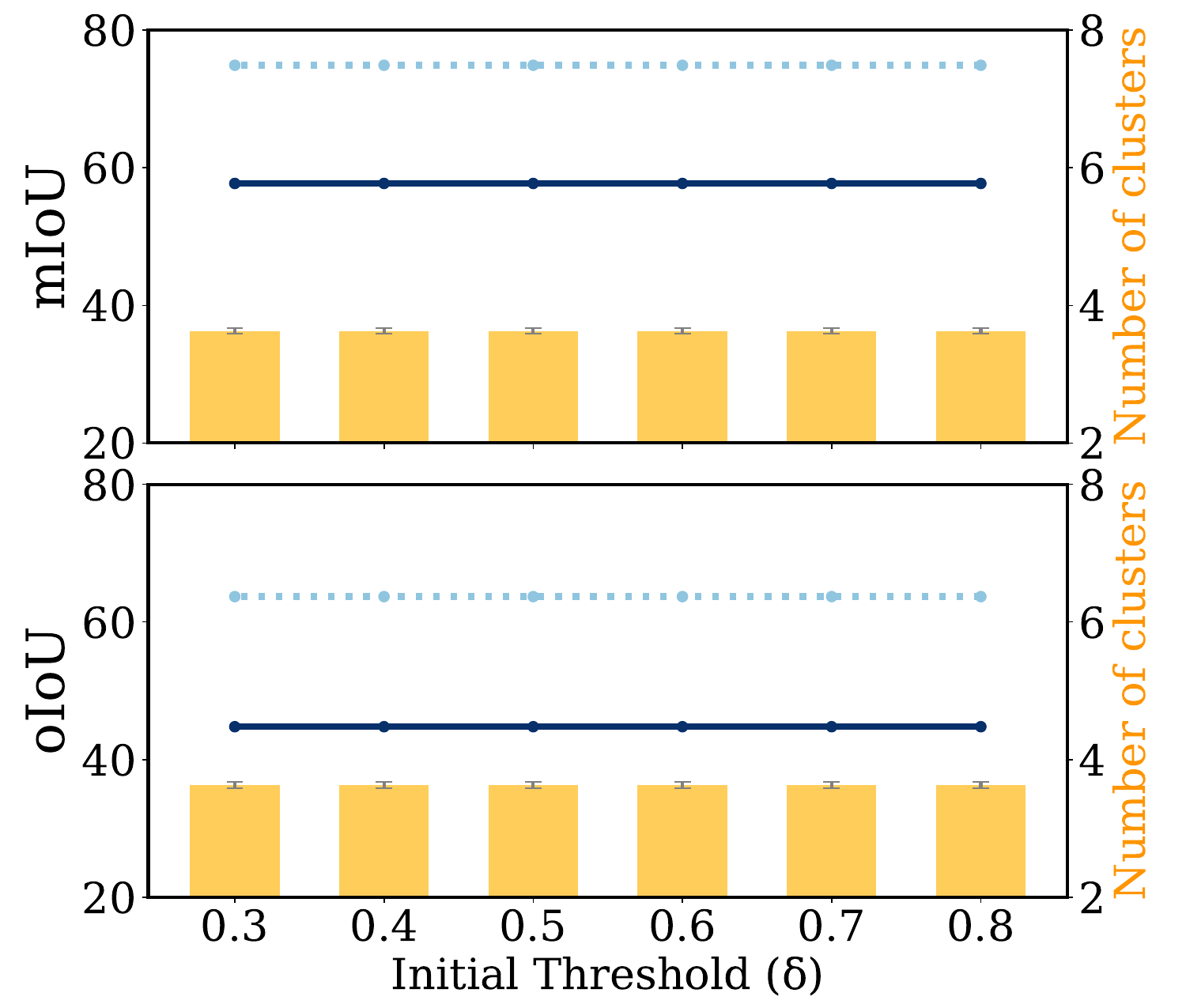}
        \caption{Threshold}
    \end{subfigure}
    \begin{subfigure}[t]{0.32\textwidth}
        \centering
        \includegraphics[width=\textwidth]{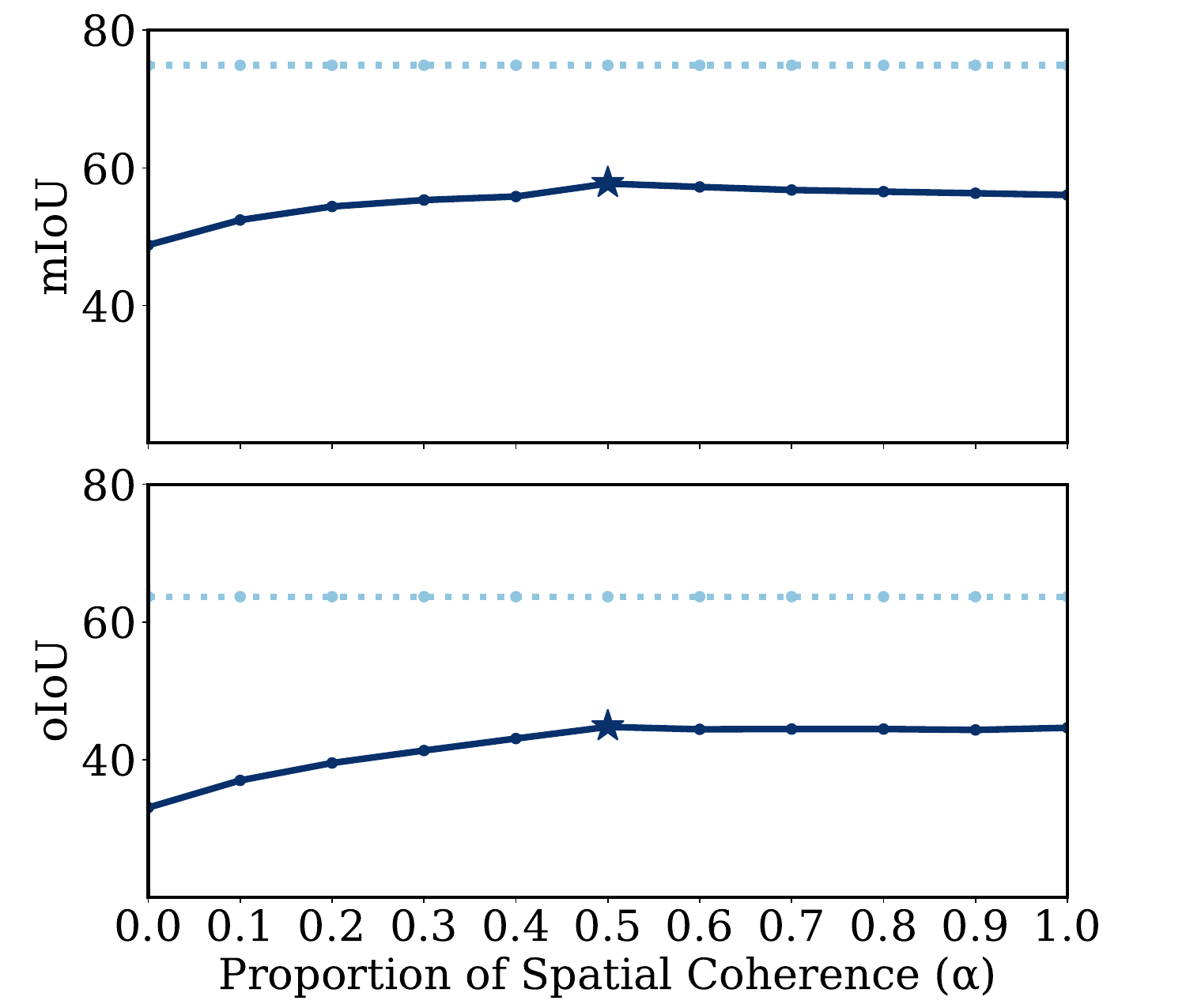}
        \caption{Spatial Coherence}
    \end{subfigure}

    \vspace{0.3em}
    \centerline{\small (c) DFN ViT-H/14}
    \vspace{-0.5em}

    \caption{
    \textbf{Hyperparameter tuning across different backbones using 10\% of the RefCOCOg val set.}
    Each row corresponds to a backbone (ViT-B/16, ViT-B/32, DFN ViT-H/14), and each column shows sensitivity to a specific hyperparameter (exit layer, threshold, spatial coherence).
    The \textcolor{topThreeBlue}{\textbf{deep-blue}} and \textcolor{topOneBlue}{\textbf{sky-blue}} lines denote \textcolor{topThreeBlue}{top-1} and \textcolor{topOneBlue}{top-3} performance, while the \textcolor{clusterGold}{\textbf{golden}} bars indicate the average number of clusters.
    Overall, performance remains stable across hyperparameter choices.
    }
    \label{fig:ablation_all}
    \vspace{-3em}
\end{figure*}

We show hyperparameter tuning results for \ours (CLIP ViT-B/16), \ours (CLIP ViT-B/32) and \ours (DFN ViT-H/14) in \cref{fig:ablation_all}. Regardless of model types, the IoU performances are not particularly sensitive to certain hyperparameter values, except for the exit layer. We select the exit layer ($l$) towards the last layer, yielding the best top-1 and top-3 performances. For instance, for \ours (CLIP ViT-B/32) and \ours (DFN ViT-H/14), we select the 10-\emph{th} ($>$ 6) and 22-\emph{th} ($>$ 18) exit layer with the highest top-3 and top-1 performances. We select the initial threshold ($\delta$) based on the top-3 performance for all three models. Lastly, since \texttt{P-Map} is not affected by the proportion of spatial coherence ($\alpha$), $\alpha$ is selected based on the top-1 performance. We set fixed $l$, $\delta$, and $\alpha$ for all evaluations ($l$: 10, 8, and 22, $\delta$: 0.5, 0.3, 0.5, and $\alpha$: 0.5, 0.7, and 0.5 for \ours (CLIP ViT-B/32, CLIP ViT-B/16, and DFN ViT-H/14). 

Note that the average number of clusters remains largely unaffected by the choice of exit layer. However, it intuitively decreases as the initial threshold increases, only for \ours (CLIP ViT-B/16) and \ours (CLIP ViT-B/32). \ours (DFN ViT-H/14) is possibly less influenced due to the increased patch number (\emph{see} visualizations in \cref{fig:maps}).

\begin{table*}[!t]
\centering
\caption{\textbf{Performance comparison between attention-based (\textsc{Attn}) and weighted-sum-based (\textsc{WS}) fusion strategies.} We observe that the best performance is achieved using exit layers from the middle layers that are closer to the final layers for both fusion methods.}
\vspace{-1em}
\resizebox{0.7\textwidth}{!}{
\begin{tabular}{lc c c c}
\toprule
\multirow{2}{*}{\textbf{Method}} & \textbf{Fusion} & 
\textbf{RefCOCOg} & 
\textbf{RefCOCO} & 
\textbf{RefCOCO+} \\
& \textbf{Layer} & oIoU / mIoU & oIoU / mIoU & oIoU / mIoU \\
\cmidrule(lr){1-1}\cmidrule(lr){2-2}\cmidrule(lr){3-3}\cmidrule(lr){4-4}\cmidrule(lr){5-5}
\textsc{Attn} & 1  & 31.65 / 42.07 & 34.57 / 41.05 & 36.32 / 41.89 \\
\textsc{Attn} & 2  & 33.33 / 43.73 & 36.11 / 43.96 & 36.69 / 42.77 \\
\textsc{Attn} & 3  & 24.80 / 35.20 & 30.03 / 37.37 & 30.29 / 35.33 \\
\textsc{Attn} & 4  & 31.36 / 43.89 & 34.29 / 41.39 & 34.01 / 39.71 \\
\textsc{Attn} & 5  & 33.13 / 45.45 & 34.92 / 42.64 & 33.80 / 38.52 \\
\textsc{Attn} & 6  & 35.47 / 49.12 & 36.24 / 44.01 & 35.82 / 41.65 \\
\textsc{Attn} & 7  & 38.70 / \textbf{51.22} & 36.99 / 44.67 & 33.80 / 39.57 \\
\textsc{Attn} & 8  & 38.65 / 51.05 & 35.62 / 45.15 & \textbf{36.09} / \textbf{44.65} \\
\textsc{Attn} & 9  & 31.94 / 48.17 & \textbf{39.56} / \textbf{48.36} & 35.42 / 42.77 \\
\textsc{Attn} & 10 & 35.37 / 50.52 & 35.35 / 44.05 & 35.56 / 43.32 \\
\textsc{Attn} & 11 & 35.85 / 51.09 & 31.09 / 41.53 & 34.50 / 42.07 \\
\rowcolor{gray!15} \textsc{WS} & -- & \textbf{39.07} / \textbf{51.22} & 36.46 / 45.35 & 35.63 / 43.73 \\
\bottomrule
\end{tabular}
}
\label{tab:fusion_results}
\vspace{-1.5em}
\end{table*}

\subsubsection{Alternative text feature fusion method.}\label{app:res_attn}
We construct hybrid text features based on a weighted sum of global- and local-level text features, following previous works \cite{yu2023zero, liu2025hybrid}. Here, we show results using an alternative fusion strategy of combining the features at an attention level. Specifically, instead of merging the text features at the final layer, we encode the sentence-level (global) and primary noun and context token-level (local) text features separately until the fusion layer and merge the attention outputs (\ie, $\text{fused}_{attn} = \gamma \text{global}_{attn} + (1-\gamma) \text{local}_{attn}$) before encoding them to construct the hybrid text features. Note that we use the fused features for the remaining layers after going through the fusion layer. As shown in \cref{tab:fusion_results} ($\gamma=0.5$), we find that there is a negligible difference between merging the text features at an attention-level and a feature-level.

\subsubsection{Exit layers of \oursfix.}\label{app:res_mid}
The choice of exit layer is critical, as it directly influences the overall quality of our spatial map, \ours. To analyze its impact, we evaluate performance across layers using the full dataset as well as 10\% subset. Specifically, we consider RefCOCOg (val), which contains long, complex referring expressions, and RefCOCO (val) and RefCOCO (testB) datasets, which feature shorter expressions with multiple objects of the same category per image~\cite{yang2022lavt}. Although the most optimal exit layer slightly differs across the datasets, the selected exit layers per dataset/domain are mostly the middle layers near the final layers (6--11), as can be shown in \cref{tab:mid_results}.

\begin{table*}[!t]
\centering
\caption{\textbf{Layer-wise performance trends on different types of datasets.} We observe that using middle-to-late layers near the final layer generally provides the most generalizable spatial and semantic information.}
\vspace{-0.5em}
\resizebox{0.8\textwidth}{!}{
\begin{tabular}{ccccc}
\toprule
\makecell[c]{\textbf{Exit} \\ \textbf{Layer}} &
\makecell[c]{\textbf{Full Refg (val)} \\ oIoU / mIoU} &
\makecell[c]{\textbf{10\% Refg (val)} \\ oIoU / mIoU} &
\makecell[c]{\textbf{10\% Ref (val)} \\ oIoU / mIoU} &
\makecell[c]{\textbf{10\% Ref (testB)} \\ oIoU / mIoU} \\
\cmidrule(lr){1-1}\cmidrule(lr){2-2}\cmidrule(lr){3-3}\cmidrule(lr){4-4}\cmidrule(lr){5-5}
1  & 36.00 / 47.20 & 37.60 / 51.59 & 35.64 / 43.97 & 28.46 / 36.89 \\
2  & 36.75 / 47.59 & 38.07 / 52.51 & 36.04 / 44.41 & 28.94 / 37.17 \\
3  & 37.02 / 47.95 & 38.20 / 52.94 & 34.01 / 42.96 & 29.48 / 38.45 \\
4  & 36.59 / 48.18 & 37.64 / 52.20 & 33.47 / 41.93 & 30.53 / 40.36 \\
5  & 37.17 / 48.60 & 37.84 / 52.50 & 32.64 / 40.57 & 31.15 / 41.00 \\
6  & 37.52 / 49.21 & 38.83 / 52.90 & 34.75 / 42.69 & \textbf{31.56} / \textbf{43.35} \\
7  & 38.45 / 50.62 & \textbf{38.90} / 52.97 & 37.02 / 45.39 & 31.37 / 42.47 \\
8  & \textbf{39.09} / 51.14 & 38.65 / 52.81 & 38.41 / \textbf{48.90} & 30.99 / 41.66 \\
9  & 38.77 / 51.02 & 37.85 / 52.62 & \textbf{38.96} / 48.05 & 30.45 / 40.63 \\
10 & 39.07 / \textbf{51.22} & 38.46 / \textbf{52.98} & 36.46 / 45.35 & 29.28 / 40.21 \\
11 & 39.63 / 50.61 & 37.81 / 52.09 & 34.89 / 44.30 & 31.09 / 41.53 \\
\bottomrule
\end{tabular}
}
\label{tab:mid_results}
\end{table*}

\subsection{Generalizability to non-CLIP-based VLE}\label{app:res_blip}
We use CLIP variants for a fair comparison with previous zero-shot RIS methods. However, to demonstrate the applicability of our method to other types of VLE, we show the results of applying our method to BLIP~\cite{li2022blip}. Whereas the final mask is selected based on the similarity between the original image and text embeddings for the Global-Local~\cite{yu2023zero} (+\texttt{CT}; Context Token) method, our method selects the best mask from the top candidates (\texttt{TC}) using the same \texttt{P-Map}-guided overlap reranking described in \cref{exp:setup} of the main paper. As can be observed in \cref{tab:blip_results}, our method consistently shows performance improvement compared to the Global-Local method.

\begin{table*}[!t]
\centering
\caption{\textbf{Performance of BLIP on RefCOCOg, RefCOCO, and RefCOCO+ datasets.} We show our method can be applicable to BLIP, achieving superior performance than the baseline - Global-Local (+\texttt{CT}).}
\vspace{-0.5em}
\resizebox{0.9\textwidth}{!}{
\begin{tabular}{cccccccccc}
\toprule
\multirow{2}{*}{\textbf{Method}} & \multicolumn{3}{c}{\textbf{RefCOCOg}} & \multicolumn{3}{c}{\textbf{RefCOCO}} & \multicolumn{3}{c}{\textbf{RefCOCO+}} \\
 & val (U) & test (U) & val (G) & val & testA & testB & val & testA & testB \\
\cmidrule(lr){1-1}\cmidrule(lr){2-4}\cmidrule(lr){5-7}\cmidrule(lr){8-10}
Global-Local (+\texttt{CT})   & 15.26 & 12.52 & 13.28 & 18.79 & 11.35 & 8.97  & 18.67 & 11.33 & 10.32 \\

 \rowcolor{gray!15} \oursfix      & 30.80 & 25.37 & 31.34 & 39.64 & 28.15 & 30.08 & 39.57 & 26.55 & 22.74 \\

\bottomrule
\end{tabular}
}
\label{tab:blip_results}
\vspace{-1em}
\end{table*}

\section{Multilingual Benchmark Details and Text-Side Analysis}\label{app:multi}

\subsection{Multilingual Evaluation Datasets and Benchmarks}
\label{app:rel_multi_data}

To rigorously evaluate the zero-shot multilingual spatial understanding of VLEs, we adopt a comprehensive 10-language evaluation framework. Our multilingual experiments are conducted on the translated referring-expression benchmark recently introduced by Nogueira \etal~\cite{nogueira2025comprehensionmultilingual}. The full resource systematically expands English-source RIS datasets into 10 languages (English, German, Dutch, Spanish, French, Italian, Korean, Portuguese, Russian, and Chinese) and contains approximately 8M multilingual referring expressions over 177,620 images.

\subsubsection{Dataset construction and semantic drift.}
The translation pipeline in \cite{nogueira2025comprehensionmultilingual} utilizes multilingual translation models, followed by reference-free quality estimation and visual-context enhancement. This construction is particularly useful for our setting because it keeps the underlying image and target annotation strictly fixed while varying only the query language. Such a design is suitable for diagnosing language-dependent geometric drift: If the visual target remains identical across translations, then cross-language performance differences can be directly attributed to the text representational interface. As noted in prior literature~\cite{nogueira2025comprehensionmultilingual}, such machine translation inevitably introduces \emph{translation-induced semantic drift}—where literal translations lose spatial nuances or introduce vocabulary biases not present in the English pretraining distribution. This creates the ultimate stress test for VLEs.

\begin{table}[!t]
\centering
\caption{\textbf{Translated RefCOCO-family benchmark used in this paper for multilingual evaluation.} The source multilingual REC resource of Nogueira \etal~\cite{nogueira2025comprehensionmultilingual} spans 12 English-origin datasets and 10 languages. For direct comparability with zero-shot RIS literature, we restrict our multilingual evaluation to the translated RefCOCO family, which yields nine canonical evaluation splits in total.}
\label{tab:translated_refcoco_family}
\resizebox{0.77\linewidth}{!}{
\begin{tabular}{c c c c}
\toprule
\textbf{Dataset} & \textbf{Splits} & \textbf{\# Lang.} & \textbf{Task} \\
\cmidrule(lr){1-1}\cmidrule(lr){2-2}\cmidrule(lr){3-3}\cmidrule(lr){4-4}
RefCOCOg & val (U), test (U), val (G) & 10 & Multilingual RIS \& Retrieval \\
RefCOCO  & val, testA, testB          & 10 & Multilingual RIS\& Retrieval \\
RefCOCO+ & val, testA, testB          & 10 & Multilingual RIS\& Retrieval \\
\bottomrule
\end{tabular}
}
\vspace{-1em}
\end{table}

\subsubsection{Dataset characteristics.}
Although the source benchmark spans 12 datasets, we restrict our multilingual evaluation to the translated RefCOCO family (\cref{tab:translated_refcoco_family}) to preserve direct comparability with the English zero-shot RIS setting. Furthermore, these datasets provide complementary linguistic properties crucial for our analysis:
\begin{itemize}
    \item \textbf{RefCOCO:} Contains short, highly spatial, and directional expressions (\eg, ``left,'' ``bottom right''). This tests absolute localization capabilities across languages.
    \item \textbf{RefCOCO+:} Strictly excludes spatial words, focusing entirely on appearance-based attributes (\eg, clothing, colors). This tests whether attribute binding survives cross-lingual translation.
    \item \textbf{RefCOCOg:} Comprises significantly longer, complex, and compositional descriptive sentences. This dataset inherently accumulates the most language-specific geometric drift during the forward pass, making it the most challenging benchmark.
\end{itemize}
Our evaluation is decomposed into nine standard evaluation splits across 10 languages, creating 90 distinct target-language/split cases. This split-level organization exposes differences between spatially explicit, attribute-centric, and long-form compositional settings much more clearly than a single dataset-level aggregate.

\subsection{Detailed Evaluation Protocols}
\label{app:rel_multi_eval}

\subsubsection{Parallel-query test-time setting.}
An important detail is that \ours operates in a \emph{parallel-query} multilingual
test-time regime. For a given target instance, the benchmark provides multiple
translated expressions that refer to the same object. We use these parallel expressions
to probe the frozen text encoder, identify the most language-stable intermediate layer,
and construct the multilingual centroid. No gradient update or adaptation of the encoder
parameters is performed.

\subsubsection{Fair comparison to baseline and final-layer fusion.}
The plain SigLIP2 baseline uses only the target-language query and encodes it with the
standard final-layer text representation. By contrast, the cross-lingual fusion-based
controls, \texttt{Final-All} and \ours, are both given the same set of parallel
translations for the same instance. \texttt{Final-All} mixes them only after the final
text representation has been formed, whereas \ours probes the parallel translations to
select the language-stable layer $l_{\mathrm{txt}}^{*}$ via cross-language consistency,
injects the multilingual centroid at that layer, and then continues the remaining forward
pass. This makes the comparison well-controlled: When multilingual evidence is used, the
difference is \emph{where} it is injected rather than \emph{whether} extra linguistic evidence is available.

\subsubsection{Multilingual zero-shot RIS setting.}
For the end-to-end segmentation task, we utilize an off-the-shelf
Mask2Former~\cite{cheng2022masked} to generate class-agnostic candidate masks. The
translated text queries are processed through the VLE. Instead of relying on the
final-layer global similarity, we extract the patch-level image features at the optimal
intermediate visual layer and the context-aware hybrid text features to generate our
\texttt{P-Map} via negative cosine similarity. The candidate masks are then reranked
based on their overlap with the \texttt{P-Map} activations. Following the main paper,
we report multilingual RIS performance primarily using mIoU and IoU@50. All models are
evaluated under the exact same candidate mask proposals.

\subsubsection{Multilingual text-to-image retrieval setting.}
To validate that the multilingual correction is not specific to mask reranking, we
additionally evaluate zero-shot multilingual text-to-image retrieval using the same
translated RefCOCO family. The most important design choice here is that the
\emph{image branch is held fixed}. Any retrieval gain can therefore be attributed to
improved text-side geometry rather than to changes in the visual representation. The
main paper reports Recall@1, while the appendix additionally analyzes Recall@5 and
Recall@10. Across the nine translated RefCOCO-family splits and 10 target languages,
this yields 90 target-language/split cases in the appendix diagnostics.

\subsubsection{Implementation and latency conventions.}
All multilingual experiments are conducted with frozen VLEs. The multilingual overhead
comes from probing intermediate text layers across the parallel translations and
aggregating the resulting centroid. For the latency analysis, we measure the end-to-end
per-sample runtime of the multilingual text-side pipeline, so that the reported overhead
reflects the actual deployment cost of the proposed stabilization step.

\subsection{Inference Latency Experimental Setup}
We measure inference latency per image--query pair using \verb|time.perf_counter|,
synchronizing CUDA with \verb|torch.cuda.synchronize()| before and after each forward
pass. The runtime includes text preprocessing, tokenization, and embedding with the
SigLIP2 text encoder, cross-language consistency computation, optimal mid-layer
selection (for \ours), and centroid aggregation. Experiments use a batch size of 1 and
report the average runtime per dataset split. As can be seen in \cref{tab:latency},
compared to the plain final-layer baseline and the \texttt{Final-All} control, our
method introduces a modest additional overhead due to mid-layer probing and centroid
aggregation.

\begin{table*}[!t]
\centering
\caption{\textbf{Full inference latency results.} Compared to the final-layer baseline
and the \texttt{Final-All} control, \ours introduces a modest additional inference
overhead due to mid-layer probing and centroid aggregation.}
\vspace{-0.5em}
\label{app:inference_latency}
\small
\resizebox{\textwidth}{!}{%
\begin{tabular}{lccccccccc}
\toprule
\multicolumn{1}{c}{\multirow{2}{*}{\textbf{Method}}} & \multicolumn{3}{c}{\textbf{RefCOCOg}} & \multicolumn{3}{c}{\textbf{RefCOCO}} & \multicolumn{3}{c}{\textbf{RefCOCO+}} \\
 & val (U) & test (U) & val (G) & val & testA & testB & val & testA & testB \\
\cmidrule(lr){1-1}\cmidrule(lr){2-4}\cmidrule(lr){5-7}\cmidrule(lr){8-10}
SigLIP2
& 0.3307 & 0.3245 & 0.3151 & 0.3138 & 0.3134 & 0.3211 & 0.3739 & 0.3931 & 0.3641 \\

SigLIP2 (Final Layer) 
& 0.3310 & 0.3257 & 0.3130 & 0.3144 & 0.3100 & 0.3216 & 0.3782 & 0.3864 & 0.3631 \\

\rowcolor{gray!15} \oursdynperm (Mid-Layer)
& 0.5267 & 0.4997 & 0.4868 & 0.5084 & 0.5070 & 0.5364 & 0.5952 & 0.6203 & 0.5800 \\
\bottomrule
\end{tabular}
\label{tab:latency}
}
\vspace{-1em}
\end{table*}

\subsection{Ablation on Zero-shot Multilingual Text-to-Image Retrieval}\label{app:res_multi_retr}

To further understand the behavior of our multilingual stabilization module, we conduct a comprehensive text-side ablation study across RefCOCO, RefCOCO+, and RefCOCOg. We compare the proposed mid-layer per-sample centroid injection against final-layer interpolation baselines and global-centroid variants while keeping the \emph{image features fixed}. This setting isolates the \emph{text-side} effect particularly cleanly, because only the text representation is modified and the image branch is held constant throughout. Accordingly, the results in this subsection should be read as \emph{diagnostic evidence} for representational geometry, rather than as end-to-end headline numbers directly comparable to the full multilingual grounding results in the main paper.

For readability, we complement the exhaustive numerical tables with four summary figures generated from the same outputs used in the raw ablation tables. Unless otherwise stated, these figures use \texttt{case\_mean} averaging over all 90 target-language/split cases, so that each case contributes equally regardless of split size. This prevents larger splits from dominating the interpretation and makes the robustness pattern across languages and datasets easier to read at a glance.

\paragraph{Reading guide and summary convention.}

\subsubsection{Compared methods.}
Throughout this subsection, we use the following shorthand:
\begin{itemize}
    \item \textbf{\texttt{Baseline (Final)}}: the original final-layer text embedding without any centroid fusion.
    \item \textbf{\texttt{Final-All}}: interpolation at the \emph{final} text embedding space using a single centroid averaged over all available languages.
    \item \textbf{\texttt{Mid-All}}: injection at a selected \emph{mid-layer} using one global centroid averaged over all samples and all languages.
    \item \textbf{\texttt{Mid-Lang}}: injection at a selected \emph{mid-layer} using one global centroid computed from a \emph{single anchor language}.
    \item \textbf{\texttt{Mid-Sample}} (\emph{ours}): \ours text-side stabilization module, which selects a language-stable mid-layer via $\Phi$ and injects a \emph{per-sample multilingual centroid} computed from all translations of the same instance.
\end{itemize}

\paragraph{How to read the figures.}
\begin{itemize}
    \item $\Delta$R@1 denotes the performance change relative to \texttt{Baseline (Final)}.
    \item \textbf{Sign counts} are reported as \texttt{(+ / 0 / -)}, indicating how many cases improve, stay unchanged, or degrade.
    \item \textbf{Case-mean} treats every target-language/split case equally; this is useful for diagnosing robustness rather than reporting a size-weighted dataset average.
    \item In the anchor heatmap, \textbf{ALL-Global} corresponds to \texttt{Mid-All}, whereas \textbf{SampleMean} corresponds to \texttt{Mid-Sample}.
\end{itemize}

\paragraph{Visual summary of the main text-side findings.}

\vspace{1em}
\begin{table}[!t]
\centering
\caption{\textbf{Translation quality \emph{vs.} representation mixing.} Retrieval performance (Recall@1) on the RefCOCOg validation set using Korean (KO) queries. Enhancing the text quality via LLM translations or detailed definitions fails to improve the baseline, whereas representation-level mixing provides significant gains.}
\vspace{-0.5em}
\label{tab:text_quality_ablation}
\resizebox{0.6\linewidth}{!}{
\begin{tabular}{lc}
\toprule
\textbf{Query Type / Method} & \textbf{Recall@1 (\%)} \\
\cmidrule(lr){1-1}\cmidrule(lr){2-2}
Baseline (Original KO) & 34.81 \\
\cmidrule(lr){1-1}\cmidrule(lr){2-2}
LLM Translation (EN $\rightarrow$ KO) & 35.04 \\
LLM Definition (EN $\rightarrow$ KO \emph{def}) & 33.99 \\
LLM Description (EN $\rightarrow$ KO \emph{desc join}) & 32.25 \\
\cmidrule(lr){1-1}\cmidrule(lr){2-2}
Translation (KO $\rightarrow$ EN) & 41.68 \\
\rowcolor{gray!10} \textbf{Representation Mix (Ours, $\alpha=0.5$)} & \textbf{44.35} \\
\bottomrule
\end{tabular}
}
\vspace{-1em}
\end{table}

\subsubsection{Translation quality \emph{vs.} geometric drift.}
A natural question is whether the multilingual degradation comes merely from poor translation quality or unnatural phrasing. However, \cref{tab:text_quality_ablation} already shows that stronger surface-form manipulations, including LLM-refined translations and expanded definitions, do not recover the gap and can even reduce retrieval accuracy. In contrast, representation-level stabilization improves performance substantially. This indicates that the bottleneck is not primarily lexical naturalness, but the \emph{language-dependent geometric drift} that emerges inside the shared multimodal space. The figure-level summaries below make this conclusion more explicit.

\begin{figure*}[!t]
\centering
\includegraphics[width=\textwidth]{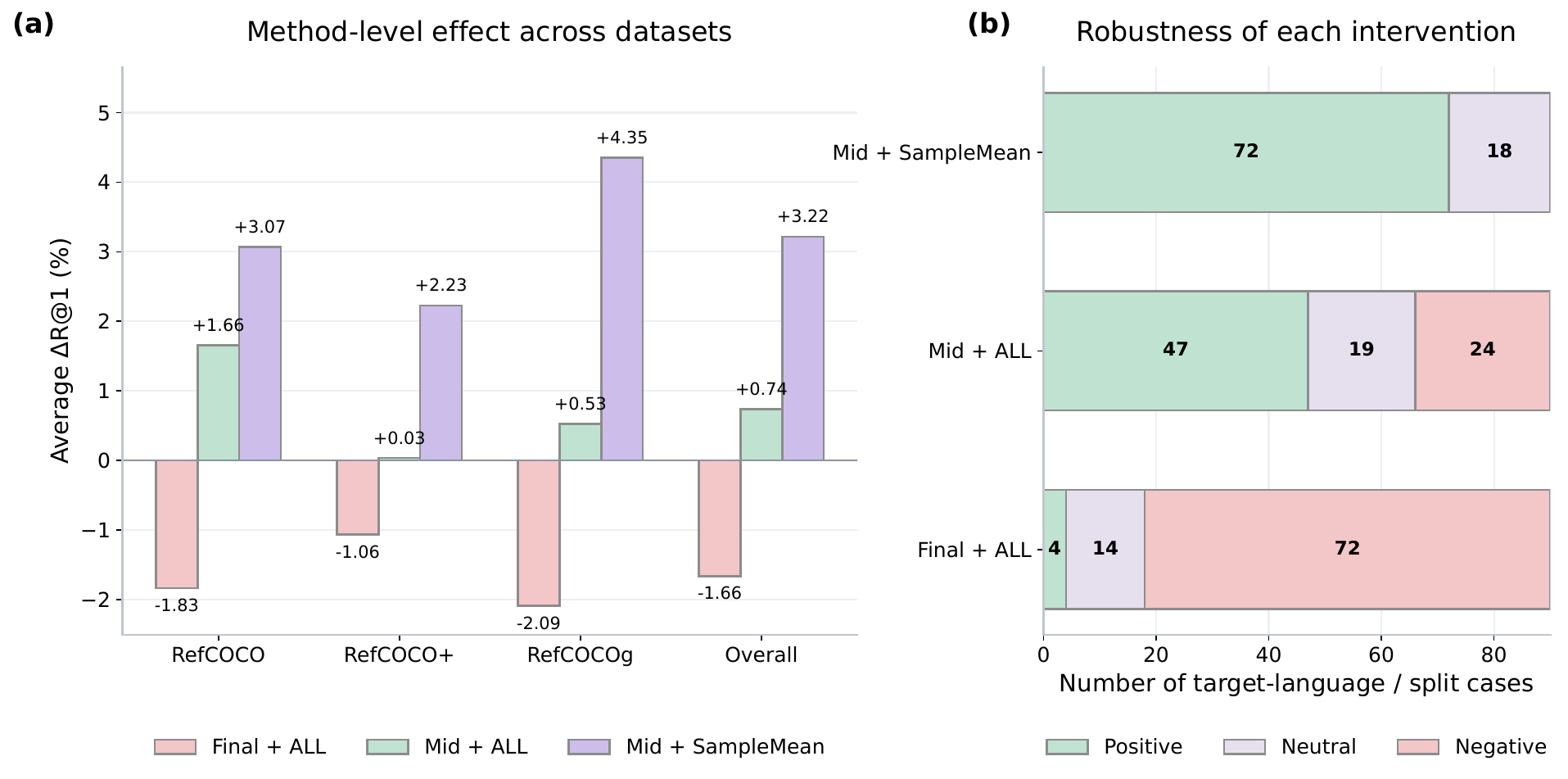}
\caption{\textbf{Case-mean visual summary of the main text-side ablation.}
Summary over 90 target-language/split cases with fixed image features. Left: mean $\Delta$R@1 by dataset and overall. Right: sign counts \texttt{(+ / 0 / -)}. The key pattern is that \texttt{Final-All} is broadly harmful, \texttt{Mid-All} is only mildly helpful, and \texttt{Mid-Sample} is both the strongest and the most consistent variant.}
\label{fig:app_multi_summary}
\end{figure*}

\subsubsection{Stage matters more than simply adding a centroid.}
The first and most important result is summarized in \cref{fig:app_multi_summary}. Averaged across all cases, \texttt{Final-All} yields an overall drop of $-1.66$ R@1, whereas \texttt{Mid-All} achieves a modest gain of $+0.74$, and \texttt{Mid-Sample} achieves the largest gain of $+3.22$. The sign pattern is even more revealing: \texttt{Final-All} improves only 4 cases and hurts 72, while \texttt{Mid-Sample} improves 72 cases and does not hurt a single case (\texttt{72 / 18 / 0}). This sharp contrast shows that the key issue is not whether a centroid is introduced, but \emph{where} the correction is applied. Once the representation has already passed through the final alignment interface, linear interpolation is usually too late and tends to destroy discriminability. By contrast, correcting the representation \emph{before} the final contextualization steps is consistently beneficial.

\subsubsection{Global \emph{vs.} instance-level context.}
\Cref{fig:app_multi_summary} also isolates the effect of centroid granularity. \texttt{Mid-All} and \texttt{Mid-Sample} both intervene at the same stage, but they differ in whether the injected signal is a single universal centroid or an instance-conditioned multilingual mean. The gap between them is substantial ($+0.74$ vs.\ $+3.22$ overall), showing that the gain is not explained by simply pulling every query toward a universal multilingual center. Instead, the strong result of \texttt{Mid-Sample} suggests that the model benefits from a correction that preserves the semantics of the \emph{current expression} while neutralizing language-specific drift.

\begin{figure*}[t]
\centering
\includegraphics[width=\textwidth]{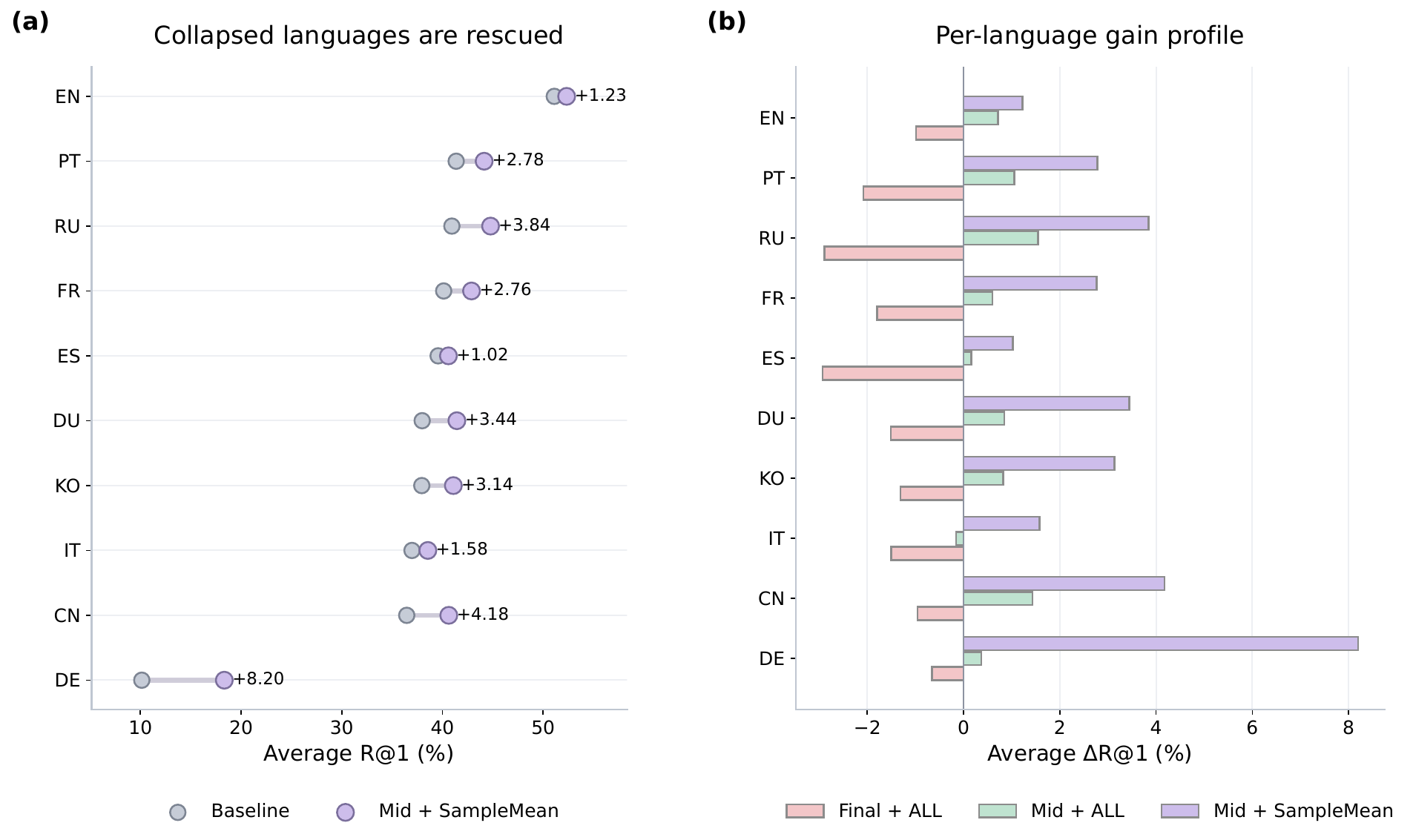}
\caption{\textbf{Language-wise rescue effect of \texttt{Mid-Sample}.}
Left: dumbbell plot comparing \texttt{Baseline (Final)} and \texttt{Mid-Sample} R@1 for each target language. Right: mean $\Delta$R@1 by target language. The largest gains occur for the weakest languages, while English is still improved on average, indicating that multilingual stabilization does not require sacrificing the dominant language.}
\label{fig:app_multi_rescue}
\end{figure*}

\subsubsection{Disproportionate rescue of collapsed languages.}
\Cref{fig:app_multi_rescue} makes clear that \texttt{Mid-Sample} acts primarily as a rescue mechanism for weak languages. German (DE) is the clearest example: It starts from a very low baseline R@1 of $10.14$ and gains $+8.20$ on average. Chinese (CN) and Russian (RU) also show large improvements of $+4.18$ and $+3.84$, respectively. Crucially, English (EN) still improves by $+1.23$ on average. This is important for the interpretation of multilingual robustness: Our method is not improving weak languages by trading away strong ones, but instead reduces a structural bias that harms the most fragile languages the most. Put differently, the method behaves more like a \emph{targeted geometric regularizer} than a language-balancing compromise.

\subsubsection{Improvement is largest where the baseline is weakest.}
A second pattern visible in \cref{fig:app_multi_rescue} is the strong negative association between the baseline language performance and the gain after stabilization. In our case-mean summary, languages with a weaker baseline tend to receive substantially larger gains, with a strong negative correlation between baseline R@1 and improvement ($r \approx -0.88$). Moreover, the dispersion across languages is reduced after stabilization, indicating that \texttt{Mid-Sample} improves not only the mean accuracy but also the \emph{cross-language stability} of the shared representation.

\subsubsection{Sensitivity to linguistic complexity.}
The dataset-wise bars in \cref{fig:app_multi_summary} further support the architectural interpretation. The largest gains are observed on RefCOCOg ($+4.35$), followed by RefCOCO ($+3.07$) and RefCOCO+ ($+2.23$). This ranking is informative: Longer and more compositional expressions accumulate language-specific drift more severely during the forward pass, so they benefit the most from stabilization before the final abstraction bottleneck. The result suggests that our method is particularly well matched to the complex expressions that are most vulnerable to multilingual fragmentation.

\begin{figure*}[t]
\centering
\includegraphics[width=\textwidth]{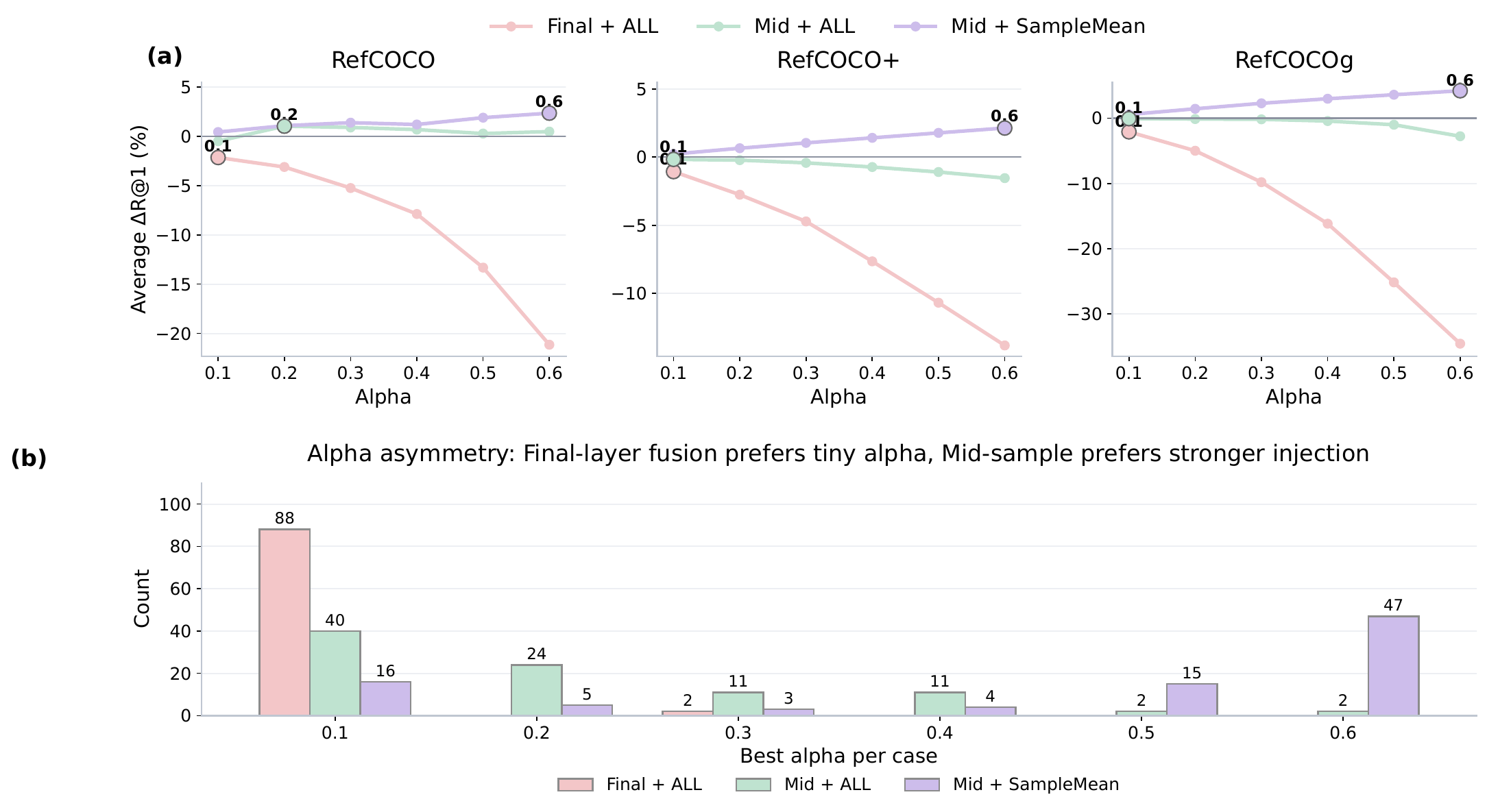}
\caption{\textbf{Alpha-asymmetry diagnostic.}
Left: fixed-$\alpha$ case-mean curves for representative methods across datasets. Right: histogram of the best $\alpha$ selected per case. \texttt{Final-All} only tolerates very small fusion weights, whereas \texttt{Mid-Sample} typically prefers strong injection, indicating that aggressive mixing is harmful in the final space but beneficial in the mid-layer regime.}
\vspace{-1em}
\label{fig:app_multi_alpha}
\end{figure*}

\subsubsection{The $\alpha$-asymmetry diagnostic.}
The interpolation-weight analysis in \cref{fig:app_multi_alpha} provides a second, highly mechanistic diagnostic. For \texttt{Final-All}, the best setting is almost always the smallest available fusion weight ($\alpha=0.1$ dominates the histogram), and the fixed-$\alpha$ curves deteriorate sharply as $\alpha$ increases. By contrast, \texttt{Mid-Sample} generally prefers much stronger injection, with the histogram peaking near $\alpha=0.6$. This asymmetry is highly informative. It shows that strong centroid mixing is destructive once the representation has already reached the biased final alignment space, but acts as a powerful regularizer when applied at an earlier, more language-stable stage. In this sense, $\alpha$ is not merely a tuning parameter; it functions as a probe that distinguishes whether the intervention is applied in a fragile or recoverable region of the text encoder.

\vspace{+1em}
\subsubsection{Why \texttt{mid-all} is not enough.}
\Cref{fig:app_multi_alpha} also helps explain why \texttt{Mid-All} remains noticeably weaker than \texttt{Mid-Sample}. Although \texttt{Mid-All} sometimes yields positive results, its gain is far less stable across datasets and $\alpha$ values. This implies that ``going mid-layer'' alone is insufficient. The representation still needs an \emph{instance-aware} corrective signal. A universal centroid can partially denoise language drift, but it cannot preserve the fine-grained semantics of the current query as effectively as the sample-wise multilingual mean.

\paragraph{Anchor synergy.}
\begin{figure*}[t]
\centering
\includegraphics[width=\textwidth]{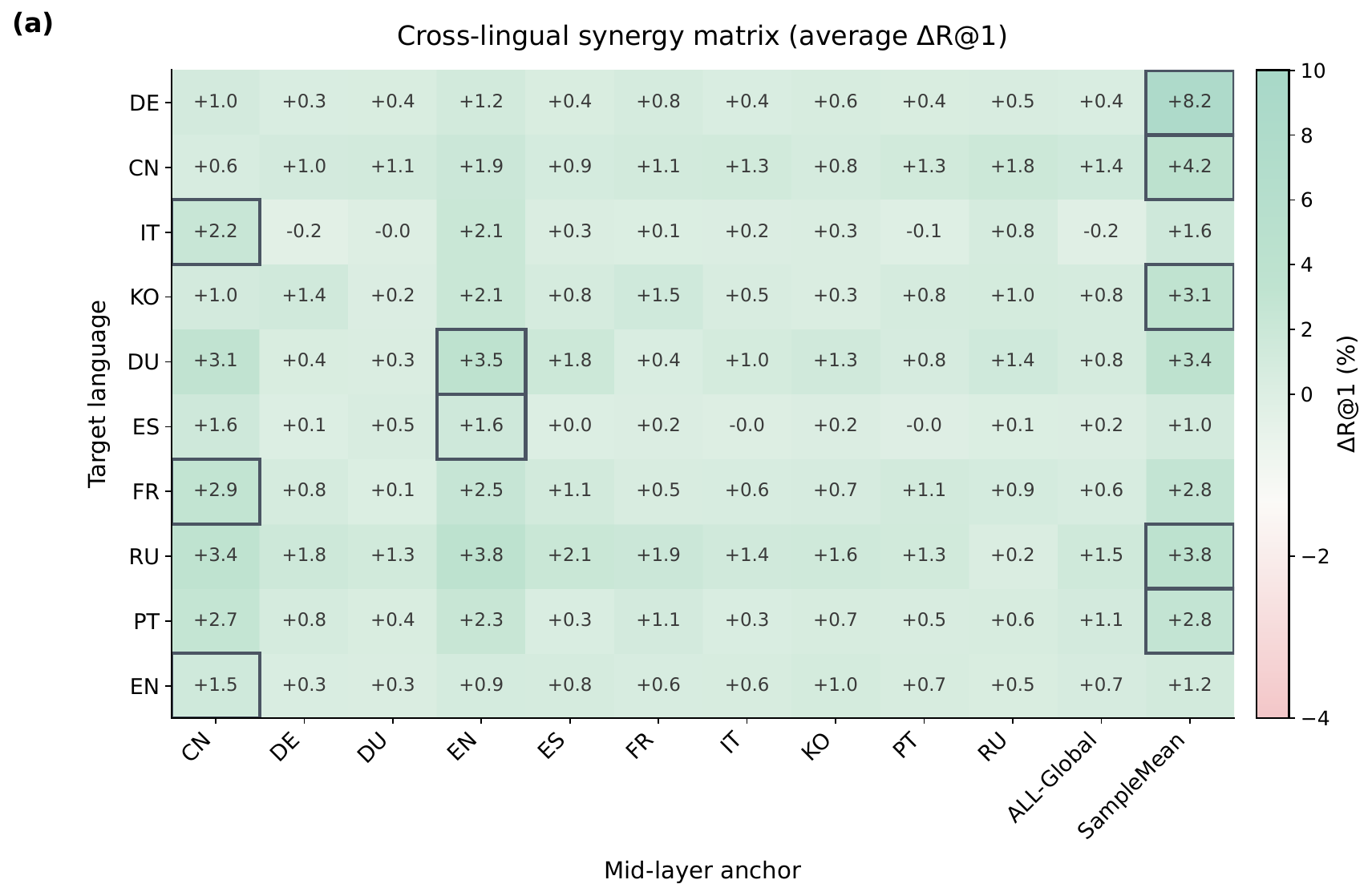}
\caption{\textbf{Cross-lingual anchor synergy at the mid-layer.}
Heatmap of case-mean $\Delta$R@1 for \texttt{Mid-Lang}, organized by target language (rows) and anchor language (columns), with additional columns for \texttt{ALL-Global} and \texttt{SampleMean}. The pattern shows that self-anchor is rarely optimal, no single donor language is universally best, and \texttt{SampleMean} is the most reliable overall strategy.}
\label{fig:app_multi_anchor}
\vspace{-1em}
\end{figure*}

\subsubsection{Stage matters more than anchor identity.}
The anchor-sweep summary in \cref{fig:app_multi_anchor} visually complements the exhaustive matrices reported later in \cref{app:res_multi_matrix}. A striking result is that the best anchor is often \emph{not} the target language itself. In many cases, a self-anchor provides only a small benefit, whereas a different anchor language performs noticeably better. This means that the improvement is not driven by surface-form matching or language identity. Rather, it comes from shifting the query toward a more language-stable region of the shared representation space.

\subsubsection{Donor-recoverable \emph{vs.} consensus-requiring languages.}
\Cref{fig:app_multi_anchor} also reveals that target languages do not all fail in the same way. Some targets are already well rescued by a single strong donor language, while others benefit only marginally from any individual donor and improve substantially only under \texttt{SampleMean}. German (DE) is a representative example of the latter regime: Single-language anchors help only weakly, but the multilingual sample-wise mean yields a much larger improvement. This suggests that some languages are \emph{consensus-requiring}: They cannot be repaired well by copying one donor geometry, but can be recovered by aggregating the common structure shared across multiple languages. This is precisely the regime in which \texttt{Mid-Sample} is most valuable.

\begin{table}[!t]
\centering
\caption{\textbf{Approaching oracle performance via centroid anchor.} Comparison of
anchor strategies (Recall@1) for Korean (KO) target queries. The row
\emph{Final-layer English translation} uses the English translation of each KO query
as a strong single-language control. Our mid-layer centroid anchor consistently
outperforms any individual static anchor and approaches the theoretical Oracle bound
(dynamically picking the best anchor per sample).}
\vspace{-0.5em}
\label{tab:oracle_comparison}
\resizebox{0.95\linewidth}{!}{
\begin{tabular}{lcc}
\toprule
\textbf{Anchor Strategy} & \textbf{RefCOCOg (UMD Val)} & \textbf{RefCOCO+ (Test B)} \\
\cmidrule(lr){1-1}\cmidrule(lr){2-2}\cmidrule(lr){3-3}
Baseline (Final-layer EN Anchor) & 41.7 & 25.3 \\
\midrule
Mid-layer (EN Anchor) & 55.2 & 29.0 \\
Mid-layer (KO Anchor) & 56.5 & 26.3 \\
\rowcolor{gray!10} \textbf{Mid-layer (Centroid Anchor, Ours)} & \textbf{59.8} & \textbf{29.5} \\
\midrule
\emph{Mid-layer (Oracle)} & \emph{60.5} & \emph{32.2} \\
\bottomrule
\end{tabular}
}
\vspace{-1em}
\end{table}

\subsubsection{Approaching oracle performance via the multilingual mean.}
The same anchor pattern also clarifies why the multilingual centroid approaches the oracle-style selection strategy. Since no single donor language is universally optimal, a method that depends on choosing one anchor inevitably suffers from anchor-selection instability. By averaging across all translations of the same instance, \texttt{Mid-Sample} bypasses this bottleneck and approximates the stable component shared across languages. This observation is fully consistent with the oracle comparison in \cref{tab:oracle_comparison}: The sample-wise multilingual mean is not merely a convenient heuristic, but a principled way to approximate the best anchor without explicitly solving a discrete language-selection problem.

\subsubsection{A mechanistic interpretation of the whole picture.}
Taken together, \cref{fig:app_multi_summary,fig:app_multi_rescue,fig:app_multi_alpha,fig:app_multi_anchor} support a coherent interpretation. The multilingual failure is not primarily a translation problem and not merely an anchor-selection problem. Instead, the evidence indicates that the final-layer text representation behaves as a biased alignment interface, where aggressive interpolation is harmful, while an earlier language-stable region still preserves recoverable structure. Correcting the representation in that intermediate region using an instance-conditioned multilingual centroid produces the strongest and most reliable gains, especially for weak languages and complex expressions. This also explains why the same stabilization principle reduces language-wise prediction fragmentation in the full multilingual grounding pipeline discussed in the main paper and in \cref{app:res_multi_qual}.

\begin{table*}[ht]
\centering
\renewcommand{\arraystretch}{1.1} 
\caption{\textbf{Comprehensive cross-lingual synergy matrices.} Absolute performance gain ($\Delta$ Recall@1, $\%$) across three datasets. Columns 1--10 show the effect of injecting a \emph{single-language} anchor. The rightmost column (\textbf{ALL}) shows our proposed \ours method utilizing the all-language centroid. Blue (\pc{+}) indicates improvement, gray (\gc{-}) indicates a drop, and black (\hc{+0.00}) indicates no change. The ALL (B2G) column provides the most consistent gains overall and reduces the variance of individual language-pair effects.}
\label{tab:matrix_all_datasets}
\resizebox{\textwidth}{!}{
\begin{tabular}{l cccccccccc  c}
\toprule
\multirow{2}{*}{\textbf{Target Query}} & \multicolumn{10}{c}{\textbf{Mid-Layer Anchor Language}} & \textbf{ALL} \\
\cmidrule(lr){2-11}
 & CN & DE & DU & EN & ES & FR & IT & KO & PT & RU & \textbf{(\ours)} \\
\cmidrule(lr){1-1}\cmidrule(lr){2-11}\cmidrule(lr){12-12}
\multicolumn{12}{c}{\textbf{RefCOCO (Test A) -- Spatial Expressions}} \\
\midrule
\textbf{CN} (Chinese)   & \pc{+1.96} & \pc{+1.96} & \pc{+1.96} & \pc{+1.96} & \pc{+1.96} & \pc{+3.92} & \pc{+1.96} & \hc{+0.00} & \pc{+1.96} & \pc{+3.92} & \pc{\textbf{+9.80}} \\
\textbf{DE} (German)    & \hc{+0.00} & \hc{+0.00} & \hc{+0.00} & \hc{+0.00} & \hc{+0.00} & \hc{+0.00} & \hc{+0.00} & \hc{+0.00} & \hc{+0.00} & \hc{+0.00} & \pc{\textbf{+11.77}}\\
\textbf{DU} (Dutch)     & \pc{+3.92} & \hc{+0.00} & \hc{+0.00} & \pc{+3.92} & \pc{+1.96} & \hc{+0.00} & \pc{+3.92} & \pc{+1.96} & \pc{+1.96} & \pc{+3.92} & \pc{\textbf{+5.88}} \\
\textbf{EN} (English)   & \pc{+1.96} & \pc{+1.96} & \pc{+1.96} & \pc{+1.96} & \pc{+1.96} & \pc{+1.96} & \pc{+3.92} & \pc{+3.92} & \pc{+3.92} & \pc{+1.96} & \pc{\textbf{+3.92}} \\
\textbf{ES} (Spanish)   & \hc{+0.00} & \hc{+0.00} & \hc{+0.00} & \hc{+0.00} & \hc{+0.00} & \hc{+0.00} & \hc{+0.00} & \hc{+0.00} & \hc{+0.00} & \hc{+0.00} & \hc{\textbf{+0.00}} \\
\textbf{FR} (French)    & \pc{+1.96} & \pc{+1.96} & \pc{+1.96} & \pc{+1.96} & \pc{+1.96} & \pc{+1.96} & \pc{+1.96} & \hc{+0.00} & \pc{+1.96} & \hc{+0.00} & \pc{\textbf{+1.96}} \\
\textbf{IT} (Italian)   & \hc{+0.00} & \hc{+0.00} & \hc{+0.00} & \pc{+1.96} & \hc{+0.00} & \hc{+0.00} & \hc{+0.00} & \hc{+0.00} & \hc{+0.00} & \hc{+0.00} & \hc{\textbf{+0.00}} \\
\textbf{KO} (Korean)    & \hc{+0.00} & \pc{+1.96} & \hc{+0.00} & \pc{+3.92} & \hc{+0.00} & \pc{+3.92} & \hc{+0.00} & \hc{+0.00} & \pc{+3.92} & \hc{+0.00} & \pc{\textbf{+5.88}} \\
\textbf{PT} (Portuguese)& \pc{+3.92} & \pc{+1.96} & \pc{+1.96} & \pc{+3.92} & \hc{+0.00} & \pc{+1.96} & \pc{+1.96} & \pc{+1.96} & \pc{+1.96} & \pc{+3.92} & \pc{\textbf{+5.88}} \\
\textbf{RU} (Russian)   & \textbf{\pc{+9.80}} & \textbf{\pc{+7.84}} & \pc{+5.88} & \textbf{\pc{+11.76}}& \pc{+5.88} & \textbf{\pc{+7.84}} & \pc{+5.88} & \pc{+3.92} & \pc{+5.88} & \hc{+0.00} & \pc{\textbf{+9.80}} \\
\midrule
\multicolumn{12}{c}{\textbf{RefCOCO+ (Test B) -- Attribute-Centric Queries}} \\
\midrule
\textbf{CN} (Chinese)   & \pc{+0.46} & \pc{+0.68} & \pc{+0.46} & \pc{+1.83} & \pc{+1.83} & \pc{+1.60} & \pc{+1.14} & \pc{+1.83} & \pc{+1.14} & \pc{+0.91} & \pc{\textbf{+2.51}} \\
\textbf{DE} (German)    & \pc{+0.91} & \hc{+0.00} & \pc{+0.23} & \pc{+0.91} & \pc{+0.46} & \pc{+0.23} & \pc{+0.23} & \hc{+0.00} & \pc{+0.23} & \pc{+0.46} & \pc{\textbf{+4.11}} \\
\textbf{DU} (Dutch)     & \pc{+2.74} & \pc{+0.46} & \pc{+0.46} & \pc{+2.74} & \pc{+0.91} & \pc{+1.14} & \pc{+1.14} & \pc{+0.68} & \pc{+1.14} & \pc{+1.37} & \pc{\textbf{+2.51}} \\
\textbf{EN} (English)   & \pc{+1.14} & \pc{+0.46} & \pc{+0.46} & \pc{+0.91} & \pc{+1.60} & \pc{+0.68} & \pc{+0.46} & \pc{+0.68} & \pc{+1.14} & \pc{+1.37} & \hc{\textbf{+0.00}} \\
\textbf{ES} (Spanish)   & \pc{+1.83} & \pc{+0.23} & \hc{+0.00} & \pc{+2.05} & \pc{+0.23} & \pc{+0.23} & \pc{+0.68} & \pc{+0.68} & \hc{+0.00} & \pc{+0.68} & \pc{\textbf{+1.14}} \\
\textbf{FR} (French)    & \pc{+2.51} & \hc{+0.00} & \hc{+0.00} & \pc{+2.74} & \pc{+0.23} & \pc{+0.23} & \pc{+0.23} & \pc{+0.68} & \pc{+0.23} & \hc{+0.00} & \pc{\textbf{+1.37}} \\
\textbf{IT} (Italian)   & \pc{+2.97} & \hc{+0.00} & \pc{+0.23} & \pc{+1.83} & \pc{+1.14} & \pc{+0.68} & \pc{+0.68} & \pc{+0.91} & \pc{+1.60} & \pc{+0.91} & \pc{\textbf{+2.05}} \\
\textbf{KO} (Korean)    & \pc{+2.51} & \pc{+1.60} & \pc{+1.14} & \pc{+2.51} & \pc{+0.68} & \pc{+0.46} & \pc{+0.91} & \gc{-0.23} & \pc{+0.68} & \pc{+2.05} & \pc{\textbf{+2.97}} \\
\textbf{PT} (Portuguese)& \pc{+2.51} & \pc{+0.23} & \pc{+0.23} & \pc{+3.20} & \pc{+0.23} & \pc{+0.68} & \pc{+0.68} & \pc{+0.91} & \hc{+0.00} & \pc{+0.68} & \pc{\textbf{+1.60}} \\
\textbf{RU} (Russian)   & \pc{+3.20} & \pc{+1.37} & \pc{+1.37} & \pc{+3.42} & \pc{+0.91} & \pc{+1.14} & \pc{+2.05} & \pc{+1.37} & \pc{+0.91} & \pc{+0.23} & \pc{\textbf{+4.11}} \\
\midrule
\multicolumn{12}{c}{\textbf{RefCOCOg (UMD Test) -- Complex Descriptions}} \\
\midrule
\textbf{CN} (Chinese)   & \gc{-0.96} & \gc{-0.48} & \pc{+0.48} & \pc{+0.96} & \hc{+0.00} & \gc{-0.48} & \hc{+0.00} & \gc{-0.48} & \gc{-0.48} & \pc{+0.96} & \pc{\textbf{+7.66}} \\
\textbf{DE} (German)    & \pc{+1.44} & \gc{-0.48} & \hc{+0.00} & \pc{+0.96} & \hc{+0.00} & \hc{+0.00} & \gc{-0.48} & \pc{+0.48} & \hc{+0.00} & \pc{+0.48} & \pc{\textbf{+14.83}}\\
\textbf{DU} (Dutch)     & \textbf{\pc{+5.26}} & \pc{+0.96} & \pc{+0.48} & \pc{+3.83} & \pc{+0.96} & \pc{+0.96} & \pc{+1.44} & \pc{+2.39} & \pc{+1.91} & \pc{+2.39} & \pc{\textbf{+5.74}} \\
\textbf{EN} (English)   & \pc{+3.83} & \pc{+0.48} & \pc{+0.48} & \pc{+0.96} & \pc{+0.96} & \pc{+1.44} & \pc{+0.48} & \pc{+2.87} & \pc{+0.48} & \pc{+0.96} & \pc{\textbf{+3.35}} \\
\textbf{ES} (Spanish)   & \pc{+2.39} & \gc{-0.48} & \gc{-0.48} & \pc{+0.48} & \gc{-1.44} & \hc{+0.00} & \gc{-0.48} & \pc{+0.48} & \gc{-0.48} & \hc{+0.00} & \pc{\textbf{+1.44}} \\
\textbf{FR} (French)    & \pc{+2.87} & \gc{-0.48} & \gc{-0.48} & \pc{+0.96} & \hc{+0.00} & \gc{-0.96} & \gc{-0.48} & \gc{-0.48} & \hc{+0.00} & \gc{-0.48} & \pc{\textbf{+2.87}} \\
\textbf{IT} (Italian)   & \pc{+3.83} & \pc{+0.48} & \hc{+0.00} & \pc{+1.44} & \pc{+0.96} & \hc{+0.00} & \hc{+0.00} & \pc{+0.48} & \pc{+0.48} & \pc{+1.91} & \pc{\textbf{+4.30}} \\
\textbf{KO} (Korean)    & \pc{+1.44} & \pc{+1.44} & \pc{+0.96} & \pc{+2.87} & \hc{+0.00} & \pc{+0.96} & \pc{+0.48} & \pc{+0.48} & \hc{+0.00} & \hc{+0.00} & \pc{\textbf{+6.70}} \\
\textbf{PT} (Portuguese)& \pc{+3.83} & \pc{+1.91} & \gc{-0.96} & \pc{+1.44} & \gc{-0.96} & \gc{-0.48} & \gc{-0.48} & \pc{+0.48} & \gc{-0.48} & \hc{+0.00} & \pc{\textbf{+6.22}} \\
\textbf{RU} (Russian)   & \pc{+2.39} & \pc{+0.96} & \pc{+0.48} & \pc{+2.39} & \pc{+0.96} & \pc{+2.39} & \pc{+1.44} & \pc{+0.96} & \pc{+0.48} & \hc{+0.00} & \pc{\textbf{+4.31}} \\
\bottomrule
\end{tabular}
}
\vspace{-1.0em}
\end{table*}

\subsection{Comprehensive Cross-Lingual Synergy Matrices by Dataset}\label{app:res_multi_matrix}

To robustly demonstrate the cross-lingual synergy discussed above, we provide unified $10 \times 10$ cross-lingual matrices across RefCOCO, RefCOCO+, and RefCOCOg. \cref{tab:matrix_all_datasets} reports the absolute performance gain ($\Delta$ Recall@1, $\%$) when a single-language mid-layer global centroid from the \emph{Anchor Language} is injected into the mid-layer representation of the \emph{Target Query}. The rightmost column (\textbf{ALL}) represents our proposed \ours method, which aggregates mid-layer representations from all available languages.

\subsubsection{Synergy in spatial expressions (RefCOCO).}
Target languages that suffer from severe spatial blindness in their baseline final-layer embeddings experience explosive recovery regardless of the anchor language used. For instance, anchoring Russian (RU) with Chinese (CN) yields a $+9.80\%$ gain. As shown in the rightmost column, using the all-language centroid (\ours) robustly achieves $+9.80\%$ for RU and maximizes gains for collapsed languages like DE ($+11.77\%$), proving that aggregating multiple languages creates the most stable structural regularizer.

\subsubsection{Synergy in attribute-centric queries (RefCOCO+).}
The middle section of \cref{tab:matrix_all_datasets} shows highly uniform improvements across nearly all $100$ individual language pairs. The exceptionally dense matrix of positive gains (highlighted in blue) indicates that mid-layer stabilization fundamentally realigns attribute-centric semantics. Here again, the \textbf{ALL} (\ours) column provides the most reliable gains on average and avoids the anchor-selection instability observed in individual language-pair choices without degrading the dominant English baseline.

\subsubsection{Robustness to sequence complexity (RefCOCOg).}
Despite the increased syntactic complexity in RefCOCOg, anchoring target languages with mid-layer centroids robustly prevents representational collapse. While intricate phrasing introduces minor fluctuations in a few single-anchor pairs (\eg, ES anchored with CN drops slightly by $-0.48\%$), the \textbf{ALL} (\ours) approach perfectly absorbs these variances, yielding strictly positive gains for all languages (\eg, $+1.44\%$ for ES, $+14.83\%$ for DE).

Overall, this unified matrix provides compelling empirical evidence for \emph{Anchor Agnosticism}. The success of \ours stems from leveraging a shared, language-agnostic contextual geometry embedded natively within the mid-layers, revealing that VLEs inherently possess a universal multimodal understanding that is only masked in their final outputs.


\clearpage

\section{Further Qualitative Results}\label{app:quali}
\subsection{Multilingual RIS Qualitative Results}\label{app:res_multi_qual}

\begin{figure*}[!t]
\centering
\includegraphics[width=\textwidth]{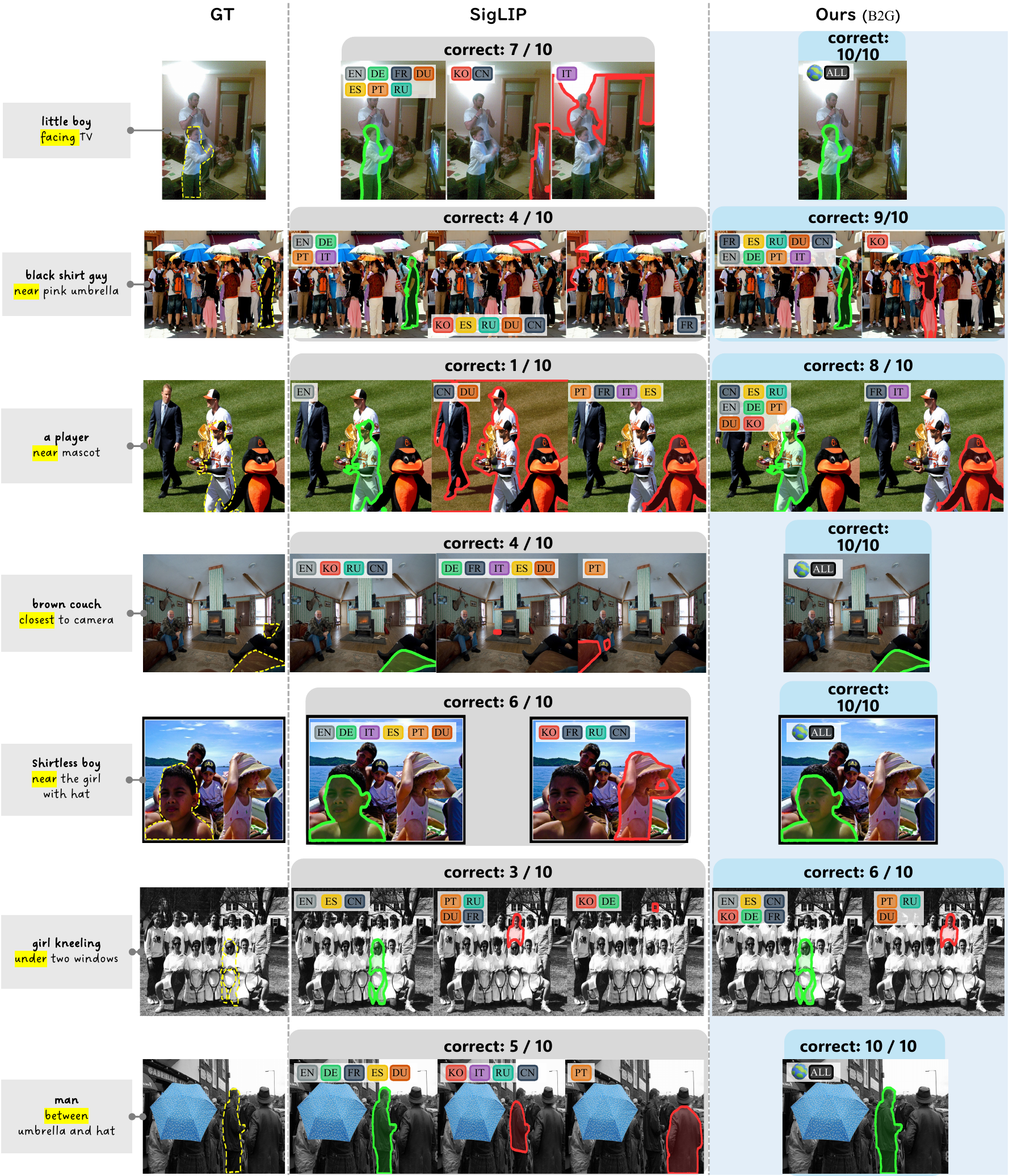}
\caption{\textbf{Supplementary multilingual qualitative results I.}
Each row shows the ground-truth mask (\textbf{GT}; yellow dashed contour), the SigLIP baseline, and \ours. Languages that produce the same prediction are grouped in the same panel, and green/red contours indicate correct/incorrect predictions, respectively. The figure highlights that the baseline often fragments into several language-specific solutions, whereas \ours typically collapses these predictions into one consistent grounding, often denoted by \texttt{ALL}.}
\label{fig:quali_ex_supp_1}
\vspace{-2em}
\end{figure*}

\begin{figure*}[!t]
\centering
\includegraphics[width=\textwidth]{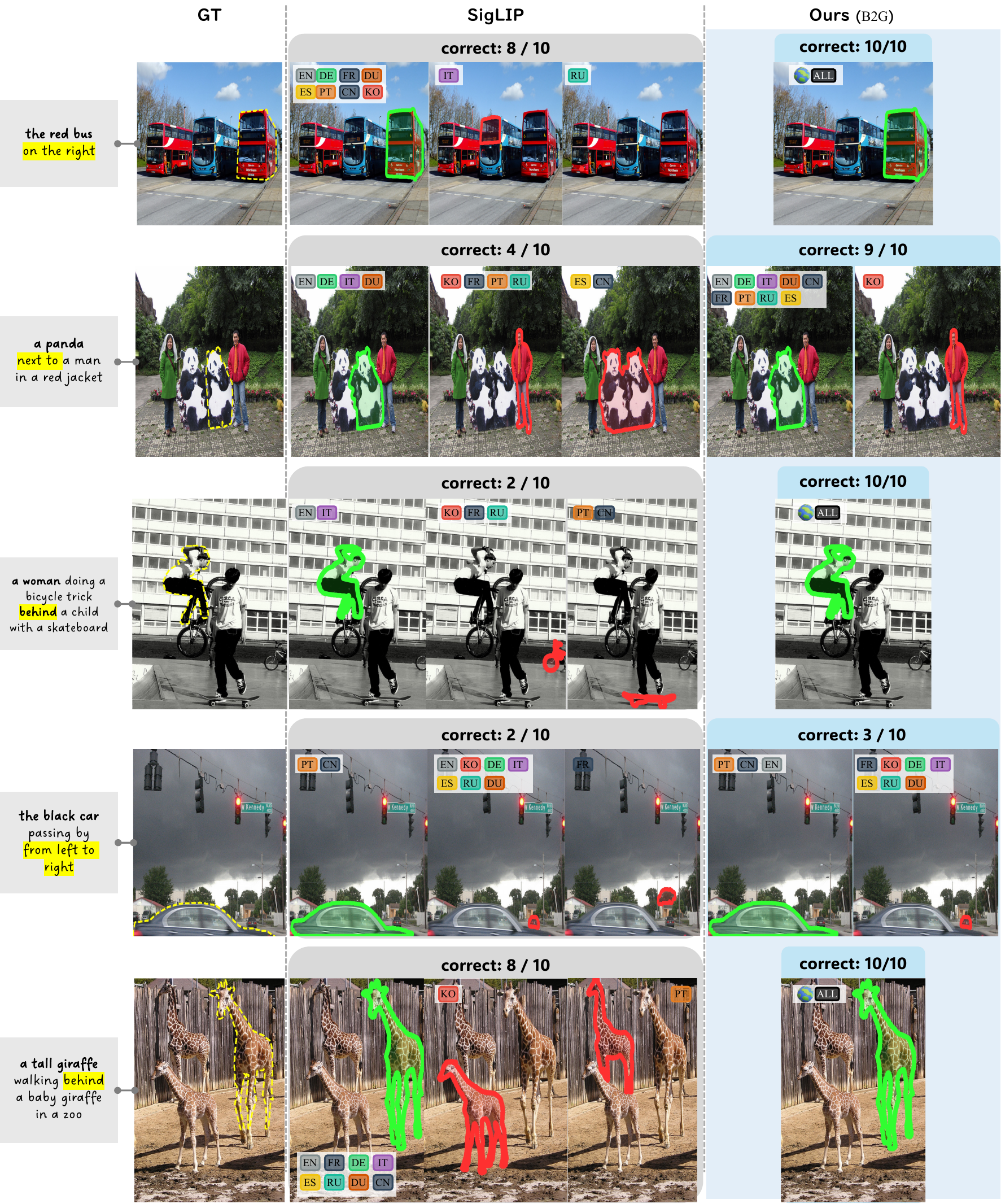}
\caption{\textbf{Supplementary multilingual qualitative results II.}
Additional examples of harder absolute-location and role-sensitive expressions. Compared with the SigLIP baseline, \ours is markedly more stable for queries such as ``the red bus on the right.'' The lower rows also illustrate remaining challenges for motion-implying or directional language, where \ours still reduces cross-language fragmentation but does not fully remove every failure mode.}
\label{fig:quali_ex_supp_2}
\vspace{-2em}
\end{figure*}

\cref{fig:quali_ex_supp_1} and \cref{fig:quali_ex_supp_2} provide additional qualitative examples. The clearest pattern is the reduction of \emph{language-wise prediction fragmentation}. In the SigLIP baseline, semantically equivalent translations often split into multiple masks, as indicated by the language tags grouped above different predictions. In contrast, \ours frequently collapses these language-dependent solutions into a single consistent prediction, often marked as \texttt{ALL}. This behavior is well aligned with our main claim that the final-layer alignment interface induces language-dependent geometric drift, whereas selecting a more language-stable intermediate text layer and injecting the multilingual centroid yields a more robust grounding signal.

This trend is particularly evident for queries requiring \emph{relative spatial reasoning}. In the first row of \cref{fig:quali_ex_supp_1} (``little boy facing TV''), the baseline is split into multiple solutions: Several languages localize the correct child, while others drift toward the adult or the doorway. Likewise, in the last row of \cref{fig:quali_ex_supp_1} (``man between umbrella and hat''), SigLIP produces multiple distinct interpretations, including nearby distractors and a far-right person, even though the relation explicitly specifies the target. In both cases, \ours resolves the ambiguity and consistently retrieves the intended object. These examples support the main paper's argument that mid-layer representations preserve structural cues that are weakened at the final layer.

Another interesting pattern is that the baseline often overweights a \emph{salient contextual noun} instead of the actual target. For instance, in \cref{fig:quali_ex_supp_1}, the query ``Shirtless boy near the girl with hat'' is correctly grounded by \ours across languages, whereas the baseline sometimes shifts to ``the girl with the hat''. A similar phenomenon appears in \cref{fig:quali_ex_supp_2} for ``a panda next to a man in a red jacket,'' where SigLIP may attend to the man in the red jacket or even to both pandas rather than to the referred panda. These examples are consistent with our multilingual formulation in the main paper: While the final-layer text feature tends to favor globally dominant semantics, the proposed mid-layer centroid injection preserves the role of the target object within the relation.

The supplementary figures also show that the benefit of \ours is not limited to catastrophic failures in non-English queries, but often acts as a \emph{cross-language structural regularizer}. In \cref{fig:quali_ex_supp_2}, ``the red bus on the right'' is already correct for many languages under the baseline, yet several outlier languages still drift to different buses. \ours removes this fragmentation and maps all translated expressions to the same rightmost bus. A similar effect is visible for ``a woman doing a bicycle trick behind a child with a skateboard,'' where the baseline occasionally attends to the child or unrelated foreground regions, while \ours consistently grounds the bicyclist behind the child. This suggests that our method improves not only the weakest languages, but also the \emph{agreement structure} among languages, which is precisely the behavior expected from a language-stable mid-layer representation.

At the same time, the supplementary results reveal informative \emph{residual failure cases}. In \cref{fig:quali_ex_supp_1}, ``black shirt guy near pink umbrella'' still leaves a Korean outlier, and ``a player near mascot'' retains a small subset of failures that attend to the mascot rather than the player. In \cref{fig:quali_ex_supp_2}, motion-implying or directional descriptions remain harder than static absolute-location cues, as some language-specific drift still persists for queries involving ``from left to right.'' These cases suggest that, although B2G substantially reduces multilingual geometric drift, extremely fine-grained role assignment and dynamic directional language remain promising directions for future work.

\subsection{Qualitative Results on Zero-shot Multilingual Text-to-Image Retrieval}
\label{app:res_multi_t2i_qual}

\begin{figure*}[!t]
    \centering
    \includegraphics[width=\textwidth]{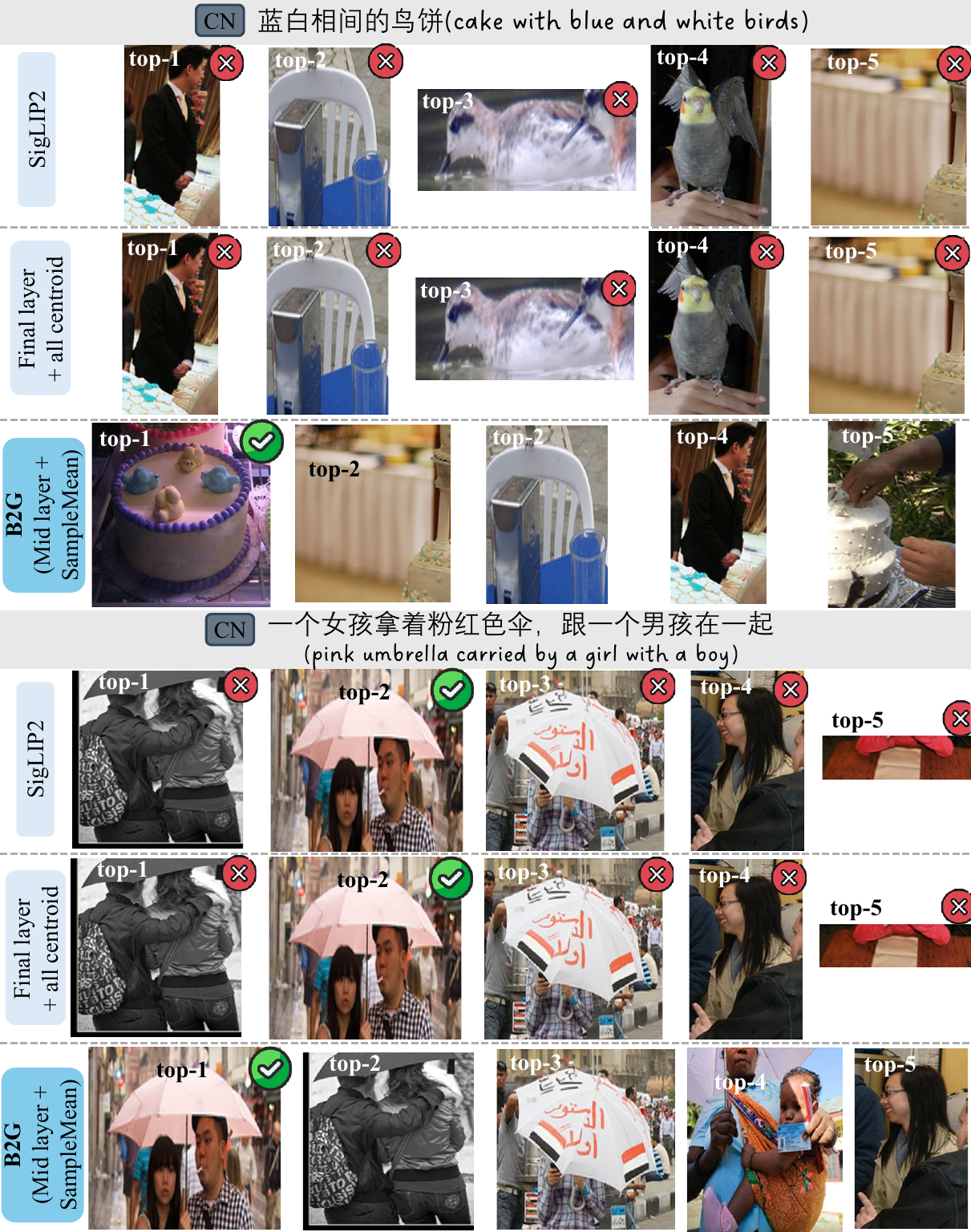}
    \caption{\textbf{Qualitative T2I retrieval results for Chinese (CN) queries.}
    The top example (``cake with blue and white birds'') shows a full rescue from
    compositional collapse: \texttt{SigLIP2} and \texttt{Final-All} miss the correct
    decorated cake in the top-5, whereas \texttt{Mid-Sample} retrieves it at
    \texttt{top-1}. The second example (``pink umbrella carried by a girl with a
    boy'') shows a milder but still meaningful rank correction, where the correct image
    rises from \texttt{top-2} to \texttt{top-1}.}
    \label{fig:t2i_qual_cn}
    \vspace{-2.5em}
\end{figure*}

\begin{figure*}[!t]
    \centering
    \includegraphics[width=\textwidth]{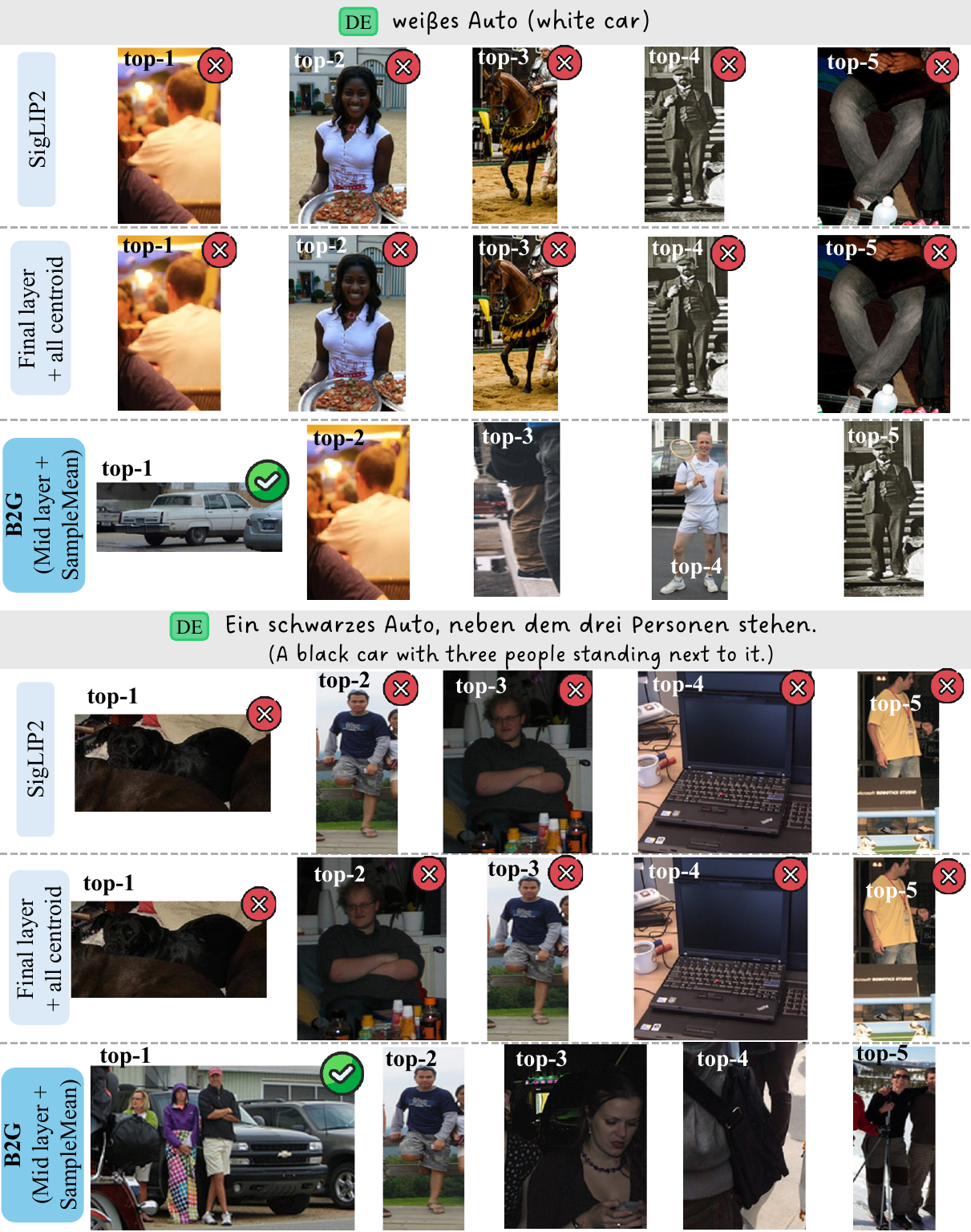}
    \caption{\textbf{Qualitative T2I retrieval results for German (DE) queries.}
    Two representative German queries illustrate catastrophic final-layer collapse.
    For both ``white car'' and ``a black car with three people standing next to it,''
    the correct image is absent from the top-5 under \texttt{SigLIP2} and remains
    absent under \texttt{Final-All}. In contrast, \texttt{Mid-Sample} retrieves the
    correct image at \texttt{top-1} in both cases.}
    \label{fig:t2i_qual_de}
    \vspace{-2.5em}
\end{figure*}

\begin{figure*}[!t]
    \centering
    \includegraphics[width=\textwidth]{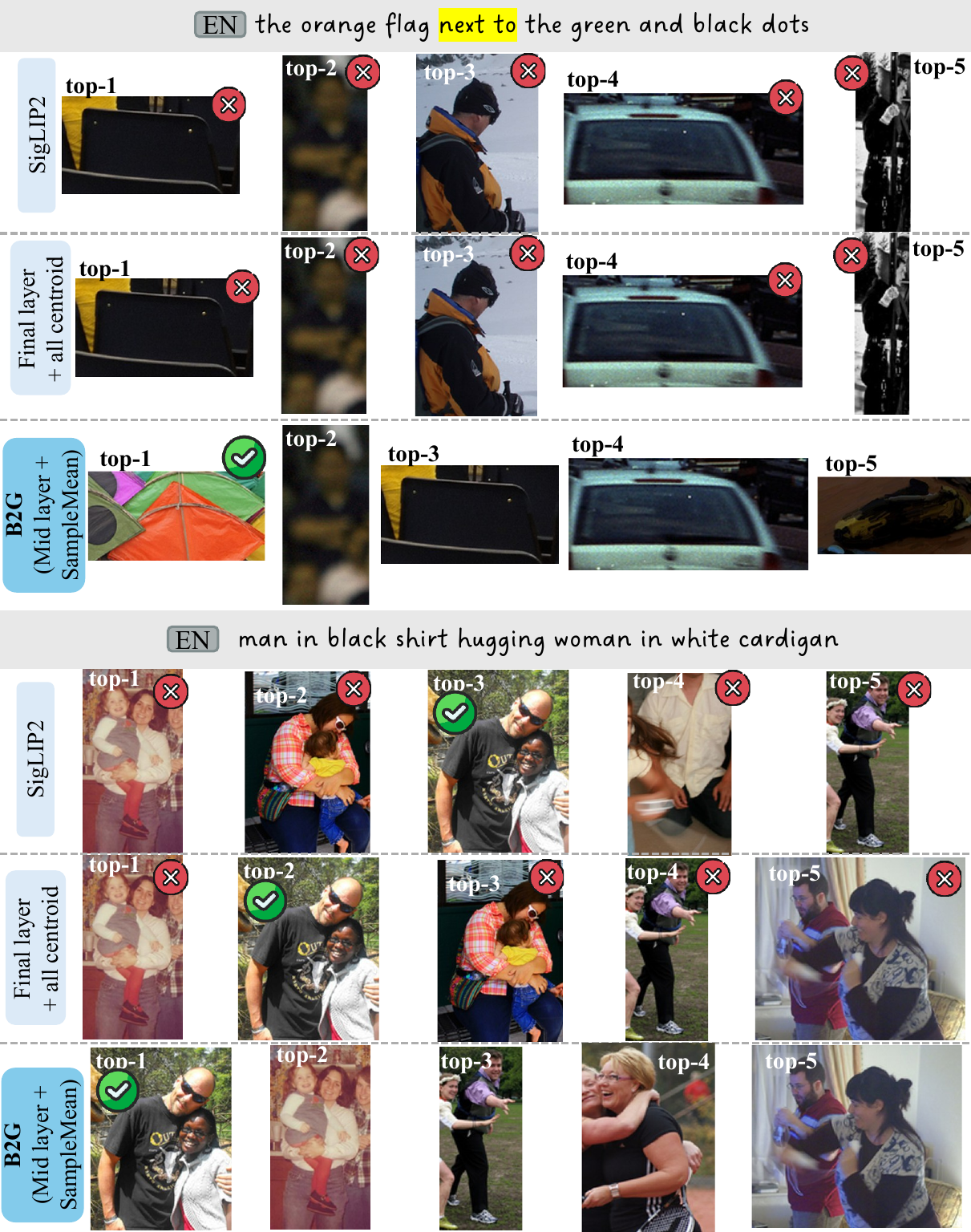}
    \caption{\textbf{Qualitative T2I retrieval results for English (EN) queries.}
    These examples confirm that there is no English trade-off. For ``the orange flag
    next to the green and black dots,'' both \texttt{SigLIP2} and \texttt{Final-All}
    fail to fully preserve the crucial spatial context, whereas \texttt{Mid-Sample}
    places the correct image at \texttt{top-1}. For ``man in black shirt hugging woman
    in a white cardigan,'' the correct image is promoted from a near miss to
    \texttt{top-1}, showing that the method sharpens compositional ranking even for the
    dominant language.}
    \label{fig:t2i_qual_en}
    \vspace{-1.5em}
\end{figure*}

\begin{figure*}[!t]
    \centering
    \includegraphics[width=\textwidth]{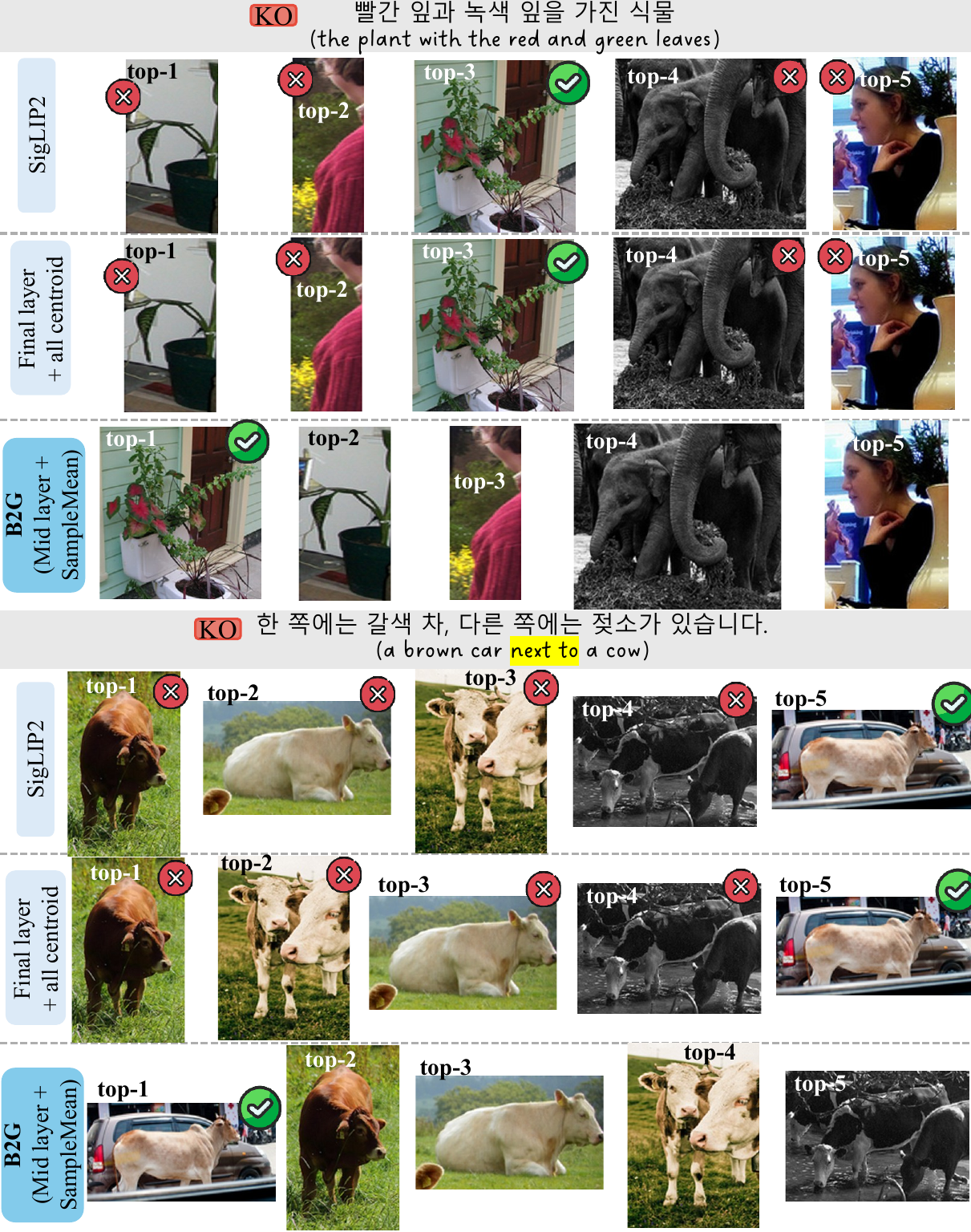}
    \caption{\textbf{Qualitative T2I retrieval results for Korean (KO) queries.}
    These examples highlight both attribute conjunction and relational binding. For
    ``the plant with the red and green leaves,'' the correct image improves from
    \texttt{top-3} to \texttt{top-1}. For ``a brown car next to a cow,'' the baseline
    retrieves isolated cows and \texttt{Final-All} still keeps the correct image only
    at \texttt{top-5}, whereas \texttt{Mid-Sample} successfully preserves the
    multi-object relation and raises the correct image to \texttt{top-1}.}
    \label{fig:t2i_qual_ko}
    \vspace{-2.5em}
\end{figure*}

\cref{fig:t2i_qual_cn}--\ref{fig:t2i_qual_ko} provide additional query-level
evidence that complements the quantitative text-side diagnostics in
\cref{app:res_multi_retr}. We present qualitative examples of text-to-image (T2I)
retrieval across four target languages: Chinese (\cref{fig:t2i_qual_cn}), German
(\cref{fig:t2i_qual_de}), English (\cref{fig:t2i_qual_en}), and Korean
(\cref{fig:t2i_qual_ko}). Each figure shows the top-5 retrieved images for the
baseline (\texttt{SigLIP2}), final-layer centroid fusion (\texttt{Final-All}), and our
proposed approach (\texttt{Mid-Sample}). Green and red markers indicate whether the
retrieved image is the paired ground-truth match. Since the image branch is strictly
fixed in this evaluation setting, the observed rank changes can be attributed directly
to text-side geometric correction.

\subsubsection{Final-layer fusion is usually too late.}
A recurring observation across the four figures is that \texttt{Final-All} usually
leaves the ranking nearly unchanged in the hardest cases and, when it does alter the
ranking, the effect is often much smaller than the change induced by \texttt{Mid-Sample}.
This visual pattern is consistent with the quantitative ablation in
\cref{app:res_multi_retr}: Final-layer interpolation is typically too late to recover
instance-level structure once language-dependent distortions have already accumulated
at the alignment interface.

\subsubsection{Collapsed-language rescue appears as a rank-level phase change.}
The clearest examples come from German and Chinese. In \cref{fig:t2i_qual_de}, both
German queries (``white car'' and ``a black car with three people standing next to
it'') are complete failures under both \texttt{SigLIP2} and \texttt{Final-All}, as the
correct image does not appear within the top-5. By contrast, \texttt{Mid-Sample}
retrieves the correct image at \texttt{top-1} in both cases. A similarly strong
recovery appears in \cref{fig:t2i_qual_cn} for ``cake with blue and white birds,''
where the baseline and \texttt{Final-All} drift toward semantically related but
structurally incorrect distractors, whereas \ours restores the correct decorated cake
to \texttt{top-1}. These examples visually explain why the largest gains are observed
for the weakest languages in the quantitative analysis.

\subsubsection{B2G also sharpens near-miss rankings.}
Not all gains are catastrophic-to-correct transitions. Some examples show a subtler
but equally informative pattern: The correct image is already present in the candidate
neighborhood, but it is under-ranked until the mid-layer correction is applied. In
\cref{fig:t2i_qual_ko}, ``the plant with the red and green leaves'' improves from
\texttt{top-3} to \texttt{top-1}, and ``a brown car next to a cow'' improves from
\texttt{top-5} to \texttt{top-1}. In \cref{fig:t2i_qual_cn}, ``pink umbrella carried
by a girl with a boy'' moves from \texttt{top-2} to \texttt{top-1}. Likewise, in
\cref{fig:t2i_qual_en}, ``man in black shirt hugging woman in white cardigan'' is
promoted from a near miss to \texttt{top-1}. These cases show that \ours is not only
rescuing collapsed languages from failure, but also refining local ranking geometry
when the correct image is already near the decision boundary.

\subsubsection{The hardest cases require compositional and relational binding.}
A shared characteristic of the successful examples is that they are not simple
object-name queries. Instead, they require preserving bindings among attributes,
objects, and relations. The English query ``the orange flag next to the green and black
dots'' in \cref{fig:t2i_qual_en} requires correct coupling between the target object and
the relational cue. The Korean query ``a brown car next to a cow'' in
\cref{fig:t2i_qual_ko} requires multi-object binding. The Chinese query ``pink umbrella
carried by a girl with a boy'' and the English query ``man in black shirt hugging woman
in white cardigan'' are even more demanding because they require role-sensitive
structure: It is not enough to retrieve images containing the right nouns; the model
must preserve who carries, who hugs, and which attributes belong to which participant.
The strong improvement of \texttt{Mid-Sample} on these examples aligns with the main
paper's interpretation that mid-layer stabilization preserves instance-specific
semantics while neutralizing language-specific drift.

\subsubsection{There is no English trade-off.}
The English examples in \cref{fig:t2i_qual_en} are especially important because they
show that the benefit of \ours is not limited to weak or collapsed languages. The first
query is a hard compositional case for both \texttt{SigLIP2} and \texttt{Final-All},
yet \ours retrieves the correct image at \texttt{top-1}. The second query shows a
milder but still meaningful refinement from a near miss to \texttt{top-1}. This is
fully consistent with the quantitative result that English also improves on average: The proposed mid-layer correction acts as a structural regularizer rather than imposing any performance trade-off.

Overall, these qualitative retrieval results support the same mechanistic picture as the
multilingual RIS examples in the main paper and in \cref{app:res_multi_qual}. The main
issue is not merely translation quality, but the fact that the final-layer text
representation behaves as a biased alignment interface. \ours corrects the query in a
more recoverable intermediate subspace, leading to more language-stable, relation-aware,
and accurate retrieval.

\section{Discussion}\label{app:dis}

\subsection{Limitations and Future Work}\label{app:dis_limit}

We provide limitations of our framework and suggest future research directions for zero-shot RIS.

\subsubsection{Spatial guidance.}
The current spatial guider~\cite{liu2025hybrid} does not always select the correct candidate among different top candidate masks. Here, our goal is not to use any other additional models except for the pretrained VLE backbone and mask generator. Future work could explore a spatial guider that can better select the final mask among the top candidate masks provided by our \texttt{P-Map}.

\subsubsection{Spatial map.}
The primary goal of this paper is to \emph{provide spatial cues} for RIS tasks using the proposed spatial map, \texttt{P-Map}, using image embeddings extracted from intermediate layers. The current spatial map could serve as a baseline for \emph{top candidate mask augmentation} to disperse the model's attention to look into multiple target objects located in different positions. We leave as a future work to extend the application of the presented spatial map into segmentation tasks that often require more refined segmentation, possibly with additional postprocessing methods~\cite{shao2024explore, bai2024self}.

\subsubsection{Parallel multilingual queries.}
Our multilingual text-side stabilization assumes access to parallel translated queries
for the same instance at test time, which enables cross-language probing and centroid
aggregation. While this setting is well-suited for controlled multilingual benchmarking
and diagnostic analysis, single-query deployment without parallel translations remains
outside the current scope. Future work could study how to approximate the same
language-stable correction from a single query or from a learned cache of language
priors.

\subsection{Societal Impact}\label{app:dis_impact}
The current study investigates zero-shot RIS, a task that requires a model to segment image regions based on free-form natural language descriptions--even for categories or concepts unseen during training. Without a need for additional training on explicit annotations, our method provides a flexible, adaptable approach applicable to many tasks related to spatial understanding. This zero-shot RIS system has the potential to significantly reduce reliance on costly human workers, paving the way for broader, scalable deployment \cite{an2023reinforcement,an2025machine}. For instance, in healthcare, zero-shot RIS could aid radiologists to highlight specific regions of interest in medical scans (\eg, ``the slightly darkened area near the lower part of the left lung'') for detailed examination. Applying zero-shot RIS to recent advanced fields could enhance the effectiveness and accessibility of vision-language systems involving precise interpretation of free-form user input of referring expressions.


\end{document}